\documentclass[]{template}

\usepackage{wrapfig}
\usepackage{makecell}
\usepackage{array}
\usepackage{multirow}
\usepackage{verbatim}
\usepackage{longtable}
\usepackage{ragged2e}
\usepackage{tablefootnote}
\usepackage{xspace}
\usepackage{lscape}
\usepackage{siunitx}
\usepackage{listings}
\usepackage{adjustbox}
\usepackage{subcaption}
\usepackage{fancyvrb}
\usepackage{fvextra}
\usepackage{float}
\usepackage{xurl}

\renewcommand{\resizebox}[3]{\adjustbox{max width={#1}}{#3}}

\makeatletter
\define@adjboxkey{width}{\@maxsizebox\height{#1}!}
\makeatother

\definecolor{graybg}{gray}{0.95}
\definecolor{subcol}{HTML}{EBF5FB}
\definecolor{catgray}{HTML}{F0F0F0}

\newcommand{\cmark}{\ding{51}}
\newcommand{\xmark}{\ding{55}}

\usepackage[capitalize,noabbrev]{cleveref}
\crefname{section}{Section}{\S\S}
\Crefname{section}{Section}{\S\S}
\crefname{table}{Table}{Tables}
\crefname{figure}{Figure}{Figures}
\crefname{equation}{eq.}{}
\crefname{appendix}{Appendix}{}
\crefformat{section}{Section #2#1#3}

\newcommand{\arxivversion}{}

\newif\ifmarkrevisions
\markrevisionsfalse
\definecolor{revred}{rgb}{0.80,0.00,0.00}
\protected\long\def\rev#1{{\ifmarkrevisions\color{revred}\fi #1}}
\newcommand{\revon}{\ifmarkrevisions\color{revred}\fi}

\newcommand\blfootnote[1]{%
  \begingroup
  \renewcommand\thefootnote{}\footnote{#1}%
  \addtocounter{footnote}{-1}%
  \endgroup
}

\setboolean{logo}{true}

\headerright{EMNLP 2026}

\title{OVO-S-Bench: A Hierarchical Benchmark for Streaming Spatial Intelligence in Multimodal LLMs}

\author[1,2,$\dagger$]{Yifei Li}
\author[3,$\dagger$]{Pengyiang Liu}
\author[2,$\ast$]{Yuhang Zang}
\author[3]{Zhongyue Shi}
\author[3]{Qi Fu}
\author[3]{Hongye Hao}
\author[1]{Jiwen Lu}

\affil[1]{Tsinghua University}
\affil[2]{Shanghai AI Laboratory}
\affil[3]{Beihang University}

\begin{abstract}
Multimodal agents in robotics, AR, and autonomous driving must reason about places and layouts from continuous egocentric streams, often using evidence outside the current view. Existing benchmarks either evaluate offline over full videos or target events rather than spatial structure. We introduce \textbf{OVO-S-Bench}, a fully human-annotated benchmark for streaming spatial intelligence, comprising 1{,}680 questions over 348 source videos. Annotation involves 12 trained annotators (each also serving as a blind cross-reviewer) across roughly 804 person-hours of multi-round quality assurance. Each question carries a query timestamp and an evidence interval, and at evaluation, the model sees only the prefix preceding the query. Questions span four levels of increasing abstraction: instantaneous egocentric perception, spatiotemporal context tracking, generative spatial reasoning, and allocentric spatial mapping. Across 38 proprietary and open-source MLLMs, Gemini-3.1-Pro \rev{trails human experts evaluated under the same prefix-access protocol by 33 points (59.2 vs.\ 92.2)}, with allocentric spatial mapping as the dominant bottleneck. Notably, streaming and spatially fine-tuned MLLMs underperform their own backbones. We further find that chain-of-thought reasoning amplifies spatial errors when ungrounded in the stream. By exposing these limitations, OVO-S-Bench establishes a demanding testbed for next-generation streaming spatial MLLMs.
\end{abstract}

\begin{document}

\blfootnote{$\dagger$ Equal contribution. \quad $\ast$ Project leader.}
\blfootnote{Project Page: \url{https://internlm.github.io/OVO-S-Bench/}}

\maketitle

\begin{figure}[h]
    \centering
    \includegraphics[width=0.95\linewidth]{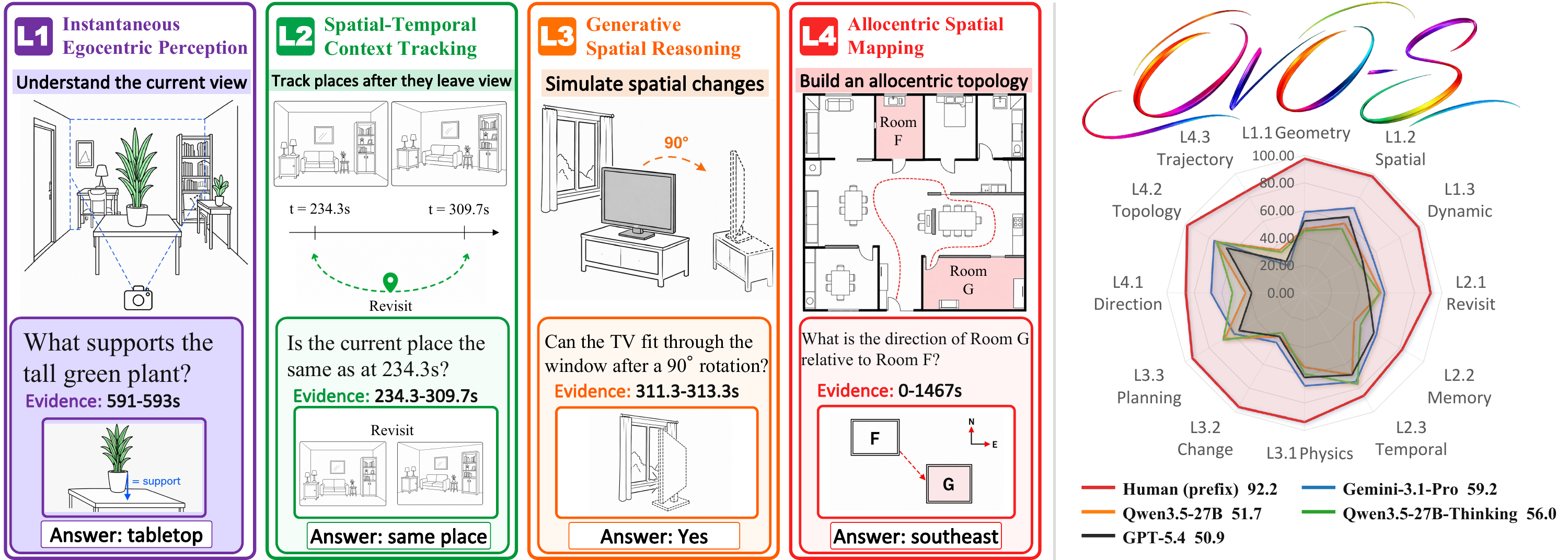}
    \caption{Overview of OVO-S-Bench. The benchmark evaluates streaming spatial understanding across four levels, from instantaneous egocentric perception and spatiotemporal context tracking to generative spatial reasoning and allocentric spatial mapping. The right panel summarizes representative model behavior across task families.}
    \label{fig:teaser}
\end{figure}

\section{Introduction}
\label{sec:intro}


\begin{table*}[t]
\centering
\footnotesize
\renewcommand{\arraystretch}{0.9}
\setlength{\tabcolsep}{3pt}
\begin{adjustbox}{width=1\textwidth}
\begin{tabular}{l c c c c c c c c c c c c}
\toprule
\textbf{Benchmark} & \textbf{Venue} & \textbf{Input} & \textbf{Stream} & \textbf{Avg.~Query Time} & \textbf{Source} & \textbf{\#Src.} & \textbf{\#Tasks} & \textbf{Anno.} & \textbf{L1} & \textbf{L2} & \textbf{L3} & \textbf{L4} \\
\midrule
\multicolumn{13}{l}{\textit{Image / Multi-image / Top-view (no video)}} \\
EmbSpatial-Bench~\citep{duan2024embspatial} & ACL'24            & Image       & \xmark & ---            & Indoor  & 3  & 6  & Auto.   & \cmark & \xmark & \xmark & \xmark \\
TopViewRS~\citep{li2024topviewrs}           & EMNLP'24          & Top-view    & \xmark & ---            & Indoor  & 1  & 9  & Hybrid  & \cmark & \xmark & \cmark & \cmark \\
MMSI-Bench~\citep{yang2025mmsi}             & NeurIPS'25        & Multi-img.  & \xmark & ---            & Mixed   & 8  & 11 & Manual  & \cmark & \cmark & \xmark & \xmark \\
\midrule
\multicolumn{13}{l}{\textit{Video (offline)}} \\
VSI-Bench~\citep{yang2025thinking}          & CVPR'25           & Video       & \xmark & 1.9~min        & Indoor  & 3  & 8  & Hybrid  & \cmark & \cmark & \cmark & \xmark \\
DISJOINT-3DQA~\citep{ravi2025disjoint}      & EMNLP'25          & Video       & \xmark & N/A            & Indoor  & 1  & 5  & Hybrid  & \cmark & \cmark & \xmark & \xmark \\
STI-Bench~\citep{sti2025bench}              & ICCV'25           & Video       & \xmark & 0.38~min       & Mixed   & 3  & 8 & Hybrid  & \cmark & \cmark & \xmark & \xmark \\
MMSI-Video-Bench~\citep{lin2025mmsivideo}   & arXiv'25          & Video       & \xmark & 1.2~min        & Mixed   & 26 & 13 & Manual  & \cmark & \cmark & \cmark & \xmark \\
VSI-SUPER~\citep{yang2025cambrians}         & ICLR'26           & Video       & \xmark & 56.6~min       & Indoor  & 3  & 2  & Hybrid  & \cmark & \cmark & \cmark & \xmark \\
\midrule
\multicolumn{13}{l}{\textit{Video (online / streaming)}} \\
StreamingBench~\citep{streamingbench2024}   & ICLR'25           & Video       & \cmark & 4.0~min        & Mixed   & 1  & 18 & Hybrid  & \cmark & \cmark & \xmark & \xmark \\
OVBench~\citep{huang2025ovbench}            & CVPR'25           & Video       & \cmark & 5.5~min        & Mixed   & 7/8  & 16 & Hybrid  & \cmark & \cmark & \xmark & \xmark \\
OVO-Bench~\citep{ovo2025bench}              & CVPR'25           & Video       & \cmark & 7.1~min        & Mixed   & 12+ & 12 & Hybrid  & \cmark & \cmark & \xmark & \xmark \\
OST-Bench~\citep{ost2025bench}              & NeurIPS'25        & Video       & \cmark & 20.12~frames   & Indoor  & 3  & 15 & Auto.   & \cmark & \cmark & \xmark & \xmark \\
ODV-Bench~\citep{odvbench2025}              & NeurIPS'25        & Video       & \cmark & 0.32~min       & Outdoor & 6  & 12 & Hybrid  & \cmark & \cmark & \cmark & \xmark \\
StreamEQA~\citep{wang2025streameqa}         & CVPR'26 Findings  & Video       & \cmark & 1.13~min       & Indoor  & 1  & 42 & Hybrid  & \cmark & \cmark & \cmark & \xmark \\
\rev{SVCBench~\citep{svcbench}}             & ECCV'26           & Video       & \cmark & 2.0~min        & Mixed   & 10 & 8  & Manual  & \cmark & \cmark & \xmark & \xmark \\
\midrule
\textbf{OVO-S-Bench (ours)}                 & EMNLP'26          & Video       & \cmark & 8.8~min        & Mixed   & 9  & 30 & Manual & \cmark & \cmark & \cmark & \cmark \\
\bottomrule
\end{tabular}
\end{adjustbox}
\vspace{-8pt}
\caption{Comparison with spatial and streaming video benchmarks. \textbf{Stream} indicates a prefix-only protocol at query time; \textbf{Avg.~Query Time} equals full video duration for offline benchmarks and the mean query timestamp (prefix length) for streaming ones (``N/A'' if not open-sourced; OST-Bench is reported in frames as it releases only sampled frames). \textbf{L1--L4} follow the taxonomy in Section~\ref{sec:taxonomy}; \cmark{} marks tasks at that level. OVO-S-Bench is the only entry with all four levels checked under a streaming protocol; per-benchmark discussion in Appendix~\ref{sec:appendix_related}.}
\label{tab:comparison}
\end{table*}

Multimodal agents acting in the physical world must reason about space from a continuous egocentric stream. A household robot fetching an item must recall where it was last seen as the camera moves between rooms, and an AR navigation assistant must guide the wearer back to a landmark they walked past minutes ago \citep{ego4d2022,lv2024aria,majumdar2024openeqa}. In each setting, the evidence required to answer a query often lies outside the current frame, requiring the agent to retain, update, and reason about spatial structure over time rather than from any single image.

Existing benchmarks leave the streaming-spatial regime untested. Spatial benchmarks study 3D relations, multi-view reasoning, and embodied question answering but assume static or offline visual context \citep{yang2025thinking,yang2025mmsi,jia2025omnispatial}, while long-video and streaming benchmarks target event understanding, narrative memory, or response timing rather than spatial structure in a continuous visual stream \citep{streamingbench2024,ovo2025bench,ost2025bench,wang2025streameqa}. No existing benchmark bridges this gap: spatial evaluation remains tied to fragmented, discrete-frame perception, while streaming evaluation targets semantic and event recall, leaving continuous spatial reasoning from egocentric video untested (Table~\ref{tab:comparison}).

We introduce \textbf{OVO-S-Bench}, a benchmark for streaming spatial intelligence over continuous egocentric video. Each question includes a query timestamp and an evidence interval, and at evaluation, the model sees only the prefix preceding the query, thereby simulating an online agent without access to future frames. Questions span four levels of increasing spatial abstraction: \textbf{L1} instantaneous egocentric perception, \textbf{L2} spatiotemporal context tracking, \textbf{L3} generative spatial reasoning, and \textbf{L4} allocentric spatial mapping (§\ref{sec:taxonomy}). Higher levels generally demand evidence that persists longer, integrates across viewpoints, and supports allocentric abstraction beyond the current frame. \rev{The scope is deliberately delimited along the response axis: adopting the prefix-only protocol of OVO-Bench~\citep{ovo2025bench}, we instantiate its \emph{Real-Time} and \emph{Backward} regimes with spatial content---answering from the current view and from evidence that has already left it---and exclude \emph{Forward} proactive timing, so that spatial state maintenance is isolated from the separate question of when a model should speak.}

Every question is written and verified by human experts. Annotators with 3D-vision backgrounds source long videos from indoor walkthroughs, egocentric activities, outdoor scenes, driving videos, and 3D environments, marking each item with options, a query timestamp, and an evidence interval. An independent group cross-reviews answerability and evidence sufficiency, and a text-only LLM probe flags items solvable from option text or world knowledge alone for revision; recurring leakage patterns are folded back into the guideline (§\ref{sec:construction}).

We evaluate 38 systems across proprietary MLLMs, general video backbones, streaming-specialized architectures, and spatially fine-tuned variants, alongside Random, Text-Only, and a Human baseline from five independent participants (§\ref{sec:main_results}). Three findings stand out. \emph{Allocentric spatial mapping is the dominant bottleneck:} L4 is the lowest-scoring level for 32 of 38 systems, and even Gemini-3.1-Pro (59.2 overall) \rev{trails the human baseline matched to its own prefix-access mode by 33 points, and the more constrained human streaming baseline by 27.} \emph{Specialized methods lag behind their backbones:} 13 of 15 streaming and spatially fine-tuned methods score below their base, with L4 as the most uniformly damaged level (mean $\Delta{=}-6.1$). \emph{Thinking gains are model- and level-dependent:} explicit reasoning aids L2 cross-frame integration (mean $\Delta{=}+3.9$) but degrades L1 current-view perception (mean $\Delta{=}-1.0$), and failure traces show that 60--80\% of wrong answers stem from ungrounded spatial claims.

\section{Related Work}
\paragraph{Benchmarking Visual Spatial Understanding.}
Spatial benchmarks span image, multi-image, and video regimes. Image and top-view benchmarks evaluate metric estimation, object relations, and 3D geometry on static scenes \citep{chen2024spatialvlm,jia2025omnispatial,li2024topviewrs}; MMSI-Bench \citep{yang2025mmsi} extends to multi-image spatial intelligence; and video benchmarks add temporal observations for spatial recall \citep{yang2025thinking}, spatio-temporal precision \citep{sti2025bench}, and dynamic camera/object motion \citep{zhang2025dsi}. Hierarchical efforts organize spatial abilities into progressively harder levels \citep{xu2025spatialbench,xiao2025spatialtree}\rev{, and concurrent work scales open-world spatial evaluation to physical dynamics and human intent \citep{gu2026escherverse}}. All assume offline visual context: models can re-attend to any frame at query time, so the streaming constraint that makes spatial evidence ephemeral never surfaces.

Embodied and egocentric settings sharpen the need for persistent spatial memory. OpenEQA \citep{majumdar2024openeqa} frames embodied QA around episodic memory and active exploration, EmbSpatial-Bench \citep{duan2024embspatial} and Open3D-VQA \citep{zhang2025open3dvqa} evaluate spatial relations in embodied or aerial environments, and DisJoint-3DQA \citep{ravi2025disjoint} studies disjoint-frame reasoning where queried objects are not co-visible. These settings still operate over a curated context window rather than a continuous causal stream, leaving open whether models sustain a spatial model across arbitrary query timestamps.

\paragraph{Multimodal Models for Spatial Reasoning.}
Methods for spatial reasoning in MLLMs cluster around three strategies. \emph{Data scaling} fine-tunes general-purpose backbones on large spatial corpora: VST pairs a 4M perception set with a 135K reasoning set under SFT+RL \citep{yang2025vst}, SenseNova-SI extends the recipe to 8M samples \citep{cai2025sensenovasi}, and Holi-Spatial mines spatial supervision from raw video streams \citep{gong2026holispatial}. \emph{Architectural augmentation} injects 3D priors via spatially grounded supervision \citep{chen2024spatialvlm}, dedicated spatial tokens \citep{wu2025spatialmllm}, or camera-aware fusion \citep{spacemind2026}. \emph{Inference-time adaptation} reshapes how the stream is consumed: Cambrian-S frames spatial supersensing as predictive world modeling with an active-memory manager \citep{yang2025cambrians}, and Spatial-TTT adapts fast weights at test time \citep{spatialttt2026}.

\paragraph{Streaming Video Understanding.}
Streaming video understanding has two arms. \emph{Long-form benchmarks}, including EgoSchema, LongVideoBench, and LVBench, test temporal understanding under offline access \citep{egoschema2023,longvideobench,wang2025lvbench,trace,videommelogical}. \emph{Streaming benchmarks}, including StreamingBench, OVBench, OVO-Bench, OST-Bench, and \rev{SVCBench} \citep{streamingbench2024,huang2025ovbench,ovo2025bench,ost2025bench,svcbench}, enforce causal input and bounded memory but target event understanding, multi-turn interaction, or counting rather than spatial structure. \rev{Concurrent streaming benchmarks extend the same causal constraint to omnimodal mobile assistants \citep{lu2026phostream}, egocentric episodic memory \citep{forte2026egostream}, and ad-hoc recall over text dialogue streams \citep{wang2026proactivememory}, each holding spatial structure outside its scope.} Streaming MLLMs respond with memory banks, token pruning, KV-cache management, and event-segment memories \citep{qian2024videostreaming,zhang2024flashvstream,zhang2025flash,yao2025timechatonline,odvbench2025,xie2026fluxmem,qian2025dispider}, and recent work asks whether such designs beat sliding-window baselines \citep{shen2026simplestream}. All target event memory; persistent spatial evidence over a continuous egocentric stream is the dimension OVO-S-Bench makes empirical for the first time.

\section{OVO-S-Bench}
\label{sec:taxonomy}
\begin{figure*}[t]
    \centering
    \includegraphics[width=1\linewidth]{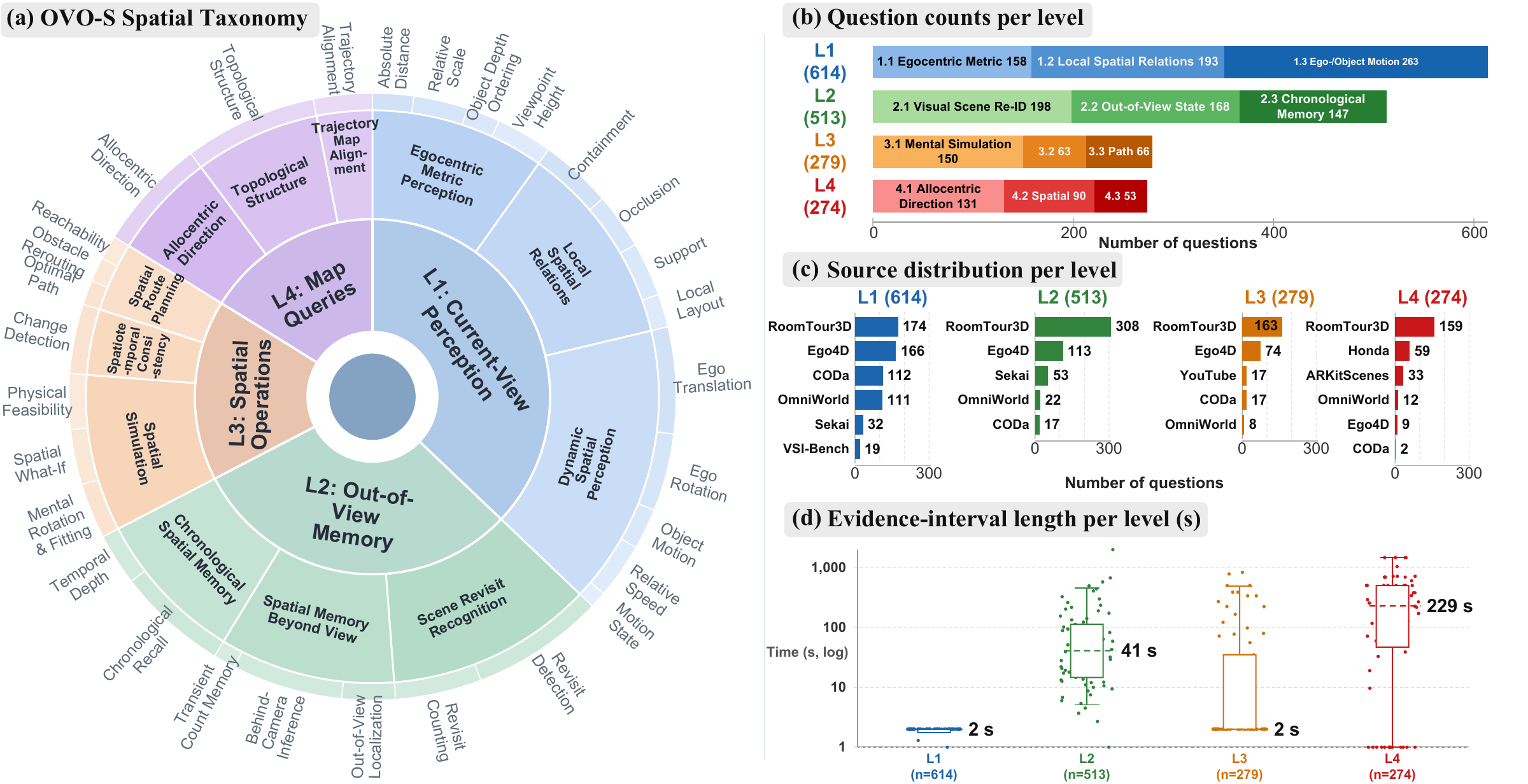}
    \vspace{-10pt}
    \caption{Taxonomy and benchmark statistics for OVO-S-Bench. The left panel gives the four-level spatial taxonomy, while the right panels report task-family counts, source distribution, and evidence-interval lengths by level.}
    \label{fig:taxonomy}
    \vspace{-10pt}
\end{figure*}

OVO-S-Bench organizes questions into four levels by the spatial state a model must access at query time (Figure~\ref{fig:taxonomy}). The levels progress from evidence directly available in the current view to allocentric map queries that require cross-viewpoint integration, reflecting a gradient of persistence and abstraction. Each level contains several task families and canonical task types; full definitions and examples are provided in Appendix~\ref{sec:appendix_categories}.

\subsection{Four-Level Streaming Spatial Taxonomy}
\label{sec:taxonomy_hier}

\begin{figure*}[t]
    \centering
    \includegraphics[width=1\textwidth]{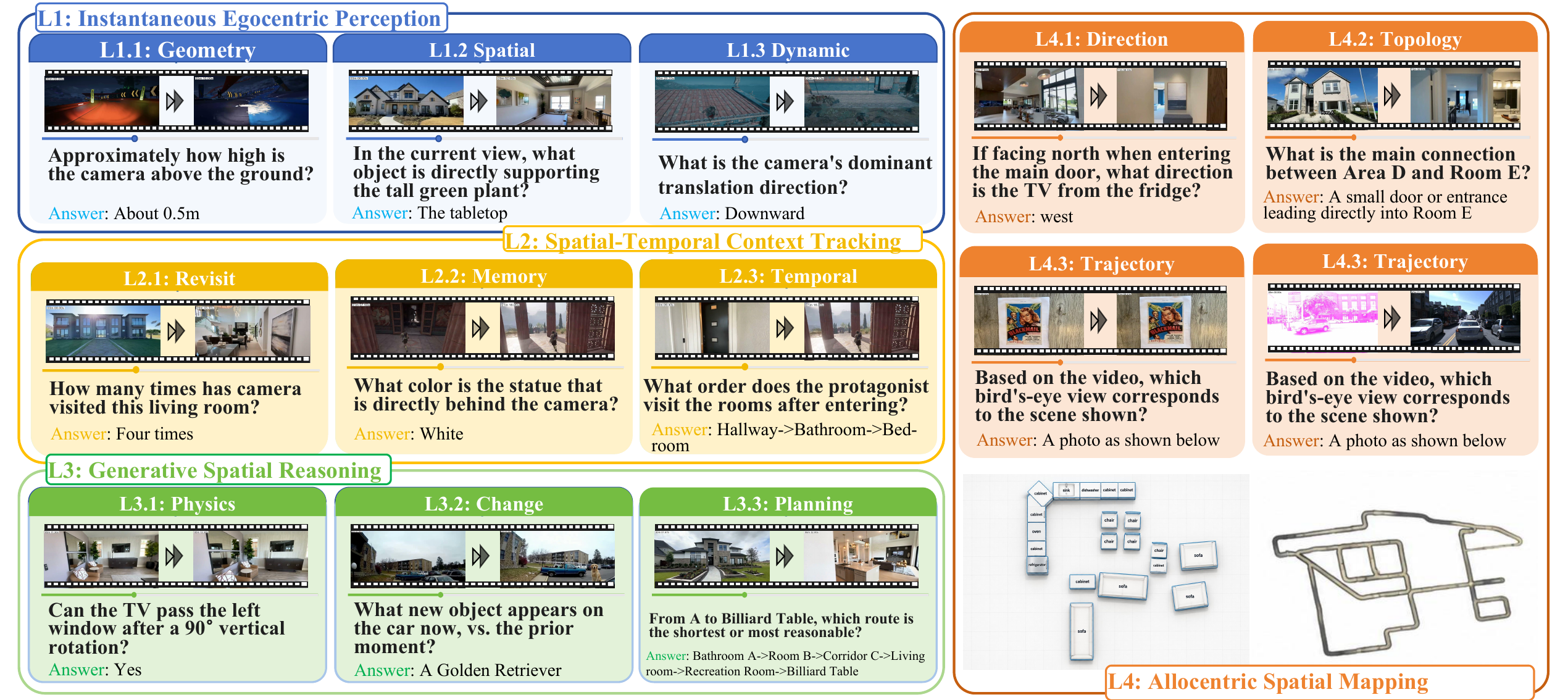}
    \vspace{-10pt}
    \caption{Representative OVO-S-Bench examples. Each card pairs a spatial question with visual evidence, illustrating the progression from current-view perception to allocentric spatial mapping.}
    \label{fig:taxonomy_examples}
    \vspace{-10pt}
\end{figure*}

\paragraph{L1: Instantaneous Egocentric Perception.}
L1 tests whether a model understands the spatial structure currently in view. Questions can be answered from the frames near the query timestamp alone, without recalling any past observation. Task families cover egocentric metric perception (distance, scale, clearance, viewpoint height), local spatial relations (containment, occlusion, support, visible layout), and dynamic spatial perception (camera motion, object motion, relative speed).


\paragraph{L2: Spatiotemporal Context Tracking.}
L2 evidence has appeared in the video prefix but is no longer visible at query time. The model must keep spatial facts bound to specific places, objects, and timestamps after visual support disappears. Task families include scene revisit recognition (whether a place has been seen and how often), spatial memory beyond the view (location and state of out-of-view entities), and chronological spatial memory (visit order and temporal-depth comparisons).


\paragraph{L3: Generative Spatial Reasoning.}
L3 requires the model to generatively operate on spatial structure rather than merely retrieve an observation. Questions involve mental rotation, state-change inference, and path feasibility; evidence spans vary by task, from local for simulation items to extended for route planning. Task families cover spatial simulation (reorientation, removal consequences, physical feasibility), spatiotemporal consistency verification (whether a state has changed between observations), and spatial route planning (route selection, detours, reachability).


\paragraph{L4: Allocentric Spatial Mapping.}
L4 requires the model to integrate the egocentric stream into an allocentric representation and query its global structure. Evidence typically spans multiple viewpoints and may cover the entire explored region. Three task families target allocentric direction reasoning (global bearings among rooms or landmarks), topological structure reasoning (adjacency, connectivity, boundaries, and transitions), and trajectory-map alignment (matching the first-person path to a bird's-eye map).

\subsection{Benchmark Construction}
\label{sec:construction}
\noindent\textbf{Video sources.}
OVO-S-Bench draws from 9 publicly available or accessible sources covering five regimes: \emph{indoor walkthroughs} (RoomTour3D~\citep{roomtour3d2024}), \emph{egocentric activities} (Ego4D~\citep{ego4d2022}), \emph{outdoor/world scenes} (Sekai~\citep{sekai2025}, OmniWorld~\citep{omniworld2025}, and YouTube walking tours), \emph{driving videos} (CODa~\citep{coda2022} and Honda HDD~\citep{hdd2018}), and \emph{spatially annotated 3D environments} (ARKitScenes~\citep{arkitscenes2021} and VSI-Bench~\citep{yang2025thinking}).

\noindent\textbf{Human annotators write every item.}
Auto-generated captions and detections rarely capture the spatial evidence our targets require. Annotators select clips from these sources with stable motion, clear viewpoints, and sufficient spatial variation at the target level. For each item, they record the video, task label, question, options, answer, query timestamp, and evidence interval. Some task types employ specialized construction techniques such as image editing to generate spatial-change contrasts.

\noindent\textbf{Each item follows the streaming setting.}
The answer must be derivable from the video prefix before the query timestamp. Annotators mark the shortest interval that contains the needed evidence and write distractors that are plausible under the visual context but wrong under the annotated evidence.

\noindent\textbf{Quality control removes shortcuts.}
We first run a text-only model to flag items that leak the answer through wording, common sense, option asymmetry, or a single lexical cue; this matters most for L2 through L4, where room priors and route stereotypes create shortcuts. We then probe the item under the intended streaming setting to expose ambiguity, easy visual shortcuts, and timestamp errors. A second annotator reviews the item without seeing the original answer and checks whether the answer and evidence interval are sufficient. Disagreements are revised, and recurring problems are folded back into the guideline. The full protocol, source details, and construction techniques are in Appendix~\ref{sec:appendix_protocols}.

\subsection{Benchmark Statistics}
\label{sec:statistics}
The released benchmark comprises 1{,}680 questions over 348 source videos from 9 datasets, organized into 30 canonical task types across four levels. \rev{Questions and scored queries differ in count: Revisit Counting asks the same question at several query timestamps, where the correct answer grows as revisits accumulate, so the 1{,}695-question public release is scored as 1{,}722 queries. Appendix tables computed on the public release state that scope explicitly; Table~\ref{tab:overall_results} and every main-text number use the 1{,}680-item evaluation set.} The mean prefix at query time is 8.8 minutes. Evidence-span duration varies by level (L1 median 2.0\,s, L2 36.8\,s, L3 2.0\,s, L4 278.7\,s), reflecting the spatial persistence required at each level. Figure~\ref{fig:taxonomy}b--d reports per-level task-family counts, source distribution, and evidence-interval lengths; L4 is reported as three map-query families.

\section{Experiments}
\label{sec:experiments}
\subsection{Evaluation Setup}
\label{sec:setup}
All systems see only the video prefix before query timestamp $t_q$, but we distinguish two evaluation modes. Under the \emph{prefix-access} mode (default for proprietary, open-source, spatially fine-tuned, and embodied models), 128 frames are uniformly sampled from $[0, t_q]$ and presented together with the question; the model may attend freely over all sampled frames. This setup follows OVO-Bench and OST-Bench~\citep{ovo2025bench,ost2025bench}. Under the \emph{streaming} mode (applied to streaming-architecture and token-compression models), the video is fed sequentially at each model's published rate, and all memory and compression decisions are \emph{query-agnostic}, made before the question arrives (Appendix~\ref{subsec:eval_streaming}). Proprietary models are queried through their official APIs, all other models run locally with their published defaults, and answers are extracted by regular expression without further post-processing.

\rev{Every scored query also carries a response type on the axis OVO-Bench defines: L1 answers from the current view (\emph{Real-Time}), L2 and L4 recall or integrate observations that have already left it (\emph{Backward}), and L3 splits per query by the gap between the end of the evidence interval and the query timestamp. The benchmark is 51\% Real-Time and 49\% Backward, and instantiates no \emph{Forward} proactive-timing queries, matching the scope stated in Section~\ref{sec:intro}. Appendix~\ref{sec:appendix_response_type} gives the labeling rule, the per-level counts, and the sensitivity of the split to the L3 gap threshold.}

We evaluate 38 systems (Tab.~\ref{tab:overall_results}) spanning six families: \emph{proprietary MLLMs} (GPT-5.4~\citep{openai2026gpt54}, Gemini-3.1-Pro/Flash-Lite~\citep{google2026gemini31pro,google2026gemini31flashlite}, Grok-4.1-Fast~\citep{xai2025grok41fast}); \emph{open-source general backbones} (InternVL-3.5~\citep{wang2025internvl3}, Qwen2.5-VL~\citep{Bai2025Qwen25VLTR}, Qwen3-VL~\citep{bai2025qwen3}, Qwen3.5~\citep{qwen2026qwen35}, Gemma-4~\citep{gemma4_2026}, GLM-4.6V-Flash~\citep{zeng2025glm}); \emph{streaming video MLLMs} (Flash-VStream~\citep{zhang2024flashvstream}, StreamForest~\citep{odvbench2025}, StreamingVLM~\citep{xu2025streamingvlm}); \emph{token-compression and memory methods} (InfiniPot-V~\citep{kim2026infinipot}, HERMES~\citep{zhang2026hermes}, FluxMem~\citep{xie2026fluxmem}, StreamingTOM~\citep{chen2025streamingtom}); \emph{spatially fine-tuned MLLMs} (Spatial-MLLM~\citep{wu2025spatialmllm}, Cambrian-S$\pm$LFP~\citep{yang2025cambrians}, VST-SFT/RL~\citep{yang2025vst}, SenseNova-SI~\citep{cai2025sensenovasi}, Spatial-TTT~\citep{spatialttt2026}); and \emph{embodied foundation models} (Cosmos-Reason1~\citep{azzolini2025cosmos}, VeBrain~\citep{luo2025visual}, RynnBrain~\citep{dang2026rynnbrain}, RoboBrain2.5~\citep{tan2026robobrain}). We anchor accuracy with three baselines: \emph{Random}, \emph{Text-Only} (GPT-5.4 with question and options only), and a \emph{Human} baseline evaluated by five independent participants on the full question set under both modes: in streaming mode, participants watch the video without knowing the question and must rely on memory; in prefix-access mode, participants receive the question alongside the video prefix and may re-scan freely. \rev{Each family is compared against the human baseline collected under its own mode: \emph{Human (streaming)} for the streaming-architecture and token-compression families, whose memory decisions are likewise query-agnostic, and \emph{Human (prefix-access)} for all remaining families.}

\begin{table*}[t]
\centering
\tiny
\setlength{\tabcolsep}{1.15pt}
\renewcommand{\arraystretch}{0.88}
\definecolor{rankGreenOne}{HTML}{8EC9A3}
\definecolor{rankGreenTwo}{HTML}{C5E5D0}
\definecolor{rankGreenThree}{HTML}{E8F4EC}
\definecolor{rankGrayOne}{HTML}{C8CDD3}
\definecolor{rankGrayTwo}{HTML}{DDE2E7}
\definecolor{rankGrayThree}{HTML}{EEF1F4}
\definecolor{groupGray}{HTML}{F3F6F8}
\begin{adjustbox}{width=1\textwidth}
\begin{tabular}{l@{\hskip 0.5pt}c | c c c c c | c c c c c | c c c c c | c c c c c | c c}
\toprule
\multirow{2}{*}{\textbf{Model}} & \multirow{2}{*}{\textbf{Params}} & \multicolumn{5}{c|}{\textbf{L1: Instant.\ Ego.\ Perc.}} & \multicolumn{5}{c|}{\textbf{L2: Context Tracking}} & \multicolumn{5}{c|}{\textbf{L3: Gen.\ Spatial Reas.}} & \multicolumn{5}{c|}{\textbf{L4: Alloc.\ Spatial Map.}} & \multirow{2}{*}{\textbf{Avg.}} & \multirow{2}{*}{\textbf{Rank}} \\
\cmidrule(lr){3-7}\cmidrule(lr){8-12}\cmidrule(lr){13-17}\cmidrule(lr){18-22}
 &  & 1.1 & 1.2 & 1.3 & Avg. & R & 2.1 & 2.2 & 2.3 & Avg. & R & 3.1 & 3.2 & 3.3 & Avg. & R & 4.1 & 4.2 & 4.3 & Avg. & R &  &  \\
\midrule
\rowcolor{groupGray}\multicolumn{24}{l}{\textit{Baseline \& Controls}} \\
Text-Only (GPT-5.4) & -- & 38.64 & 41.37 & 35.21 & 38.41 & -- & 36.01 & 30.08 & 40.78 & 35.62 & -- & 54.00 & 9.52 & 53.03 & 38.85 & -- & 31.68 & 54.17 & 20.75 & 35.53 & -- & 37.10 & -- \\
Random Baseline & -- & 33.93 & 25.49 & 30.00 & 29.81 & -- & 37.44 & 26.77 & 41.08 & 35.10 & -- & 50.00 & 25.00 & 25.00 & 33.33 & -- & 25.00 & 31.25 & 25.00 & 27.08 & -- & 31.33 & -- \\
\rev{Human (prefix-access)} & -- & 97.47 & 97.93 & 95.44 & 96.95 & -- & 91.41 & 81.55 & 85.71 & 86.22 & -- & 93.33 & 95.24 & 93.94 & 94.17 & -- & 86.14 & 98.33 & 83.02 & 89.16 & -- & 92.20 & -- \\
Human (streaming) & -- & 91.77 & 95.85 & 92.02 & 93.21 & -- & 86.87 & 77.98 & 78.23 & 81.03 & -- & 88.67 & 87.30 & 83.33 & 86.43 & -- & 79.21 & 86.67 & 71.70 & 79.19 & -- & 86.61 & -- \\
\midrule
\rowcolor{groupGray}\multicolumn{24}{l}{\textit{Closed-source proprietary MLLMs}} \\
GPT-5.4 & -- & 52.08 & 63.70 & 48.13 & 54.64 & 4 & 46.25 & 57.92 & 68.61 & 57.59 & \cellcolor{rankGrayTwo}2 & 61.33 & 36.51 & 54.55 & 50.80 & 9 & 32.67 & 62.50 & 26.42 & 40.53 & 17 & 50.89 & 5 \\
Gemini-3.1-Pro & -- & 58.69 & \textbf{71.18} & \textbf{55.89} & \textbf{61.92} & \cellcolor{rankGrayOne}1 & \textbf{57.92} & \textbf{59.97} & \textbf{74.23} & \textbf{64.04} & \cellcolor{rankGrayOne}1 & \textbf{67.33} & 41.27 & 59.09 & 55.90 & \cellcolor{rankGrayThree}3 & \textbf{60.40} & \textbf{81.67} & 22.64 & \textbf{54.90} & \cellcolor{rankGrayOne}1 & \textbf{59.19} & \cellcolor{rankGreenOne}1 \\
Gemini-3.1-Flash-Lite & -- & 51.19 & 63.38 & 47.70 & 54.09 & 5 & 47.98 & 47.61 & 60.98 & 52.19 & 8 & 58.67 & \textbf{49.21} & 54.55 & 54.14 & 5 & 35.64 & 68.33 & 24.53 & 42.84 & 7 & 50.81 & 7 \\
Grok-4.1-Fast & -- & 48.34 & 49.95 & 36.13 & 44.80 & 17 & 42.26 & 42.03 & 55.48 & 46.59 & 19 & 56.00 & 34.92 & 54.55 & 48.49 & 15 & 24.75 & 55.83 & 24.53 & 35.04 & 25 & 43.73 & 19 \\
\midrule
\rowcolor{groupGray}\multicolumn{24}{l}{\textit{Open-source general video MLLMs}} \\
InternVL-3.5 & 8B & 47.86 & 54.60 & 35.31 & 45.92 & 13 & 40.95 & 40.31 & 56.22 & 45.83 & 21 & 54.00 & 26.98 & 60.61 & 47.20 & 18 & 34.65 & 55.00 & 28.30 & 39.32 & 18 & 44.57 & 16 \\
InternVL-3.5 & 38B & 58.64 & 55.56 & 49.74 & 54.65 & \cellcolor{rankGrayThree}3 & 50.30 & 48.00 & 64.91 & 54.40 & 7 & 61.33 & 22.22 & 53.03 & 45.53 & 22 & 38.61 & 68.33 & 18.87 & 41.94 & 10 & 49.13 & 8 \\
InternVL-3.5 & 241B-A28B & \textbf{61.30} & 59.36 & 45.98 & 55.55 & \cellcolor{rankGrayTwo}2 & 48.75 & 49.86 & 68.54 & 55.72 & \cellcolor{rankGrayThree}3 & 53.33 & 33.33 & 68.18 & 51.62 & 8 & 42.57 & 65.83 & 13.21 & 40.54 & 16 & 50.85 & 6 \\
Qwen2.5-VL & 7B & 42.89 & 40.26 & 38.87 & 40.67 & 28 & 40.24 & 36.31 & 59.98 & 45.51 & 22 & 54.67 & 23.81 & 59.09 & 45.86 & 21 & 47.52 & 65.83 & 20.75 & 44.70 & 5 & 44.19 & 17 \\
Qwen3-VL & 4B & 32.91 & 53.19 & 43.65 & 43.25 & 20 & 47.26 & 34.25 & 63.04 & 48.18 & 14 & 57.33 & 38.10 & 68.18 & 54.54 & 4 & 25.74 & 70.00 & 28.30 & 41.35 & 11 & 46.83 & 10 \\
Qwen3-VL & 32B & 48.83 & 57.07 & 44.51 & 50.14 & 8 & 50.71 & 34.58 & 70.43 & 51.91 & 9 & 60.67 & 26.98 & 68.18 & 51.94 & 7 & 32.67 & 68.33 & 22.64 & 41.22 & 12 & 48.80 & 9 \\
Qwen3-VL & 235B-A22B & 50.36 & 56.09 & 50.91 & 52.45 & 6 & 49.76 & 45.61 & 70.17 & 55.18 & 5 & 63.33 & 46.03 & 74.24 & \textbf{61.20} & \cellcolor{rankGrayOne}1 & 45.54 & 70.83 & 20.75 & 45.71 & \cellcolor{rankGrayThree}3 & 53.64 & \cellcolor{rankGreenTwo}2 \\
Qwen3.5 & 4B & 40.28 & 51.13 & 44.82 & 45.41 & 15 & 47.44 & 34.19 & 62.98 & 48.20 & 13 & 52.00 & 39.68 & 56.06 & 49.25 & 14 & 27.72 & 65.83 & 22.64 & 38.73 & 19 & 45.40 & 13 \\
Qwen3.5 & 9B & 43.25 & 55.03 & 44.13 & 47.47 & 11 & 44.52 & 38.53 & 65.05 & 49.37 & 12 & 56.67 & 33.33 & 60.60 & 50.20 & 11 & 25.74 & 63.33 & 20.75 & 36.61 & 22 & 45.91 & 12 \\
Qwen3.5 & 27B & 46.50 & 58.13 & 49.88 & 51.50 & 7 & 55.24 & 41.53 & 68.67 & 55.15 & 6 & 54.00 & 36.51 & 66.67 & 52.39 & 6 & 35.64 & 71.67 & \textbf{35.85} & 47.72 & \cellcolor{rankGrayTwo}2 & 51.69 & 4 \\
Qwen3.5 & 397B-A17B & 46.71 & 53.32 & 48.68 & 49.57 & 9 & 55.65 & 41.22 & 69.30 & 55.39 & 4 & 57.33 & 38.10 & \textbf{78.79} & 58.07 & \cellcolor{rankGrayTwo}2 & 34.65 & 75.00 & 26.42 & 45.36 & 4 & 52.10 & \cellcolor{rankGreenThree}3 \\
Gemma-4 & E2B & 42.12 & 42.36 & 31.93 & 38.80 & 33 & 35.00 & 31.97 & 42.60 & 36.52 & 35 & 55.33 & 12.70 & 50.00 & 39.34 & 33 & 28.71 & 37.50 & 22.64 & 29.62 & 35 & 36.07 & 36 \\
Gemma-4 & E4B & 41.33 & 47.33 & 34.10 & 40.92 & 26 & 41.43 & 38.69 & 48.16 & 42.76 & 30 & 52.00 & 14.29 & 62.12 & 42.80 & 27 & 24.75 & 47.50 & 24.53 & 32.26 & 32 & 39.69 & 30 \\
Gemma-4 & 26B-A4B & 47.90 & 56.23 & 43.86 & 49.33 & 10 & 47.26 & 43.31 & 49.10 & 46.55 & 20 & 62.67 & 25.40 & 46.97 & 45.01 & 25 & 19.80 & 55.00 & 13.21 & 29.34 & 36 & 42.56 & 24 \\
GLM-4.6V-Flash & 9B & 41.93 & 56.01 & 35.87 & 44.60 & 19 & 43.81 & 43.36 & 56.79 & 47.99 & 16 & 48.67 & 34.92 & 56.06 & 46.55 & 19 & 28.71 & 57.50 & 15.09 & 33.77 & 28 & 43.23 & 22 \\
\midrule
\rowcolor{groupGray}\multicolumn{24}{l}{\textit{Streaming video MLLMs}} \\
Flash-VStream & 7B & 15.26 & 22.70 & 17.99 & 18.65 & 38 & 31.85 & 21.61 & 36.34 & 29.93 & 38 & 35.33 & 9.52 & 22.73 & 22.53 & 38 & 14.85 & 54.17 & 16.98 & 28.67 & 38 & 24.94 & 38 \\
StreamForest & 7B & 47.69 & 56.74 & 35.48 & 46.64 & 12 & 42.74 & 36.64 & 56.22 & 45.20 & 24 & 60.00 & 30.16 & 59.09 & 49.75 & 12 & 26.73 & 59.17 & 18.87 & 34.92 & 26 & 44.13 & 18 \\
StreamingVLM & 7B & 40.29 & 41.43 & 34.25 & 38.65 & 35 & 47.74 & 39.72 & 64.11 & 50.52 & 10 & 46.67 & 28.57 & 50.00 & 41.75 & 30 & 38.61 & 64.17 & 20.75 & 41.18 & 13 & 43.03 & 23 \\
\midrule
\rowcolor{groupGray}\multicolumn{24}{l}{\textit{Token-compression and memory-based methods}} \\
InfiniPot-V & 7B & 38.66 & 41.66 & 36.82 & 39.05 & 31 & 34.82 & 26.81 & 45.53 & 35.72 & 36 & 56.67 & 17.46 & 51.52 & 41.88 & 29 & 33.66 & 47.50 & --$^{*}$ & 40.58 & 15 & 39.31 & 32 \\
HERMES & 7B & 42.00 & 46.35 & 34.29 & 40.88 & 27 & 40.48 & 36.92 & 58.85 & 45.41 & 23 & 56.00 & 30.16 & 62.12 & 49.43 & 13 & 41.58 & 70.00 & 16.98 & 42.86 & 6 & 44.64 & 15 \\
FluxMem & 7B & 41.46 & 44.58 & 42.89 & 42.98 & 22 & 43.33 & 39.36 & 59.98 & 47.56 & 17 & 53.33 & 28.57 & 54.55 & 45.48 & 24 & 48.51 & 64.17 & 15.09 & 42.59 & 9 & 44.65 & 14 \\
StreamingTOM & 7B & 35.60 & 43.34 & 32.79 & 37.24 & 36 & 50.48 & 37.14 & 56.91 & 48.18 & 14 & 51.33 & 23.81 & 40.91 & 38.68 & 34 & 23.76 & 54.17 & 22.64 & 33.52 & 29 & 39.41 & 31 \\
\midrule
\rowcolor{groupGray}\multicolumn{24}{l}{\textit{Spatially fine-tuned MLLMs}} \\
Spatial-MLLM & 7B & 34.39 & 34.35 & 38.41 & 35.72 & 37 & 40.00 & 29.03 & 48.53 & 39.19 & 33 & 54.00 & 17.46 & 31.82 & 34.43 & 36 & 23.76 & 55.00 & 30.19 & 36.32 & 23 & 36.41 & 35 \\
Cambrian-S & 7B & 42.79 & 43.75 & 34.14 & 40.23 & 29 & 38.57 & 32.00 & 49.53 & 40.03 & 32 & 54.00 & 15.87 & 40.91 & 36.93 & 35 & 18.81 & 50.00 & 20.75 & 29.86 & 34 & 36.76 & 34 \\
Cambrian-S-LFP & 7B & 38.72 & 42.14 & 35.62 & 38.82 & 32 & 37.80 & 25.61 & 50.72 & 38.04 & 34 & 52.00 & 14.29 & 36.36 & 34.22 & 37 & 22.77 & 42.50 & 20.75 & 28.68 & 37 & 34.94 & 37 \\
VST-7B-SFT & 7B & 48.70 & 42.92 & 38.11 & 43.25 & 20 & 41.85 & 28.69 & 61.41 & 43.98 & 27 & 50.67 & 26.98 & 53.03 & 43.56 & 26 & 24.75 & 62.50 & 26.42 & 37.89 & 21 & 42.17 & 26 \\
VST-7B-RL & 7B & 51.07 & 44.82 & 41.10 & 45.66 & 14 & 44.23 & 29.03 & 59.35 & 44.20 & 25 & 52.00 & 20.63 & 50.00 & 40.88 & 32 & 26.73 & 60.83 & 26.42 & 37.99 & 20 & 42.18 & 25 \\
SenseNova-SI-1.5 & 8B & 43.99 & 44.67 & 37.74 & 42.13 & 25 & 45.83 & 27.69 & 53.66 & 42.40 & 31 & 54.00 & 31.75 & 42.42 & 42.72 & 28 & 21.78 & 42.50 & 33.96 & 32.75 & 30 & 40.00 & 29 \\
Spatial-TTT & 2B & 47.25 & 39.84 & 29.08 & 38.73 & 34 & 40.24 & 23.92 & 42.16 & 35.44 & 37 & 48.67 & 38.10 & 36.36 & 41.04 & 31 & 23.76 & 44.17 & 30.19 & 32.71 & 31 & 36.98 & 33 \\
\midrule
\rowcolor{groupGray}\multicolumn{24}{l}{\textit{Embodied foundation models}} \\
Cosmos-Reason1 & 7B & 48.20 & 43.85 & 42.31 & 44.79 & 18 & 36.85 & 35.94 & 58.29 & 43.69 & 28 & 55.33 & 20.63 & 60.61 & 45.52 & 23 & 30.69 & 50.00 & 15.09 & 31.93 & 33 & 41.48 & 28 \\
VeBrain & 7B & 43.88 & 45.36 & 38.98 & 42.74 & 24 & 40.24 & 32.25 & 60.04 & 44.18 & 26 & 54.00 & 25.40 & 59.09 & 46.16 & 20 & 42.57 & 66.67 & 13.21 & 40.82 & 14 & 43.47 & 21 \\
RynnBrain & 8B & 45.20 & 49.42 & 41.33 & 45.32 & 16 & 49.29 & 38.94 & 62.54 & 50.26 & 11 & 54.00 & 33.33 & 54.55 & 47.29 & 17 & 35.64 & 64.17 & 28.30 & 42.70 & 8 & 46.39 & 11 \\
RoboBrain2.5 & 4B & 36.70 & 45.60 & 38.04 & 40.12 & 30 & 42.92 & 31.64 & 55.72 & 43.43 & 29 & 52.00 & 30.16 & 62.12 & 48.09 & 16 & 22.77 & 54.17 & 30.19 & 35.71 & 24 & 41.84 & 27 \\
RoboBrain2.5-NV & 8B & 34.68 & 51.38 & 42.53 & 42.86 & 23 & 45.95 & 35.56 & 58.41 & 46.64 & 18 & 56.67 & 28.57 & 66.67 & 50.63 & 10 & 26.73 & 49.17 & 26.42 & 34.10 & 27 & 43.56 & 20 \\
\bottomrule
\end{tabular}
\end{adjustbox}
\vspace{-10pt}
\caption{\textbf{Main results on OVO-S-Bench} (\rev{prefix-only protocol}, multiple-choice accuracy). \rev{The streaming-architecture and token-compression families run in query-agnostic \emph{streaming} mode and are compared against \emph{Human (streaming)}; all other families run in \emph{prefix-access} mode and are compared against \emph{Human (prefix-access)} (\S\ref{sec:setup}).} \textbf{R} is rank; top-three level ranks are shaded gray and overall ranks green (darker is better); bold marks the best per column; baselines and controls are unranked. $^{*}$L4.3 uses image options unsupported by InfiniPot-V.}
\label{tab:overall_results}
\end{table*}

\subsection{Benchmark Results}
\label{sec:main_results}
Tab.~\ref{tab:overall_results} supports four observations about the current state of streaming spatial intelligence.

\noindent \textbf{Significant Gap with Human Performance.}
\rev{The strongest system, Gemini-3.1-Pro, reaches 59.2, trailing the human baseline collected under its own prefix-access mode (92.2) by 33.0 points; the best open-source model, Qwen3-VL-235B-A22B, reaches 53.6, a 38.6-point gap under the same pairing. Measured instead against the more constrained \emph{Human (streaming)} score of 86.6---the reference for the query-agnostic families---the two gaps are 27.4 and 33.0 points.} Random (31.3) and Text-Only (37.1) baselines fall below all general-backbone and proprietary systems, confirming that the human-model gap reflects genuine visual-streaming difficulty rather than language priors.

\noindent \textbf{Allocentric Spatial Mapping Bottleneck.}
L4 is the lowest-scoring level for 32 of 38 systems, with an average gap of 9.3\% between L1--L3 and L4 and gaps above 10\% even on the largest open-source backbones (Qwen3-VL-235B-A22B: 10.6, InternVL-3.5-241B-A28B: 13.8). The six exceptions all have L1 below 41 (Qwen2.5-VL-7B, Flash-VStream, and four others), so their flipped ordering reflects degraded current-view perception rather than competent allocentric spatial mapping. Among L1--L3 the ordering is unstable (33 of 38 systems peak at L2 or L3): although evidence demand grows monotonically from L1 to L4, accuracy does not, and drops sharply once an allocentric map must be abstracted from the entire explored region. 

\noindent \textbf{Narrow but Uneven Closed-Source Lead.}
The closed-source advantage is only 5.6 points overall (Gemini-3.1-Pro 59.2 vs.\ Qwen3-VL-235B-A22B 53.6), narrower than the 10+ point gap reported on recent video and multimodal benchmarks~\citep{fu2026video}. The gap is uneven across levels: closed-source widens its lead on the memory-heavy L2 (+5.9, above the +5.6 overall) but narrows on L4 (+4.1), and on L3 the best open-source backbone exceeds Gemini-3.1-Pro by 5.3 points (61.2 vs.\ 55.9). The closed-source group is itself heterogeneous: GPT-5.4 (50.9) and Grok-4.1-Fast (43.7) trail Gemini-3.1-Pro by 8 and 15 points, suggesting streaming spatial competence is not yet a generic capability of frontier multimodal systems.

\noindent \textbf{Specialized Methods Lag Behind Backbones.}
No streaming-architecture or spatially fine-tuned variant outperforms its comparable general backbone, and 13 of 15 lag behind their own base. Streaming MLLMs are tuned to compress memory for narrative QA, and spatial fine-tuning~\citep{wu2025spatialmllm,cai2025sensenovasi} is tuned on discrete-frame QA; neither regime exercises persistent spatial state over a continuous prefix, consistent with the L2--L4 losses we measure. We dissect this further in Section~\ref{sec:method_vs_base}.

\subsection{Analysis Experiments}
\label{sec:analysis}

\subsubsection{Effect of Thinking Mode}
\label{sec:thinking}
\begin{figure}[t]
\centering
\begin{minipage}[c]{0.46\linewidth}
\centering
\setlength{\tabcolsep}{2pt}
\renewcommand{\arraystretch}{0.85}
\definecolor{deltaPos}{HTML}{E2F0E8}
\definecolor{deltaNeg}{HTML}{F7E0DC}
\providecommand{\dpos}[1]{\cellcolor{deltaPos}#1}
\providecommand{\dneg}[1]{\cellcolor{deltaNeg}#1}
\resizebox{\linewidth}{!}{%
\small
\begin{tabular}{l|r|rrrr}
\toprule
\textbf{Model} & $\Delta$\textbf{Overall} & $\Delta$\textbf{L1} & $\Delta$\textbf{L2} & $\Delta$\textbf{L3} & $\Delta$\textbf{L4} \\
\midrule
GLM-4.6V-Flash             & \dneg{-0.9} & \dpos{+1.7} & \dneg{-1.1} & \dneg{-2.2} & \dneg{-1.9} \\
InternVL-3.5-8B            & \dpos{+0.3} & \dneg{-1.3} & \dpos{+6.0} & \dneg{-1.8} & \dneg{-1.7} \\
InternVL-3.5-38B           &  0.0        & \dneg{-7.7} & \dpos{+1.7} & \dpos{+3.2} & \dpos{+2.7} \\
Qwen3-VL-4B                & \dneg{-2.5} & \dneg{-1.8} & \dpos{+0.1} & \dneg{-4.1} & \dneg{-4.3} \\
Qwen3-VL-32B               & \dpos{+1.0} & \dneg{-0.8} & \dpos{+3.5} & \dpos{+2.3} & \dneg{-1.2} \\
Qwen3.5-4B                 & \dneg{-1.3} & \dneg{-4.9} & \dpos{+4.3} & \dneg{-6.1} & \dpos{+1.5} \\
Qwen3.5-9B                 & \dpos{+6.7} & \dpos{+3.1} & \dpos{+9.4} & \dpos{+4.3} & \dpos{+10.0} \\
Qwen3.5-27B                & \dpos{+4.3} & \dneg{-0.5} & \dpos{+6.3} & \dpos{+1.5} & \dpos{+10.0} \\
Grok-4.1-Fast              & \dpos{+3.2} & \dpos{+3.2} & \dpos{+5.2} & \dneg{-0.3} & \dpos{+4.5} \\
\bottomrule
\end{tabular}%
}
\captionof{table}{\textbf{Thinking-mode versus non-thinking-mode variants} on OVO-S-Bench. $\Delta$ is thinking minus non-thinking accuracy.}
\label{tab:thinking}
\end{minipage}\hfill
\begin{minipage}[c]{0.50\linewidth}
\centering
\includegraphics[width=0.8\linewidth]{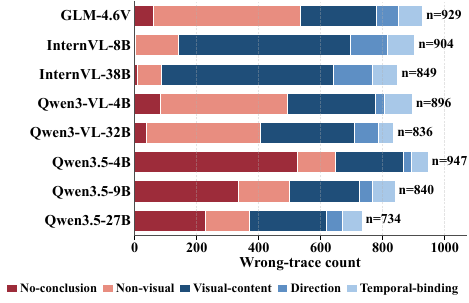}
\vspace{-12pt}
\captionof{figure}{\textbf{Wrong-trace counts (bar length) and error-class composition (segments)} across eight thinking-mode models.}
\label{fig:cot_failures}
\end{minipage}
\end{figure}

Tab.~\ref{tab:thinking} compares thinking-mode and non-thinking-mode variants for nine models, where the paired delta isolates the effect of explicit chain-of-thought.

\noindent \textbf{Model-Dependent Thinking Gains.}
Gains do not scale with backbone size: mid-size Qwen3.5 posts the strongest gain (from $-1.3$ at 4B to $+6.7$ at 9B), while Qwen3-VL and InternVL-3.5 hover near zero, Grok-4.1-Fast rises by $+3.2$, and GLM-4.6V-Flash falls by $-0.9$. Across levels, thinking consistently helps L2 (mean $\Delta = +3.9$, 8/9 pairs positive) and shows a small mean drop on L1 (mean $\Delta = -1.0$, 6/9 pairs negative), suggesting chain-of-thought aids the cross-frame integration but offers little for clean current-view perception.

\noindent \textbf{Probing Thinking-Mode Failure Patterns.}
We probe failure modes of thinking-mode models in Tab.~\ref{tab:thinking} by submitting every wrong-answer trace, together with the question, options, ground-truth, and predicted letters, and $4$--$6$ frames sampled at midpoints of the annotated evidence intervals, to GPT-5.4 as a judge. The judge assigns one of five labels: \textbf{no-conclusion error} (no extractable letter), \textbf{non-visual error} (commits via world priors without citing visible content), \textbf{visual-content error} (asserts a spatial relation the frames do not show), \textbf{direction error} (correct objects, inverted L/R or F/B), and \textbf{temporal-binding error} (correct object, wrong moment or visit). Three votes per trace are aggregated by majority. Please refer to Appendix~\ref{sec:appendix_cot} for full setup, cross-judge calibration, per-level breakdowns, and worked examples of each error class with annotated CoT traces. Across the eight thinking-mode models (Fig.~\ref{fig:cot_failures}), failures concentrate on mis-grounded visual evidence: non-visual and visual-content errors jointly cover $60$--$80\%$ of wrong traces in GLM-4.6V-Flash, Qwen3-VL, and InternVL-3.5. Qwen3.5-thinking is the sole exception, exhausting its 32K thinking-token budget without emitting an answer letter.

\begin{figure*}[t]
\centering
\includegraphics[width=0.9\linewidth]{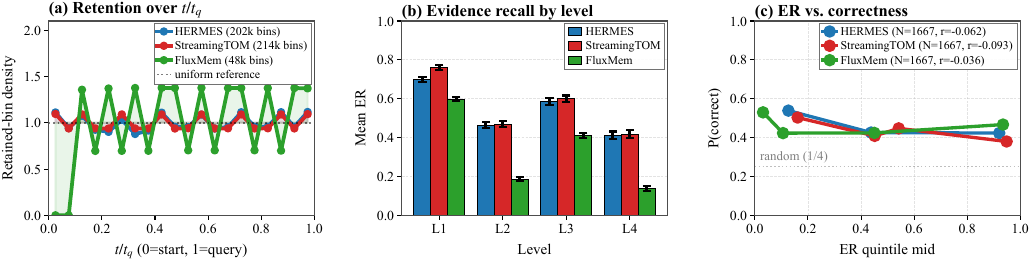}
\vspace{-12pt}
\caption{\textbf{Token-compression diagnostics.} \textbf{(a)} Retained 1-s bin density vs.\ $t/t_q$ (dotted: uniform). \textbf{(b)} Evidence Recall (ER) by level. \textbf{(c)} ER vs.\ $P(\text{correct})$; legend shows per-method Pearson $r$.}
\label{fig:compression}
\end{figure*}

\subsubsection{Frame Sampling Sensitivity}
\label{sec:frame_sampling}
We ask whether frame-selection strategy explains the gap to human performance. Eight policies cover four regimes: a one-frame query baseline (\emph{1@q}); a 16-frame sliding window at 4\, fps (\emph{16f@4fps}); uniform budgets \emph{U-32/64/128/256} over the entire prefix; a recency-weighted variant \emph{log-128} (60\% within 30\,s of the query, 30\% in 30\,s--5\,min, 10\% earlier), and an offline oracle \emph{oracle-evidence} that places 128 frames inside the evidence intervals. Formal definitions are in Appendix~\ref{sec:appendix_sampling}.

\begin{table}[t]
\centering
\small
\setlength{\tabcolsep}{3pt}
\renewcommand{\arraystretch}{0.88}
\resizebox{\columnwidth}{!}{%
\begin{tabular}{l r r r r r r}
\toprule
\textbf{Model} & \textbf{1@q} & \textbf{16f@4fps} & \textbf{U-128} & \textbf{U-256} & \textbf{log-128} & \textbf{oracle} \\
\midrule
Gemini-3.1-Flash-Lite & 47.95 & 49.72 & \textbf{50.88} & 48.33 & 48.55 & 49.70 \\
Qwen3-VL-32B          & 43.95 & 45.80 & 48.71 & 48.93 & \textbf{51.40} & 48.80 \\
Qwen3.5-27B           & 47.49 & 48.42 & 51.56 & 50.64 & 51.74 & \textbf{51.87} \\
VST-7B-RL             & 41.24 & 39.10 & 42.13 & 41.99 & \textbf{44.35} & 42.13 \\
RoboBrain2.5-NV-8B    & 39.91 & 41.22 & 43.56 & 43.87 & \textbf{46.84} & 43.56 \\
\bottomrule
\end{tabular}%
}
\vspace{-10pt}
\caption{\textbf{Frame-sampling sensitivity on a five-model representative subset} (Overall accuracy; bold per row). Policy definitions and the full policy / model matrix are in Appendix~\ref{sec:appendix_sampling}.}
\label{tab:frame_sampling}
\end{table}

\noindent \textbf{Sampling Strategies Help Little.}
Tab.~\ref{tab:frame_sampling} reports overall accuracy under six policies on a representative subset. Neither the oracle nor \emph{U-256} improves over \emph{U-128}: the oracle gain is at most $+0.3$ across all five models, and \emph{U-256} stays in $[-2.6, +0.3]$. The recency-weighted \emph{log-128} matches or exceeds the oracle on three of five models (Qwen3-VL-32B $+2.6$, VST-7B-RL $+2.2$, RoboBrain2.5-NV $+3.3$ vs.\ \emph{U-128}) and underperforms on the other two, so smarter sampling does not consistently close the gap. Short-context priors are level-selective: \emph{1@q} wins Qwen3.5-27B L1 by $+7.8$ but loses L4 by $-4.1$ versus uniform, replicating the perception--memory trade-off.
Together, these results indicate that the 27-point gap to human performance does \textbf{not} reduce to a retrieval problem solvable by better frame selection.

\subsubsection{Specialized Methods versus Backbones}
\label{sec:method_vs_base}
Streaming, token-compression, and spatially fine-tuned MLLMs target the regime OVO-S-Bench evaluates. Tab.~\ref{tab:method_vs_base} pairs each specialized method with its base backbone under the same protocol.

\begin{table}[t]
\centering
\small
\setlength{\tabcolsep}{2.5pt}
\renewcommand{\arraystretch}{0.85}
\definecolor{deltaPos}{HTML}{E2F0E8}
\definecolor{deltaNeg}{HTML}{F7E0DC}
\providecommand{\dpos}[1]{\cellcolor{deltaPos}#1}
\providecommand{\dneg}[1]{\cellcolor{deltaNeg}#1}
\resizebox{\columnwidth}{!}{%
\begin{tabular}{l l|r|rrrr}
\toprule
\textbf{Method} & \textbf{Base} & $\Delta$\textbf{Overall} & $\Delta$\textbf{L1} & $\Delta$\textbf{L2} & $\Delta$\textbf{L3} & $\Delta$\textbf{L4} \\
\midrule
\multicolumn{7}{l}{\textit{Spatially fine-tuned}} \\
Spatial-MLLM-7B     & Qwen2.5-VL-7B   & \dneg{-7.70}  & \dneg{-4.82}  & \dneg{-6.16}  & \dneg{-11.43} & \dneg{-8.39}  \\
VST-7B-SFT          & Qwen2.5-VL-7B   & \dneg{-1.98}  & \dpos{+2.57}  & \dneg{-1.36}  & \dneg{-2.30}  & \dneg{-6.82}  \\
VST-7B-RL           & Qwen2.5-VL-7B   & \dneg{-1.98}  & \dpos{+4.92}  & \dneg{-1.15}  & \dneg{-4.98}  & \dneg{-6.71}  \\
Spatial-TTT-2B      & Qwen3-VL-2B     & \dneg{-3.54}  & \dneg{-1.81}  & \dneg{-10.12} & \dneg{-4.28}  & \dpos{+2.04}  \\
SenseNova-SI-1.5    & InternVL-3.5-8B & \dneg{-4.55}  & \dneg{-3.72}  & \dneg{-3.43}  & \dneg{-4.47}  & \dneg{-6.57}  \\
\midrule
\multicolumn{7}{l}{\textit{Streaming architectures}} \\
StreamingVLM-7B     & Qwen2.5-VL-7B   & \dneg{-1.14}  & \dneg{-2.09}  & \dpos{+5.17}  & \dneg{-4.11}  & \dneg{-3.53}  \\
Flash-VStream-7B    & Qwen2-VL-7B     & \dneg{-18.35} & \dneg{-21.69} & \dneg{-14.68} & \dneg{-20.28} & \dneg{-16.73} \\
\midrule
\multicolumn{7}{l}{\textit{Token-compression and memory}} \\
FluxMem             & Qwen2.5-VL-7B   & \dpos{+0.45}  & \dpos{+2.23}  & \dpos{+2.05}  & \dneg{-0.37}  & \dneg{-2.11}  \\
HERMES              & Qwen2.5-VL-7B   & \dpos{+0.44}  & \dpos{+0.14}  & \dneg{-0.09}  & \dpos{+3.57}  & \dneg{-1.85}  \\
StreamingTOM        & Qwen2.5-VL-7B   & \dneg{-4.78}  & \dneg{-3.43}  & \dpos{+2.67}  & \dneg{-7.18}  & \dneg{-11.18} \\
\midrule
\multicolumn{7}{l}{\textit{Embodied foundation models}} \\
Cosmos-Reason1-7B   & Qwen2.5-VL-7B   & \dneg{-2.63}  & \dpos{+4.25}  & \dneg{-1.66}  & \dneg{-0.33}  & \dneg{-12.78} \\
VeBrain-7B          & Qwen2.5-VL-7B   & \dneg{-0.64}  & \dpos{+2.21}  & \dneg{-1.17}  & \dpos{+0.31}  & \dneg{-3.89}  \\
RoboBrain2.5-4B     & Qwen3-VL-4B     & \dneg{-4.90}  & \dneg{-2.93}  & \dneg{-4.60}  & \dneg{-6.44}  & \dneg{-5.64}  \\
RynnBrain-8B        & Qwen3-VL-8B     & \dneg{-1.24}  & \dneg{-0.05}  & \dneg{-0.28}  & \dneg{-4.95}  & \dpos{+0.33}  \\
RoboBrain2.5-8B-NV  & Qwen3-VL-8B     & \dneg{-4.07}  & \dneg{-2.51}  & \dneg{-3.89}  & \dneg{-1.61}  & \dneg{-8.27}  \\
\bottomrule
\end{tabular}%
}
\vspace{-10pt}
\caption{\textbf{Specialized methods versus their base backbones on OVO-S-Bench}. $\Delta$ is specialized minus base accuracy.}
\label{tab:method_vs_base}
\end{table}

\noindent \textbf{Specialization Loses Across Levels.}
\textbf{Specialization is net-negative overall:} 13 of 15 methods lose on overall accuracy (median $-2.0$, range $-18.4$ to $+0.5$), and only HERMES ($+0.4$) and FluxMem ($+0.5$) exceed their base. \textbf{L4 is the most uniformly damaged level:} 13 of 15 methods regress on allocentric spatial mapping (mean $\Delta = -6.1$, only Spatial-TTT $+2.0$ and RynnBrain $+0.3$ are positive), and Flash-VStream-7B ($-16.7$) and Cosmos-Reason1-7B ($-12.8$) show the largest drops. \textbf{L1 is occasionally improved:} perception-style specialization lifts current-view scores (VST-7B-RL $+4.9$, Cosmos-Reason1 $+4.3$, VST-7B-SFT $+2.6$). The L1/L4 split is consistent with regimes optimized for fragmented spatial QA, event memory, or robotic action grounding rather than long-range geometric evidence. This pattern spans all three families despite non-overlapping training data, making a pure domain-shift explanation unlikely (though controlled fine-tuning would be needed to confirm).

\noindent \textbf{Retention Is Not the Bottleneck.}
For HERMES, StreamingTOM, and FluxMem, we instrument inference to dump, at every query timestamp, the original-video seconds surviving in memory, then compute the per-query \emph{Compression Ratio} (CR; fraction of the prefix kept) and \emph{Evidence Recall} (ER; fraction of the annotated evidence interval that survives). \textbf{Compression sheds evidence on memory-heavy levels:} mean ER on L4 ranges from $0.14$ (FluxMem) to $0.42$ (StreamingTOM), against $0.60$--$0.76$ on L1 (Figure~\ref{fig:compression}b), so the L2/L4 cost of compression originates before reasoning begins. \textbf{Yet retention alone does not predict correctness:} the per-query Pearson correlation between ER and correctness is essentially zero across all three methods ($r \in [-0.07, 0.00]$, panel c), and the retained-frame time distribution stays close to the uniform reference (panel a) rather than concentrating on the evidence interval. Consistent with Sec.~\ref{sec:frame_sampling}, higher retention does not translate into correctness under these compressors.

\section{Conclusion}
We introduce OVO-S-Bench, a fully human-annotated benchmark that exposes the streaming spatial intelligence gap of current video MLLMs. Its 1{,}680 questions span four levels of spatial abstraction under a strict prefix-only protocol with annotated query timestamps and evidence intervals. Across 38 evaluated systems, Gemini-3.1-Pro (59.2) \rev{trails its matched prefix-access human baseline by 33 points and the more constrained human streaming baseline by 27}, with allocentric spatial mapping as the dominant bottleneck; streaming and spatially fine-tuned variants score below their own backbones, and chain-of-thought amplifies spatial errors when ungrounded in the stream. We hope OVO-S-Bench helps shift spatial-intelligence evaluation from offline, discrete-frame benchmarks to continuous, streaming ones, where models must perceive, remember, and reason about space as it unfolds in real time, bringing evaluation closer to the embodied setting in which spatial intelligence must operate.

\section*{Limitations}
While OVO-S-Bench covers the broadest mix of streaming-spatial scenes to date, from indoor walkthroughs to driving videos, the evaluated model remains a passive observer of pre-recorded streams and cannot close the perception-action loop that real embodied agents rely on. \rev{Each item is also scored via multiple-choice under a fixed prefix-only constraint. A 200-item open-ended pilot measures what that format costs: strict free-form accuracy falls 10.5 points below multiple choice on the same items, and 20\% of free-form answers are partially correct, a band a single letter cannot express (Appendix~\ref{sec:appendix_open_ended}). The format therefore trades item-level resolution for deterministic scoring.} The specialization analysis (Section~\ref{sec:method_vs_base}) does not control for domain shift between each method's training data and OVO-S-Bench; controlled fine-tuning would be needed to fully isolate architectural limitations. This reflects a current trade-off in the field, as prior streaming-spatial benchmarks recover the loop only by trading scene diversity for controlled 3D environments, via camera-pose graphs over indoor scans \citep{ost2025bench} or a full physics simulator \citep{hong2026esibench}. Moving forward, we plan to extend the protocol to a closed-loop regime that preserves the same diversity, toward benchmarks that evaluate not only what models perceive but also what they choose to perceive.

{\revon   
\section*{Ethics Statement}

\paragraph{Intended use.}
The capability OVO-S-Bench measures matters most where spatial evidence outside the current view is safety-critical: assistive AR navigation and aids for blind and low-vision users must recall what lies behind the user or what was passed several minutes earlier. Measuring this under a streaming constraint, rather than over a fully observed video, is what makes the measurement informative for those settings.

\paragraph{Risks.}
Spatially unreliable models deployed in embodied or assistive systems can cause physical harm, and the failure modes documented here---ungrounded spatial claims and evidence discarded before reasoning begins---are the ones an offline, fully observed evaluation hides. We release the benchmark so that such failures surface before deployment rather than after.

\paragraph{Data provenance and licensing.}
All source videos come from publicly released datasets or public walking-tour footage and are used under their original terms; we do not relicense or redistribute video (Appendix~\ref{subsec:source_datasets} lists the per-source terms). Our release contains human-written annotations only---questions, options, answers, query timestamps, evidence intervals, and per-source metadata---together with the source identifiers and access instructions needed to obtain the underlying videos from their original providers. The evaluation harness, prompt templates, and scoring code are available at \url{https://github.com/InternLM/OVO-S-Bench} and the annotations at \url{https://huggingface.co/datasets/JoeLeelyf/OVO-S-Bench}. No personally identifiable information was collected, and items describe scene structure rather than identifiable individuals.

\paragraph{Annotation.}
The 12 annotators were recruited from the authors' institutions and annotated as part of their research duties; recruitment, training, adjudication, and compensation arrangements are documented in Appendix~\ref{subsec:annotation_overview}.
}       

\section*{Acknowledgments}
This work was supported by Shanghai Artificial Intelligence Laboratory.

\bibliographystyle{plainnat}
\bibliography{custom}

\newpage
\appendix
\clearpage
\section*{\Large Appendix}
\renewcommand{\thesection}{\Alph{section}}
\setcounter{section}{0}

\textbf{Content of Appendices}\\
\textbf{Section}~\ref{sec:appendix_scaling}. Per-Family Scaling Curves.\\
\hspace*{1.5em}$\bullet$ Per-level accuracy versus backbone scale for Qwen3-VL, Qwen3.5, and InternVL-3.5.\\
\textbf{Section}~\ref{sec:appendix_sampling}. Frame-Sampling Sensitivity.\\
\hspace*{1.5em}$\bullet$ \textbf{\S}~\ref{subsec:sampling_defs}. Sampling Policy Definitions.\\
\hspace*{1.5em}$\bullet$ \textbf{\S}~\ref{subsec:sampling_overall}. Overall-Accuracy Matrix.\\
\hspace*{1.5em}$\bullet$ \textbf{\S}~\ref{subsec:sampling_perlevel}. Per-Level Breakdown.\\
\hspace*{1.5em}$\bullet$ \textbf{\S}~\ref{subsec:sampling_analysis}. Additional Analysis.\\
\hspace*{1.5em}$\bullet$ \textbf{\S}~\ref{subsec:length_difficulty}. Length versus Difficulty.\\
\textbf{Section}~\ref{sec:appendix_cot}. CoT Failure Taxonomy.\\
\hspace*{1.5em}$\bullet$ \textbf{\S}~\ref{subsec:cot_setup}. Judge Setup and Pre-filtering.\\
\hspace*{1.5em}$\bullet$ \textbf{\S}~\ref{subsec:cot_calibration}. Cross-Judge Calibration.\\
\hspace*{1.5em}$\bullet$ \textbf{\S}~\ref{subsec:cot_perlevel}. Per-Level Failure Distribution.\\
\hspace*{1.5em}$\bullet$ \textbf{\S}~\ref{subsec:cot_examples}. Worked Examples.\\
\textbf{Section}~\ref{sec:appendix_categories}. Detailed Category Definitions.\\
\hspace*{1.5em}$\bullet$ L1--L4 subcategory definitions with accepted/rejected examples.\\
\textbf{Section}~\ref{sec:appendix_protocols}. Annotation Protocol \& Construction Pipeline.\\
\hspace*{1.5em}$\bullet$ \textbf{\S}~\ref{subsec:source_datasets}. Source Datasets and Licensing.\\
\hspace*{1.5em}$\bullet$ \textbf{\S}~\ref{subsec:general_protocol}. General Annotation Protocol.\\
\hspace*{1.5em}$\bullet$ \textbf{\S}~\ref{subsec:construction_techniques}. Level-Specific Construction Techniques.\\
\hspace*{1.5em}$\bullet$ \textbf{\S}~\ref{subsec:qc_pipeline}. Quality Control Pipeline.\\
\textbf{Section}~\ref{sec:appendix_related}. Extended Comparison with Prior Benchmarks.\\
\hspace*{1.5em}$\bullet$ Per-benchmark discussion of what each benchmark does, shared points with OVO-S-Bench, and key differences.\\
\hspace*{1.5em}$\bullet$ \textbf{\S}~\ref{subsec:text_only_shortcut}. Text-Only Shortcut Analysis.\\
\hspace*{1.5em}$\bullet$ \textbf{\S}~\ref{subsec:source_robustness}. Source-Robustness Analysis.\\
\textbf{Section}~\ref{sec:appendix_eval}. Evaluation Methods Details.\\
\hspace*{1.5em}$\bullet$ \textbf{\S}~\ref{subsec:eval_models}. Models Evaluated.\\
\hspace*{1.5em}$\bullet$ \textbf{\S}~\ref{subsec:eval_decoding}. Decoding Parameters.\\
\hspace*{1.5em}$\bullet$ \textbf{\S}~\ref{subsec:eval_sampling}. Default Frame Sampling.\\
\hspace*{1.5em}$\bullet$ \textbf{\S}~\ref{subsec:eval_streaming}. Streaming-Model Ingestion Protocol.\\
\hspace*{1.5em}$\bullet$ \textbf{\S}~\ref{subsec:eval_prompts}. Prompt Templates.\\
\hspace*{1.5em}$\bullet$ \textbf{\S}~\ref{subsec:eval_policy_prompts}. Policy-Specific Inputs.\\
\hspace*{1.5em}$\bullet$ \textbf{\S}~\ref{subsec:eval_parsing}. Answer Extraction.\\
\hspace*{1.5em}$\bullet$ \textbf{\S}~\ref{subsec:eval_runtime}. Hardware and Software.\\
\textbf{Section}~\ref{sec:future_work}. Future Work.\\
\textbf{Section}~\ref{sec:appendix_examples}. Additional Examples.\\

\section{Per-Family Scaling Curves}
\label{sec:appendix_scaling}
This appendix accompanies the corollary in Section~\ref{sec:main_results} that L4 saturates early. We trace per-level accuracy as a function of backbone scale within three open-source families: Qwen3-VL (2B/4B/8B/30B-A3B/32B/235B-A22B), Qwen3.5 (0.8B/2B/4B/9B/27B/35B-A3B/122B-A10B/397B-A17B), and InternVL-3.5 (1B/2B/4B/8B/30B-A3B/38B/241B-A28B), all under default 128-frame uniform sampling with reasoning off, so the curves isolate scale. MoE checkpoints are marked with open rings, since the total parameter count overstates per-token compute.

\begin{figure*}[t]
\centering
\includegraphics[width=\textwidth]{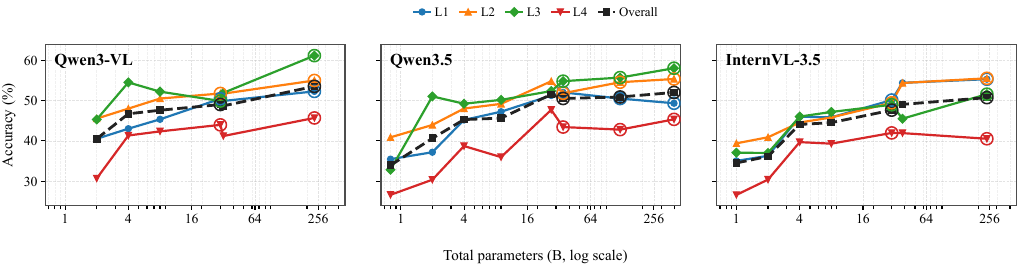}
\caption{Per-level accuracy versus total parameter count (log scale) within three open-source families. Solid lines: L1--L4. Dashed line with square markers: overall accuracy. Open rings overlay MoE checkpoints.}
\label{fig:scaling}
\end{figure*}

\paragraph{Three Scaling Regimes.}
Figure~\ref{fig:scaling} reveals three regimes. \textbf{L1 and L2 scale log-linearly:} end-to-end gains are $+11.8 / +9.5$ on Qwen3-VL, $+13.9 / +14.5$ on Qwen3.5, and $+20.3 / +16.2$ on InternVL-3.5, with the two curves tracking each other within 3 points at every size. \textbf{L3 climbs but unevenly:} Qwen3.5 gains $+25.2$ overall yet Qwen3-VL drops from 54.5 at 4B to 49.9 at 30B-A3B before recovering to 61.2 at 235B-A22B, and the InternVL-3.5 38B dense checkpoint (45.5) underperforms its 30B-A3B MoE sibling (49.0), suggesting L3 depends on training-data composition as much as raw parameter count. \textbf{L4 saturates early:} all three families plateau in the 40--46 range from the mid-tier onward; InternVL-3.5 L4 moves only $+0.8$ over a 60$\times$ parameter increase (39.7 at 4B vs.\ 40.5 at 241B-A28B), and remains roughly 39 points below human-streaming (79.2), indicating the allocentric bottleneck is not a capacity problem and will not close from scaling alone.

\section{Frame-Sampling Sensitivity}
\label{sec:appendix_sampling}
This appendix accompanies Section~\ref{sec:frame_sampling}. We formalize the eight sampling policies, then report the full eight-model $\times$ eight-policy accuracy matrix at the Overall and per-level granularity. Three observations not included in the main text follow.

\subsection{Sampling Policy Definitions}
\label{subsec:sampling_defs}
Let the video frame rate be $f$, the query time $t_q$ seconds, and the query frame index $n_q = \lfloor t_q \cdot f \rfloor$. Let $N$ be the total frame budget and $\mathcal{E} = \{[s_i, e_i]\}_{i=1}^{K}$ the set of annotated evidence intervals (in seconds). Every policy outputs an ordered frame index set $\mathcal{T} \subseteq \{0, 1, \dots, n_q\}$, i.e.\ never consumes content posterior to the query, satisfying the streaming protocol.

\paragraph{single@query.}
The naive streaming baseline that uses only the query frame: $\mathcal{T}_{\text{sgl}} = \{\, n_q \,\}$ ($N{=}1$).

\paragraph{nearest-16f@4fps.}
A short causal sliding window with step $\Delta = \max(1, \text{round}(f / f_w))$:
\begin{equation*}
\mathcal{T}_{\text{rcw}} = \{\, n_q - k\Delta : 0 \le k < N,\; n_q - k\Delta \ge 0 \,\}.
\end{equation*}
The default $N{=}16$, $f_w{=}4\,\text{fps}$ covers the last $(N-1)/f_w = 3.75$\,s before the query.

\paragraph{uniform-$N$.}
Uniform sampling of $N$ frames over the entire prefix $[0, n_q]$:
\begin{equation*}
\mathcal{T}_{\text{uni}} = \bigl\{\, \text{round}\bigl(\tfrac{k}{N-1}\, n_q\bigr) : 0 \le k < N \,\bigr\}.
\end{equation*}
We sweep $N \in \{32, 64, 128, 256\}$; \emph{uniform-128} is the default policy used in Table~\ref{tab:overall_results}.

\paragraph{oracle-evidence.}
An offline oracle that allocates the $N{=}128$ budget inside the annotated evidence intervals only. Each interval $[s_i, e_i]$ maps to a frame-domain interval $[\tilde s_i, \tilde e_i]$ clipped to $\le n_q$, with length $L_i = \tilde e_i - \tilde s_i$ and per-interval budget $N_i = \max(1, \text{round}(N L_i / \sum_j L_j))$ (rounding corrected on the longest interval). Each interval is then uniformly sampled:
\begin{equation*}
\mathcal{T}_{\text{evid}} = \bigcup_{i=1}^{K} \text{linspace}(\tilde s_i, \tilde e_i, N_i).
\end{equation*}
When $\mathcal{E} = \emptyset$ the policy degrades to \emph{uniform-$N$} over $[0, n_q]$. This serves as an upper bound: \emph{how well a model could do if it had access only to the truly relevant frames.}

\paragraph{log-decay-128.}
A recency-weighted three-band schedule with $N{=}128$, band boundaries $\beta_1 = 30\,\text{s},\, \beta_2 = 300\,\text{s}$, and weights $(\alpha_1, \alpha_2, \alpha_3) = (0.60, 0.30, 0.10)$:
\begin{equation*}
\begin{aligned}
\mathcal{B}_1 &= [n_q - \beta_1 f,\; n_q],            & N_1 &= \text{round}(\alpha_1 N), \\
\mathcal{B}_2 &= [n_q - \beta_2 f,\; n_q - \beta_1 f], & N_2 &= \text{round}(\alpha_2 N), \\
\mathcal{B}_3 &= [0,\; n_q - \beta_2 f],              & N_3 &= N - N_1 - N_2.
\end{aligned}
\end{equation*}
Each band is uniformly sampled. When the prefix is shorter than $\beta_2 f$ the missing band is skipped and its budget rolls into the surviving bands. This policy realizes a coarse exponential prior of near-importance without using ground-truth evidence.

Table~\ref{tab:sampling_summary} compares the five families at $N{=}128$.

\begin{table}[t]
\centering
\small
\setlength{\tabcolsep}{4pt}
\renewcommand{\arraystretch}{1.0}
\resizebox{\columnwidth}{!}{%
\begin{tabular}{l l l}
\toprule
\textbf{Policy} & \textbf{Prefix coverage} & \textbf{Type} \\
\midrule
single@query ($N{=}1$)      & last frame only                          & streaming \\
nearest-16f@4fps ($N{=}16$) & last 3.75\,s at 4\,fps                   & streaming \\
uniform-128                 & uniform over $[0, n_q]$                  & streaming \\
log-decay-128               & 3-band recency-weighted (60/30/10\%)     & streaming \\
oracle-evidence             & only inside $\mathcal{E}$                & offline \\
\bottomrule
\end{tabular}%
}
\caption{Frame-sampling policies, $N{=}128$ unless noted. Streaming policies obey the prefix-only constraint; the oracle requires evidence annotations and is not available at deployment.}
\label{tab:sampling_summary}
\end{table}

\subsection{Overall-Accuracy Matrix}
\label{subsec:sampling_overall}
Table~\ref{tab:sampling_overall_full} gives the Overall accuracy of eight models under all eight policies. Empty cells (``--'') indicate that the model's context window cannot accommodate the required frame count.

\begin{table*}[t]
\centering
\small
\setlength{\tabcolsep}{4pt}
\renewcommand{\arraystretch}{0.95}
\begin{adjustbox}{width=\textwidth}
\begin{tabular}{l r r r r r r r r}
\toprule
\textbf{Model} & \textbf{single@q} & \textbf{nearest-16f} & \textbf{U-32} & \textbf{U-64} & \textbf{U-128} & \textbf{U-256} & \textbf{oracle-E} & \textbf{log-128} \\
\midrule
Gemini-3.1-Flash-Lite & 47.95 & 49.72 & \textbf{52.45} & 51.26 & 50.88 & 48.33 & 49.70 & 48.55 \\
InternVL-3.5-38B      & 45.04 & 44.45 & 48.94 & 49.49 & 49.08 & --    & \textbf{50.18} & 49.33 \\
Qwen3-VL-32B          & 43.95 & 45.80 & 45.89 & 47.18 & 48.71 & 48.93 & 48.80 & \textbf{51.40} \\
Qwen3.5-27B           & 47.49 & 48.42 & 48.12 & 49.72 & 51.56 & 50.64 & \textbf{51.87} & 51.74 \\
VST-7B-RL             & 41.24 & 39.10 & 41.50 & 43.62 & 42.13 & 41.99 & 42.13 & \textbf{44.35} \\
Cambrian-S-7B         & 35.93 & 36.01 & 34.93 & 36.66 & 36.76 & 36.20 & 37.17 & \textbf{37.67} \\
VeBrain-7B            & 43.51 & 45.17 & 41.65 & 43.82 & 43.47 & 43.30 & 43.53 & \textbf{45.40} \\
RoboBrain2.5-NV-8B    & 39.91 & 41.22 & 40.82 & 42.63 & 43.56 & 43.87 & 43.56 & \textbf{46.84} \\
\bottomrule
\end{tabular}
\end{adjustbox}
\caption{Full Overall-accuracy matrix on OVO-S-Bench (\%). Eight representative models $\times$ eight frame-sampling policies. Per-row best is bolded. ``--'' = context-window overflow.}
\label{tab:sampling_overall_full}
\end{table*}

\subsection{Per-Level Breakdown}
\label{subsec:sampling_perlevel}
Table~\ref{tab:sampling_perlevel} reports the per-level accuracies (L1--L4) of the same eight models under the same eight policies. Per-row best is bolded. The level breakdown drives the three observations in \S\ref{subsec:sampling_analysis}.

\begin{table*}[t]
\centering
\setlength{\tabcolsep}{2pt}
\renewcommand{\arraystretch}{0.90}
\begin{minipage}[t]{0.49\textwidth}
\centering
{\small\textbf{(a) L1: Instantaneous Egocentric Perception}}\\[2pt]
\resizebox{\linewidth}{!}{%
\begin{tabular}{l rrrrrrrr}
\toprule
\textbf{Model} & \textbf{1@q} & \textbf{16f} & \textbf{U32} & \textbf{U64} & \textbf{U128} & \textbf{U256} & \textbf{Ora} & \textbf{Log} \\
\midrule
Gemini-3.1-Flash-Lite & \textbf{59.53} & 59.46 & 53.12 & 53.14 & 53.88 & 53.40 & 53.28 & 53.60 \\
InternVL-3.5-38B      & 59.87 & \textbf{61.15} & 53.35 & 54.04 & 54.44 & --    & 54.97 & 58.48 \\
Qwen3-VL-32B          & 56.46 & \textbf{60.74} & 49.36 & 49.66 & 49.93 & 51.37 & 50.19 & 55.34 \\
Qwen3.5-27B           & 59.12 & \textbf{61.88} & 49.52 & 50.87 & 51.29 & 50.63 & 51.06 & 53.87 \\
VST-7B-RL             & \textbf{57.81} & 52.45 & 45.47 & 45.96 & 45.45 & 46.16 & 45.45 & 50.73 \\
Cambrian-S-7B         & 39.14 & 41.12 & 38.25 & 40.16 & 40.23 & 39.81 & 40.23 & \textbf{42.21} \\
VeBrain-7B            & 52.64 & \textbf{54.43} & 41.67 & 42.08 & 42.74 & 43.02 & 42.74 & 44.52 \\
RoboBrain2.5-NV-8B    & \textbf{53.79} & 53.25 & 41.14 & 43.42 & 42.86 & 44.06 & 42.86 & 50.14 \\
\bottomrule
\end{tabular}%
}
\end{minipage}
\hfill
\begin{minipage}[t]{0.49\textwidth}
\centering
{\small\textbf{(b) L2: Context Tracking}}\\[2pt]
\resizebox{\linewidth}{!}{%
\begin{tabular}{l rrrrrrrr}
\toprule
\textbf{Model} & \textbf{1@q} & \textbf{16f} & \textbf{U32} & \textbf{U64} & \textbf{U128} & \textbf{U256} & \textbf{Ora} & \textbf{Log} \\
\midrule
Gemini-3.1-Flash-Lite & 36.90 & 39.25 & \textbf{56.88} & 54.11 & 52.03 & 47.96 & 49.03 & 42.41 \\
InternVL-3.5-38B      & 33.32 & 34.89 & 52.67 & \textbf{55.03} & 54.40 & --    & 54.41 & 52.98 \\
Qwen3-VL-32B          & 33.93 & 34.55 & 50.13 & 52.27 & 51.75 & \textbf{53.53} & 52.31 & 50.45 \\
Qwen3.5-27B           & 35.82 & 35.71 & 48.89 & 52.35 & 54.83 & 54.09 & \textbf{55.16} & 52.46 \\
VST-7B-RL             & 31.63 & 28.71 & 44.17 & 44.86 & 44.20 & 43.65 & 44.20 & \textbf{45.22} \\
Cambrian-S-7B         & 34.56 & 34.29 & 36.79 & 38.56 & 40.03 & 39.75 & 39.99 & \textbf{42.18} \\
VeBrain-7B            & 31.82 & 33.21 & 42.45 & 43.89 & 44.18 & 46.00 & 44.18 & \textbf{47.73} \\
RoboBrain2.5-NV-8B    & 28.45 & 31.10 & 42.30 & 44.12 & \textbf{46.64} & 46.31 & 46.64 & 46.64 \\
\bottomrule
\end{tabular}%
}
\end{minipage}

\vspace{6pt}

\begin{minipage}[t]{0.49\textwidth}
\centering
{\small\textbf{(c) L3: Gen.\ Spatial Reas.}}\\[2pt]
\resizebox{\linewidth}{!}{%
\begin{tabular}{l rrrrrrrr}
\toprule
\textbf{Model} & \textbf{1@q} & \textbf{16f} & \textbf{U32} & \textbf{U64} & \textbf{U128} & \textbf{U256} & \textbf{Ora} & \textbf{Log} \\
\midrule
Gemini-3.1-Flash-Lite & 50.85 & 53.12 & 54.56 & \textbf{55.18} & 54.14 & 51.32 & 54.29 & 53.65 \\
InternVL-3.5-38B      & 46.06 & 44.70 & \textbf{51.07} & 48.94 & 45.53 & --    & 46.01 & 47.35 \\
Qwen3-VL-32B          & 52.25 & 50.25 & 50.13 & 48.42 & 51.94 & \textbf{55.03} & 52.45 & 54.02 \\
Qwen3.5-27B           & 51.40 & 50.54 & 50.98 & 50.86 & 52.39 & 51.80 & 53.85 & \textbf{56.04} \\
VST-7B-RL             & 38.60 & 40.56 & 39.00 & \textbf{42.11} & 40.88 & 41.07 & 40.88 & 40.87 \\
Cambrian-S-7B         & \textbf{41.36} & 38.64 & 36.32 & 38.47 & 36.93 & 36.77 & 37.01 & 38.47 \\
VeBrain-7B            & 46.01 & 45.87 & 42.23 & \textbf{47.46} & 46.16 & 44.88 & 46.67 & 43.78 \\
RoboBrain2.5-NV-8B    & 43.65 & 44.28 & 42.81 & 46.66 & 50.63 & \textbf{51.08} & 50.63 & 49.22 \\
\bottomrule
\end{tabular}%
}
\end{minipage}
\hfill
\begin{minipage}[t]{0.49\textwidth}
\centering
{\small\textbf{(d) L4: Alloc.\ Spatial Map.}}\\[2pt]
\resizebox{\linewidth}{!}{%
\begin{tabular}{l rrrrrrrr}
\toprule
\textbf{Model} & \textbf{1@q} & \textbf{16f} & \textbf{U32} & \textbf{U64} & \textbf{U128} & \textbf{U256} & \textbf{Ora} & \textbf{Log} \\
\midrule
Gemini-3.1-Flash-Lite & 44.51 & \textbf{47.03} & 45.27 & 42.61 & 43.46 & 40.66 & 42.17 & 44.55 \\
InternVL-3.5-38B      & 40.90 & 37.07 & 38.68 & 39.94 & 41.94 & --    & \textbf{45.32} & 38.50 \\
Qwen3-VL-32B          & 33.15 & 37.65 & 33.95 & 38.35 & 41.22 & 35.77 & 40.26 & \textbf{45.80} \\
Qwen3.5-27B           & 43.61 & 45.54 & 43.10 & 44.79 & \textbf{47.72} & 46.03 & 47.42 & 44.57 \\
VST-7B-RL             & 36.94 & 34.68 & 37.34 & \textbf{41.54} & 37.99 & 37.09 & 37.99 & 40.56 \\
Cambrian-S-7B         & 28.68 & 29.98 & 28.37 & 29.45 & 29.86 & 28.49 & \textbf{31.44} & 27.84 \\
VeBrain-7B            & 43.58 & \textbf{47.16} & 40.25 & 41.84 & 40.82 & 39.29 & 40.54 & 45.59 \\
RoboBrain2.5-NV-8B    & 33.74 & 36.24 & 37.01 & 36.32 & 34.10 & 34.02 & 34.10 & \textbf{41.36} \\
\bottomrule
\end{tabular}%
}
\end{minipage}
\caption{Per-level accuracy (\%) under eight frame-sampling policies. Per-row best is bolded. ``--'' = context-window overflow.}
\label{tab:sampling_perlevel}
\end{table*}

\subsection{Additional Analysis}
\label{subsec:sampling_analysis}
\paragraph{The oracle is not an upper bound.}
On five of eight models (Cambrian-S-7B, VeBrain-7B, VST-7B-RL, Qwen3-VL-32B, RoboBrain2.5-NV-8B), \emph{log-128} outperforms \emph{oracle-evidence} on Overall by 0.5--3.3 points. The oracle places all 128 frames inside the annotated evidence interval, whereas the rest of the prefix is discarded; for questions whose answer also depends on global context (e.g.\ L3 counterfactuals and L4 trajectory queries), this aggressive concentration removes useful framing. The \emph{log-128} schedule, in contrast, retains 10\% of the budget on the far past, which appears to compensate.

\paragraph{Backbone-dependent value of long context.}
Doubling the uniform budget from 128 to 256 frames yields negligible or negative deltas for all seven models with data (range $-2.6$ to $+0.3$): Gemini-3.1-Flash-Lite $-2.6$, Qwen3.5 $-0.9$, Cambrian-S $-0.6$, VeBrain $-0.2$, VST-7B-RL $-0.1$, Qwen3-VL $+0.2$, RoboBrain2.5-NV $+0.3$. No model benefits by more than 0.3 points, consistent with the scaling-curve finding (\S\ref{sec:appendix_scaling}) that L4 saturates early across families and with the observation of~\citet{shen2026simplestream} that the marginal value of longer visual context is model-dependent.

\paragraph{Short-context priors and the perception--memory trade-off.}
Across all four large general backbones the one-frame query (\emph{1@q}) wins L1 by 5--8 points over \emph{uniform-128} (Gemini $+5.7$, InternVL-3.5-38B $+5.4$, Qwen3-VL-32B $+6.5$, Qwen3.5-27B $+7.8$) but loses L2 by 15--21 points (Gemini $-15.1$, Qwen3-VL $-17.8$, Qwen3.5 $-19.0$, InternVL-3.5 $-21.1$). The trade-off is sharpest for VST-7B-RL ($+12.4$ on L1 vs.\ $-12.6$ on L2) and is the empirical signature of the perception--memory trade-off~\citep{shen2026simplestream} for OVO-S-Bench: a clean current view helps egocentric perception but cannot answer questions that require integrating across the prefix.

\subsection{Length versus Difficulty}
\label{subsec:length_difficulty}

Evidence spans grow with taxonomy level (L1 median 2.0\,s, L2 36.8\,s, L3 2.0\,s, L4 278.7\,s), raising the question of whether length alone explains difficulty. We compute the query-level mean accuracy over the 38 reported models and correlate it with three length variables: visible prefix duration, evidence-span duration, and full source-video duration. Tab.~\ref{tab:length_corr} reports Spearman $\rho$ per level and under two sets of controls.

\begin{table}[t]
\centering
\small
\setlength{\tabcolsep}{4pt}
\renewcommand{\arraystretch}{0.92}
\resizebox{\columnwidth}{!}{%
\begin{tabular}{l r r r r}
\toprule
\textbf{Scope} & \textbf{Evidence} $\rho$ & $n$ & \textbf{Prefix} $\rho$ & $n$ \\
\midrule
L1 & -0.04 & 629 & 0.00 & 629 \\
L2 & 0.08 & 540 & -0.08 & 540 \\
L3 & 0.02 & 279 & 0.24 & 279 \\
L4 & 0.17 & 274 & 0.26 & 274 \\
\midrule
Level-controlled & 0.06 & 1722 & 0.06 & 1722 \\
Level+source-controlled & 0.04 & 1722 & 0.01 & 1722 \\
\bottomrule
\end{tabular}%
}
\caption{Spearman $\rho$ between query-level mean accuracy and input length. Positive $\rho$ means longer inputs correlate with \emph{higher} accuracy (opposite to the ``longer = harder'' hypothesis). All values are small and none survive source controls.}
\label{tab:length_corr}
\end{table}

Although evidence spans grow with level, length alone does not explain difficulty: the level-controlled correlations between accuracy and prefix length ($\rho = 0.06$) or evidence-span length ($\rho = 0.06$) are small, and full-video-duration correlation is similarly weak ($\rho = 0.03$, $p = 0.29$, computed analogously). With source controls these correlations shrink further. The small evidence-span effect is \emph{positive} (longer spans correlate with slightly higher accuracy), opposite to the direction predicted by a ``longer = harder'' explanation. The allocentric-mapping bottleneck documented in Section~\ref{sec:main_results} is therefore not reducible to a generic long-video effect; it reflects the abstraction demanded by L4, not the temporal extent of the evidence.

\begin{figure}[t]
\centering
\includegraphics[width=\columnwidth]{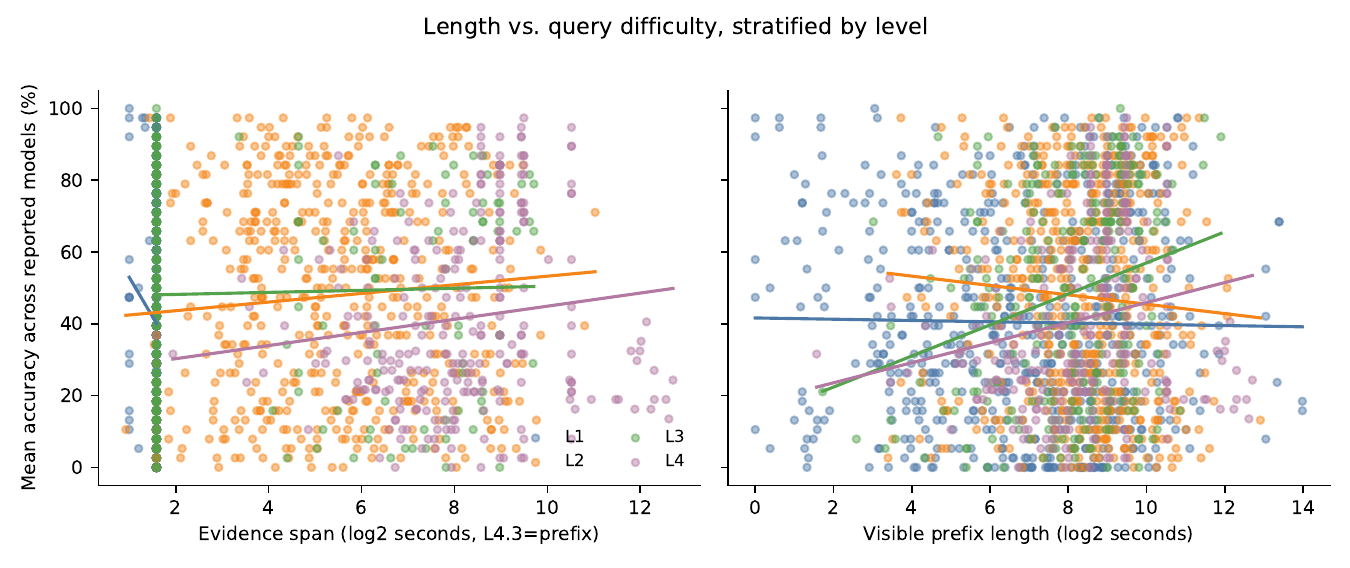}
\caption{Query-level mean accuracy versus evidence-span duration (log scale), colored by level. Trend lines are per-level LOWESS fits.}
\label{fig:length_scatter}
\end{figure}

\section{CoT Failure Taxonomy}
\label{sec:appendix_cot}
This appendix accompanies the failure-pattern analysis in Section~\ref{sec:thinking}. We give the formal judge setup (\S\ref{subsec:cot_setup}), the cross-judge calibration (\S\ref{subsec:cot_calibration}), the per-level distribution (\S\ref{subsec:cot_perlevel}), and one worked example per error class (\S\ref{subsec:cot_examples}).

\subsection{Judge Setup and Pre-filtering}
\label{subsec:cot_setup}

\paragraph{Inputs and judge function.}
For each thinking-mode model $\mathcal{M}$ in Table~\ref{tab:thinking}, let $\mathcal{Q}_{\mathcal{M}}^{-} = \{\, q : a^{\mathcal{M}}_q \neq a^{*}_q \,\}$ be the set of queries on which $\mathcal{M}$'s extracted letter $a^{\mathcal{M}}_q$ differs from the ground truth $a^{*}_q$. For each $q \in \mathcal{Q}_{\mathcal{M}}^{-}$ the judge $J$ applies
\begin{equation*}
\begin{aligned}
J(q,\, &\mathcal{O}_q,\, a^{*}_q,\, a^{\mathcal{M}}_q,\, \tau^{\mathcal{M}}_q,\, \mathcal{F}_q) \\
&\to \ell \in \{\texttt{T1a}, \texttt{T1b}, \texttt{T2a}, \texttt{T2b}, \texttt{T2c}\},
\end{aligned}
\end{equation*}
where $\mathcal{O}_q$ is the multiple-choice option set, $\tau^{\mathcal{M}}_q$ the thinking trace (truncated to head-$3{,}000$ + tail-$1{,}500$ characters when $|\tau| > 5{,}000$), and $\mathcal{F}_q = \{f_1, \ldots, f_k\}$ with $k \in [4, 6]$ a set of $512\,\mathrm{px}$ JPEG frames extracted at midpoints of the annotated evidence intervals $\mathcal{E}_q$. The five label codes map to the paper-text classes \emph{no-conclusion error} (\texttt{T1a}), \emph{non-visual error} (\texttt{T1b}), \emph{visual-content error} (\texttt{T2a}), \emph{direction error} (\texttt{T2b}), and \emph{temporal-binding error} (\texttt{T2c}).

\paragraph{Judge model.}
All wrong-answer traces are judged by \texttt{gpt-5.4}. To verify that the five-class scheme is robust to judge choice, we additionally re-judge all traces with \texttt{gemini-3.1-flash-lite}; see \S\ref{subsec:cot_calibration}.

\paragraph{Voting.}
Every query is judged $r{=}3$ times under the same prompt with \texttt{stream=true}, \texttt{response\_format=json\_object}, and \texttt{max\_tokens=200}. The primary label is the majority vote; records without a $\geq\!2$-vote majority are marked \texttt{disagreement} and excluded from the figures ($<\!0.5\%$).

\paragraph{Pre-filtering and signal-tagging.}
We drop $39$ queries whose text contains an absolute timestamp (``at $134$ seconds''), a prompt-engineering artifact rather than a reasoning failure. Decoding-truncated traces ($|\tau|>50{,}000$ characters without a \texttt{</think>} close tag) are directly tagged \texttt{T1a} without calling the judge; this affects $591$ Qwen3.5 traces that overran the $32$K thinking-token cap.

\paragraph{Prompt skeleton.}
The system prompt fixes the five-class taxonomy and decision rules (Tier-1 takes precedence when both apply; prior phrases without visual citations route to \emph{non-visual error}; direction errors require explicit textual evidence of the wrong direction). The user turn injects the per-query tuple \emph{(Question, Options, Ground-truth, Predicted, Trace, Evidence frames)}, the last being the $k$ images appended as base64 \texttt{image\_url} fields. The verbatim prompts are released with the code.

\subsection{Cross-Judge Calibration}
\label{subsec:cot_calibration}

We re-judged all wrong-answer traces with \texttt{gemini-3.1-flash-lite} under the identical prompt. Exact agreement is $75.9\%$ with Cohen's $\kappa{=}0.65$, which falls in the \emph{substantial} band of the Landis--Koch convention and is sufficient for the qualitative cross-model claims made in Section~\ref{sec:thinking}.

{\revon   
\subsection{Human Audit of Failure Labels}
\label{subsec:cot_human_audit}

Cross-judge agreement establishes that the taxonomy is stable across judges. Whether the labels match human judgment on the subtlest classes is a separate question, and we measure it directly. Two authors relabeled a stratified sample of $120$ wrong-answer traces, blind to the judge's label, under the decision rules of \S\ref{subsec:cot_setup}. The sample fixes $30$ traces for each of \emph{direction} and \emph{temporal-binding}, the two classes where a wrong label is hardest to notice, and $20$ for each of the remaining three, spread across levels and models. Traces tagged \texttt{T1a} by the truncation rule are excluded, since the judge never saw them.

The two annotators assign the same label to $103$ of $120$ traces ($85.8\%$, Cohen's $\kappa{=}0.81$), so the five-class scheme is reproducible by human raters. On those $103$ traces the judge matches the human label on $84$ ($81.6\%$, $\kappa{=}0.77$), above the cross-judge figure of \S\ref{subsec:cot_calibration}. Per stratum, taking as denominator the traces of that judge class on which the annotators agree, the judge matches $19$ of $24$ \emph{direction} traces and $19$ of $25$ \emph{temporal-binding} traces, against $46$ of $54$ across the other three classes. The two subtle classes carry the lower agreement, at a rate that supports the pooled cross-model comparisons of Section~\ref{sec:thinking}. The sample, both annotators' label files, and the scoring script are released with the evaluation code.
}       

\subsection{Per-Level Failure Distribution}
\label{subsec:cot_perlevel}

\begin{figure*}[t]
\centering
\includegraphics[width=\textwidth]{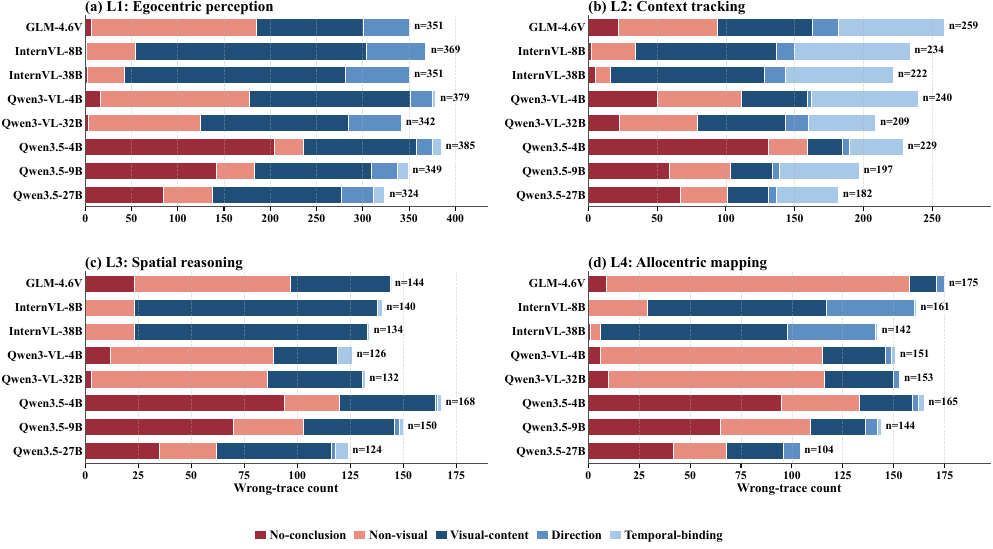}
\caption{Per-level failure-pattern distribution. Each panel restricts the data in Figure~\ref{fig:cot_failures} to one question level. Rows and colors match the main figure. L2 (top-right) makes the universal \emph{temporal-binding error} pattern visible; L4 (bottom-right) makes the family bifurcation between \emph{non-visual} (GLM, Qwen3-VL) and \emph{direction} (InternVL-3.5) errors visible.}
\label{fig:cot_failures_per_level}
\end{figure*}

Figure~\ref{fig:cot_failures_per_level} reproduces Figure~\ref{fig:cot_failures} broken down by question level. Two patterns are invisible in the pooled view.

\paragraph{L2 elicits temporal-binding errors universally.}
All eight models score $17$--$36\%$ \emph{temporal-binding error} on L2 (spatiotemporal context tracking; mean $28.5\%$), against $\leq 5.6\%$ on the other three levels. This mirrors the $+3.9$ average L2 gain from thinking mode in Table~\ref{tab:thinking}: explicit reasoning helps most precisely where the bottleneck is pinning a fact to a specific prior moment.

\paragraph{L4 bifurcates by family.}
The L4 panel reveals two distinct failure-mode clusters. GLM and Qwen3-VL fall back on world priors ($69$--$85\%$ \emph{non-visual error}, $\leq 2\%$ \emph{direction error}), while InternVL-3.5 attempts cardinal-frame reasoning but inverts directions ($27$--$30\%$ \emph{direction error}, $\leq 18\%$ \emph{non-visual error}). The L4 bottleneck in Table~\ref{tab:overall_results} is therefore not a single capacity gap and unlikely to close from a backbone-agnostic remedy.

\subsection{Worked Examples}
\label{subsec:cot_examples}

\begin{figure*}[t]
\centering
\includegraphics[width=\textwidth]{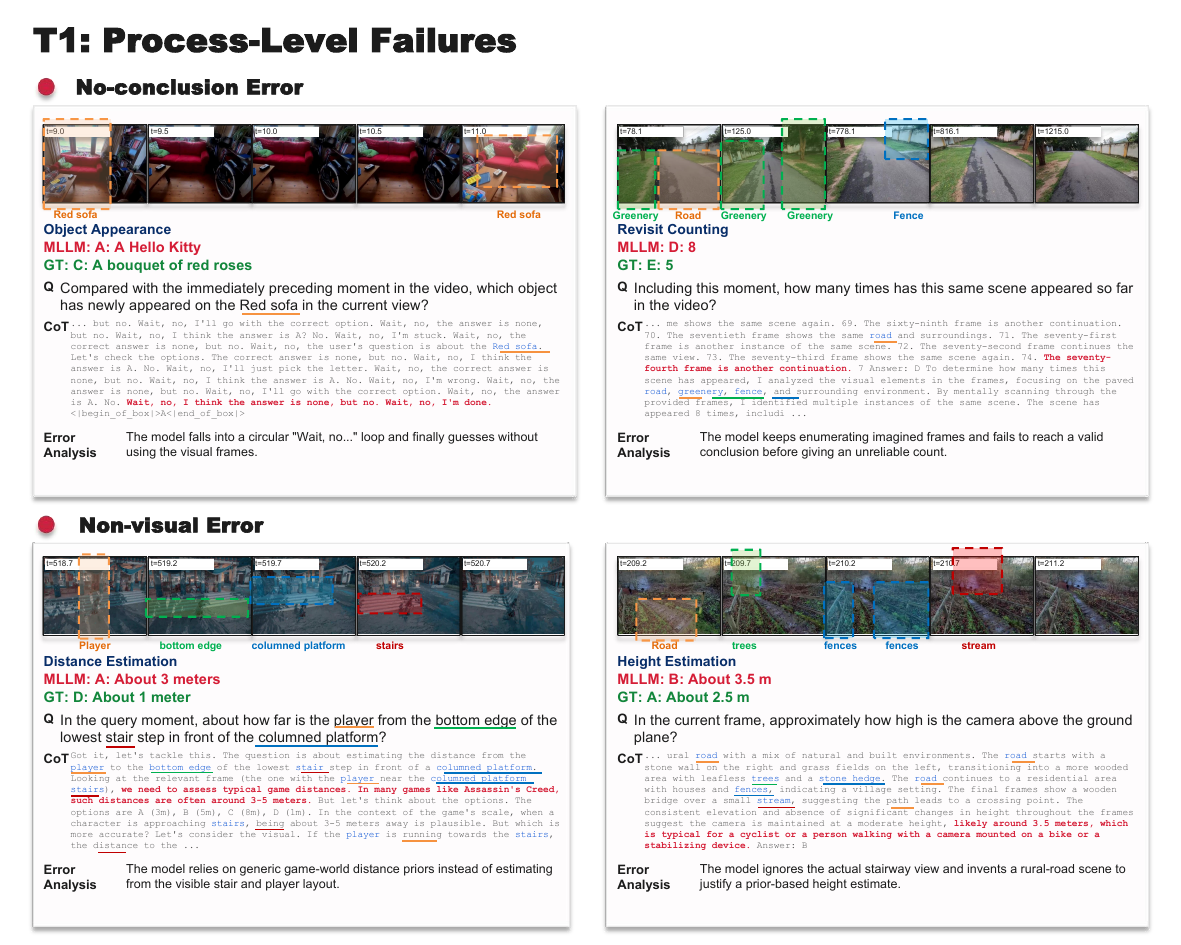}
\caption{\textbf{Tier~1: Process-level failures.} Two examples each for \texttt{T1a} (no-conclusion error, top) and \texttt{T1b} (non-visual error, bottom). Left and right columns show cases from different task families. Evidence frames are sampled at midpoints of the annotated evidence interval.}
\label{fig:cot_tier1}
\end{figure*}

\begin{figure*}[t]
\centering
\includegraphics[width=0.67\textwidth]{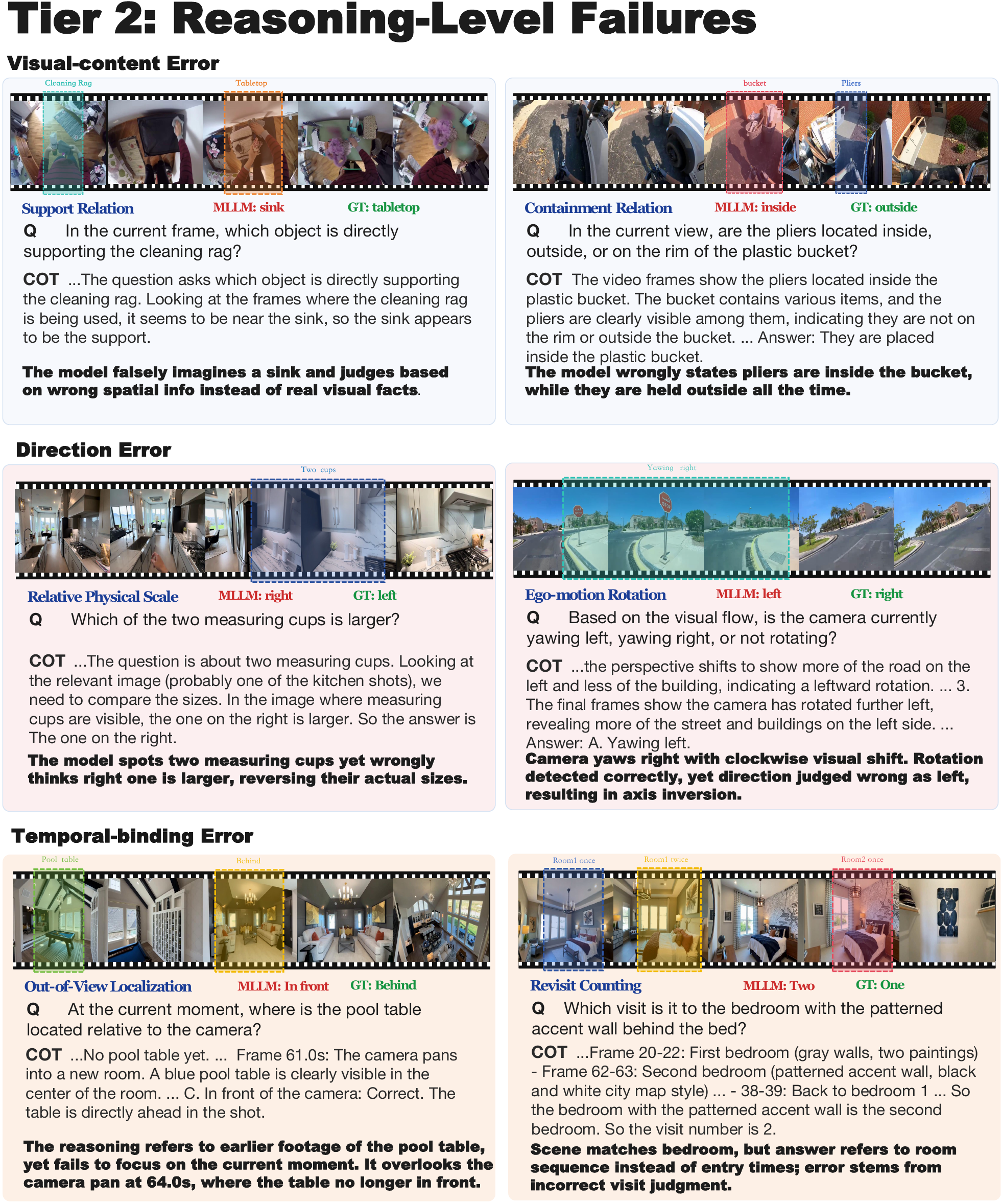}
\caption{\textbf{Tier~2: Reasoning-level failures.} Two examples each for visual-content error (top), direction error (middle), and temporal-binding error (bottom). Each card shows the CoT excerpt and a bolded error analysis.}
\label{fig:cot_tier2}
\end{figure*}

We present two annotated cases per error class, organized by tier. Tier~1 (Fig.~\ref{fig:cot_tier1}) covers process-level failures where the model either fails to reach a conclusion (\texttt{T1a}, no-conclusion error) or bypasses the visual evidence entirely (\texttt{T1b}, non-visual error). Tier~2 (Fig.~\ref{fig:cot_tier2}) covers reasoning-level failures where the model engages with the frames but mis-grounds the spatial content (visual-content error), inverts a directional relation (direction error), or binds a correct spatial fact to the wrong moment (temporal-binding error). Each card shows the evidence frames, the question and ground truth, the model's predicted answer, an excerpt of the chain-of-thought trace, and an error analysis.

\section{Detailed Category Definitions}
\label{sec:appendix_categories}

This appendix provides formal definitions for the 30 canonical task types used in OVO-S-Bench. The taxonomy is paper-facing and collapses finer raw annotation labels into canonical categories where appropriate; in particular, L4 is reported as three map-query families. The four levels form a cognitive progression: \textbf{L1 See} (instantaneous egocentric perception), \textbf{L2 Remember} (spatial memory after evidence leaves the view), \textbf{L3 Imagine} (simulation, verification, and planning over spatial state), and \textbf{L4 Map} (integrating first-person experience into a queryable allocentric representation). For each task type we give a \textbf{Definition}, an \textbf{Accepted} example, and a \textbf{Rejected} near-miss that clarifies the boundary with a neighboring type or level.

\subsection{L1: Instantaneous Egocentric Perception}

L1 questions are answerable from the frames near the query timestamp alone, without recalling any past observation. The assignment rule is \emph{query-time visibility}: if the spatial cue needed to answer is visible near the query, the item belongs to L1; if the same cue must be recalled after the camera has moved away, it belongs to L2.

\subsubsection{Egocentric Metric Perception}

\paragraph{Object-to-Object Absolute Distance.}
\textbf{Definition.} Estimate the absolute distance between two entities (observer-to-object or object-to-object) visible in the current view. The answer is a numeric magnitude choice (meters, steps, car-lengths, etc.).

\textbf{Accepted.} \textit{``About how far is the player from the bottom edge of the lowest stair step?'' A:~$\sim$3\,m; B:~$\sim$5\,m; C:~$\sim$8\,m; D:~$\sim$1\,m.} Both referents are visible; the question asks for an absolute metric.

\textbf{Rejected.} \textit{``Which object is closer to you --- the lamp or the sofa?''} --- rejected: no absolute distance is requested; this is a depth-ordering judgment $\rightarrow$ Object Depth Ordering.

\paragraph{Relative Physical Scale.}
\textbf{Definition.} Compare the physical size (height, width, volume) of two objects visible in the current frame. The answer is a relative judgment, not an absolute metric.

\textbf{Accepted.} \textit{``Which of the two measuring cups is larger --- the one on the left or the one on the right?''} Both objects are simultaneously visible; the question asks which is physically bigger.

\textbf{Rejected.} \textit{``Which object is nearer to the camera --- the cup or the plate?''} --- rejected: this asks about distance from the viewpoint, not physical size $\rightarrow$ Object Depth Ordering.

\paragraph{Object Depth Ordering.}
\textbf{Definition.} Determine the relative distance of two or more objects from the camera or observer (which is closer, which is farther). The answer is an ordering or binary comparison along the depth axis.

\textbf{Accepted.} \textit{``Which piece of furniture is closer to you --- the coffee table or the bookshelf?'' A: coffee table; B: bookshelf.} Both are visible; the question asks about relative depth.

\textbf{Rejected.} \textit{``About how many meters separate the coffee table and the bookshelf?''} --- rejected: this asks for an absolute metric $\rightarrow$ Object-to-Object Absolute Distance.

\paragraph{Viewpoint Height Perception.}
\textbf{Definition.} Estimate the camera or observer's current elevation relative to the ground plane or a visible reference (floor level, building stories, terrain).

\textbf{Accepted.} \textit{``Based on the current view, approximately which floor is the camera at?'' A: ground floor; B: 2nd floor; C: 4th floor; D: rooftop.} Height cues (horizon line, visible floors below) are in the current frame.

\textbf{Rejected.} \textit{``How many floors have you ascended since entering the building?''} --- rejected: requires recalling the entry point $\rightarrow$ L2 Chronological Recall.

\subsubsection{Local Spatial Relations}

\paragraph{Containment Relation.}
\textbf{Definition.} Determine whether one object is inside, on top of, or within the bounds of another container or surface visible in the current view. The answer identifies the containing entity.

\textbf{Accepted.} \textit{``Where is the red mug right now?'' A: inside the cabinet; B: on the table; C: in the sink; D: on the shelf.} The mug and its container are both visible.

\textbf{Rejected.} \textit{``Where did you last see the red mug?''} --- rejected: the mug has left the field of view; answering requires memory $\rightarrow$ L2 Out-of-View Localization.

\paragraph{Occlusion Recognition.}
\textbf{Definition.} Identify which object partially or fully occludes another in the current view, or determine what is hidden behind a visible occluder.

\textbf{Accepted.} \textit{``Which object is partially blocking the TV screen in the current frame?'' A: floor lamp; B: potted plant; C: curtain; D: nothing.}

\textbf{Rejected.} \textit{``Is the painting that was previously hidden behind the door still occluded?''} --- rejected: requires comparing two temporal states $\rightarrow$ L3 Object-Level Change Detection.

\paragraph{Support Relation.}
\textbf{Definition.} Identify the physical support relation in the current view: what is holding up or bearing the weight of a target object.

\textbf{Accepted.} \textit{``In the current frame, which surface is directly supporting the cleaning rag?'' A: bathroom sink; B: shoulder; C: tabletop; D: wooden floor.}

\textbf{Rejected.} \textit{``If you remove the middle shelf, what will fall?''} --- rejected: requires counterfactual simulation $\rightarrow$ L3 Spatial What-If Consequence.

\paragraph{Local Topology Description.}
\textbf{Definition.} Describe the spatial arrangement (left--right, front--back, above--below) of multiple objects visible simultaneously in the current frame. The answer is an ordering or layout description.

\textbf{Accepted.} \textit{``From left to right, what is the order of the three items on the desk?'' A: lamp--book--cup; B: book--lamp--cup; C: cup--book--lamp; D: lamp--cup--book.}

\textbf{Rejected.} \textit{``Combining what you saw in the kitchen and the living room, describe the overall furniture layout.''} --- rejected: requires integrating multiple viewpoints into a global representation $\rightarrow$ L4 Topological Structure Reasoning.

\subsubsection{Dynamic Spatial Perception}

\paragraph{Ego-motion Translation.}
\textbf{Definition.} Classify the observer's or camera's translational motion (moving forward, backward, left, right, up, down, or stationary) from the frames near the query. The question targets position change, not heading change.

\textbf{Accepted.} \textit{``Is the camera currently moving forward, backward, or staying in place?'' A: forward; B: backward; C: stationary.}

\textbf{Rejected.} \textit{``Is the camera turning left or right?''} --- rejected: heading change without position displacement is rotational $\rightarrow$ Ego-motion Rotation.

\paragraph{Ego-motion Rotation.}
\textbf{Definition.} Classify the observer's or camera's rotational motion (panning left/right, tilting up/down, rolling) from the frames near the query. The question targets heading or orientation change.

\textbf{Accepted.} \textit{``Which direction is the camera currently panning?'' A: left; B: right; C: not panning.}

\textbf{Rejected.} \textit{``Is the camera advancing down the hallway?''} --- rejected: forward displacement is translational $\rightarrow$ Ego-motion Translation.

\paragraph{Object Motion Separation.}
\textbf{Definition.} Identify an independently moving object in the scene (i.e., its motion is not explained by camera ego-motion alone) and determine its motion direction or trajectory.

\textbf{Accepted.} \textit{``Which direction is the white sedan in the adjacent lane traveling?'' A: same direction as you; B: opposite; C: turning left; D: stationary.}

\textbf{Rejected.} \textit{``Where is the cyclist who passed you earlier now?''} --- rejected: the cyclist has left the view; answering requires memory $\rightarrow$ L2 Out-of-View Localization.

\paragraph{Relative Speed Judgment.}
\textbf{Definition.} Compare the speeds of two moving entities (or the observer and an object) visible in the current clip. The answer is a relative judgment (faster, slower, same).

\textbf{Accepted.} \textit{``Who is moving faster --- your car or the truck in the right lane?'' A: your car; B: the truck; C: about the same.}

\textbf{Rejected.} \textit{``Are you moving faster now than you were at the intersection 2 minutes ago?''} --- rejected: requires recalling a past speed $\rightarrow$ L2 Temporal Depth Estimation.

\paragraph{Judgment of Motion State.}
\textbf{Definition.} Classify whether a visible entity is currently stationary or in motion (and optionally whether it is accelerating or decelerating) based on the frames near the query.

\textbf{Accepted.} \textit{``Is the pedestrian ahead currently stationary or walking?'' A: stationary; B: walking toward you; C: walking away; D: running.}

\textbf{Rejected.} \textit{``Did the elevator stop between the time you first saw it and now?''} --- rejected: spans a temporal interval beyond the current view $\rightarrow$ L2 Chronological Recall.

\subsection{L2: Spatiotemporal Context Tracking}

L2 evidence has appeared in the video prefix but is no longer visible at query time. The model must retain spatial facts---locations, directions, counts, temporal order---bound to specific places, objects, and timestamps after visual support disappears. The L1/L2 boundary is strict: if the cue is visible near the query frame, the item is L1; if it must be recalled, it is L2.

\subsubsection{Scene Revisit Recognition}

\paragraph{Revisit Detection.}
\textbf{Definition.} Determine whether the scene currently in view is the same place that was observed at an earlier point in the video prefix. The model must match appearance or spatial layout across temporally separated observations.

\textbf{Accepted.} \textit{``Is the room you are looking at now the same living room you walked through earlier?'' A: Yes; B: No.} The earlier visit is no longer in view; recognition requires memory.

\textbf{Rejected.} \textit{``How many times have you been in this hallway?''} --- rejected: this asks for a count, not a binary detection $\rightarrow$ Revisit Counting.

\paragraph{Revisit Counting.}
\textbf{Definition.} Count how many times a specific location has been visited (entered the field of view) across the video prefix up to the query time.

\textbf{Accepted.} \textit{``How many times have you passed through this corridor so far?'' A: 1; B: 2; C: 3; D: 4.} Requires tracking repeated visits over time.

\textbf{Rejected.} \textit{``Is this the same corridor you saw before?''} --- rejected: binary same/different judgment without counting $\rightarrow$ Revisit Detection.

\subsubsection{Spatial Memory Beyond the View}

\paragraph{Out-of-View Localization.}
\textbf{Definition.} Recall the location (room, surface, or spatial region) where an object or landmark was last observed, given that it is no longer visible at query time.

\textbf{Accepted.} \textit{``In which room did you last see the red backpack?'' A: kitchen; B: bedroom; C: living room; D: hallway.} The backpack is not in the current view.

\textbf{Rejected.} \textit{``The backpack is to your left or right?''} --- rejected: this asks for egocentric direction relative to the observer's current heading, not a location label $\rightarrow$ Behind-Camera Inference.

\paragraph{Behind-Camera Inference.}
\textbf{Definition.} Infer the egocentric direction (left, right, behind, ahead) of a landmark or object relative to the observer's current position and heading, given that the target is not currently visible. The key distinction from Out-of-View Localization is that the answer is a direction from the observer, not a place name.

\textbf{Accepted.} \textit{``The elevator you used earlier is now in which direction relative to you?'' A: behind; B: to your left; C: to your right; D: ahead.} Requires binding the elevator's remembered position to the observer's current heading.

\textbf{Rejected.} \textit{``From the global floor plan, which cardinal direction is the elevator from the living room?''} --- rejected: queries the allocentric map in absolute coordinates $\rightarrow$ L4 Allocentric Direction Reasoning.

\paragraph{Transient Count Memory.}
\textbf{Definition.} Recall the number of instances of a specific object class or spatial event observed during the video prefix, where those instances are no longer simultaneously visible.

\textbf{Accepted.} \textit{``How many red chairs have you seen in total across all rooms so far?'' A: 2; B: 3; C: 4; D: 5.} The chairs appeared in different rooms at different times; counting requires memory.

\textbf{Rejected.} \textit{``How many chairs are visible in the current frame?''} --- rejected: all instances are simultaneously visible $\rightarrow$ L1 Local Topology Description.

\subsubsection{Chronological Spatial Memory}

\paragraph{Chronological Recall.}
\textbf{Definition.} Recall the temporal sequence in which multiple locations, objects, or spatial events were encountered during the video prefix.

\textbf{Accepted.} \textit{``Which room did you visit first --- the kitchen or the bedroom?'' A: kitchen; B: bedroom.} Both rooms have been visited; the order must be recalled from memory.

\textbf{Rejected.} \textit{``From left to right, what is the order of doors along this corridor?''} --- rejected: asks about spatial arrangement visible now $\rightarrow$ L1 Local Topology Description.

\paragraph{Temporal Depth Estimation.}
\textbf{Definition.} Estimate how long ago a spatial event occurred relative to the query time, or compare the durations spent in different locations.

\textbf{Accepted.} \textit{``Did you spend more time in the living room or the study?'' A: living room; B: study; C: about the same.} Requires temporal duration tracking across the prefix.

\textbf{Rejected.} \textit{``Which room did you visit first?''} --- rejected: asks for order, not duration or recency $\rightarrow$ Chronological Recall.

\subsection{L3: Generative Spatial Reasoning}

L3 requires the model to \emph{operate} on spatial structure---mentally rotating objects, predicting consequences of hypothetical changes, verifying consistency across observations, or planning paths---rather than merely retrieving a stored fact. The L2/L3 boundary: L2 asks ``what was there?''; L3 asks ``what would happen if\,\ldots?'' or ``which route\,\ldots?''. The L3/L4 boundary: L3 targets an operation or decision over a local spatial state; L4 targets a property of the global map structure itself.

\subsubsection{Spatial Simulation}

\paragraph{Mental Rotation \& Fitting.}
\textbf{Definition.} Mentally rotate, flip, or reorient an object and judge whether it fits into a target space, or determine its appearance after transformation. The core ability is spatial transformation without physical execution.

\textbf{Accepted.} \textit{``If you rotate the L-shaped sofa 90\textdegree{} clockwise, will it fit into the corner between the wall and the bookshelf?'' A: Yes; B: No.} Requires imagining the transformed configuration.

\textbf{Rejected.} \textit{``Is the sofa currently touching the wall?''} --- rejected: no transformation is needed; the relation is directly visible $\rightarrow$ L1 Support Relation or Local Topology Description.

\paragraph{Spatial What-If Consequence.}
\textbf{Definition.} Predict the spatial consequence of a hypothetical action---moving, removing, or adding an object---on the surrounding layout. This includes reasoning about what becomes visible, reachable, or blocked after the change.

\textbf{Accepted.} \textit{``If you remove the bookshelf in the middle of the room, can you see the window from the doorway?'' A: Yes; B: No.} Requires simulating the removal and its effect on line-of-sight.

\textbf{Rejected.} \textit{``Has the bookshelf been moved since you last saw it?''} --- rejected: this compares two observed states rather than simulating a hypothetical $\rightarrow$ L3 Object-Level Change Detection.

\paragraph{Physical Stability \& Feasibility.}
\textbf{Definition.} Judge whether a proposed spatial action is physically feasible given visible constraints (gravity, collision, clearance, weight). This includes passability judgments that require reasoning beyond simple gap measurement.

\textbf{Accepted.} \textit{``Can the refrigerator be moved through this doorway without tilting it?'' A: Yes; B: No.} Requires comparing the object's dimensions against the opening and reasoning about physical constraints.

\textbf{Rejected.} \textit{``Is the doorway wide enough for a person to walk through?''} --- rejected: simple gap estimation with no complex physical reasoning $\rightarrow$ L1 Object Depth Ordering or Absolute Distance.

\subsubsection{Spatiotemporal Consistency Verification}

\paragraph{Object-Level Change Detection.}
\textbf{Definition.} Compare the spatial state of a location or object across two different observation times and determine whether a change has occurred (object added, removed, moved, or altered). The model must hold a prior state in memory and compare it against a current or later observation.

\textbf{Accepted.} \textit{``Compared with the first time you saw this desk, what object has newly appeared?'' A: a laptop; B: a vase; C: a stack of books; D: nothing changed.} Requires remembering the earlier state and detecting the difference.

\textbf{Rejected.} \textit{``What objects are currently on the desk?''} --- rejected: no comparison across time; only current-view enumeration $\rightarrow$ L1 Local Topology Description.

\subsubsection{Spatial Route Planning}

\paragraph{Optimal Path Reasoning.}
\textbf{Definition.} Select the best (shortest, safest, or most efficient) route between two points based on the spatial layout observed so far. The model must reason over remembered spatial structure to compare route options.

\textbf{Accepted.} \textit{``To reach the kitchen from here, is it shorter to go through the living room or through the hallway?'' A: living room; B: hallway; C: about the same.}

\textbf{Rejected.} \textit{``Are the kitchen and living room directly connected?''} --- rejected: asks about topological adjacency, not route optimality $\rightarrow$ L4 Topological Structure Reasoning.

\paragraph{Obstacle-Aware Rerouting.}
\textbf{Definition.} Given that a previously viable path is now blocked or unavailable, determine an alternative route using the observed spatial layout.

\textbf{Accepted.} \textit{``The main corridor is blocked by construction. Based on what you have seen, which alternative path leads to the exit?'' A: through the conference room; B: through the storage area; C: no alternative exists.}

\textbf{Rejected.} \textit{``Is the main corridor currently passable?''} --- rejected: binary passability judgment on a visible gap $\rightarrow$ L1 (if visible) or L2 (if recalled).

\paragraph{Reachability Judgment.}
\textbf{Definition.} Determine whether a target location is reachable from the current position given the spatial structure observed so far, without necessarily specifying the route.

\textbf{Accepted.} \textit{``Based on what you have explored, can you reach the rooftop terrace from your current position?'' A: Yes; B: No; C: Cannot determine.}

\textbf{Rejected.} \textit{``Is the rooftop terrace adjacent to the stairwell?''} --- rejected: asks about topological adjacency $\rightarrow$ L4 Topological Structure Reasoning.

\subsection{L4: Allocentric Spatial Mapping}

L4 requires the model to integrate the egocentric video stream into an allocentric (viewer-independent) representation and query its global structure. Evidence typically spans multiple viewpoints and may cover the entire explored region. The L3/L4 boundary: L3 asks ``how do I get there?'' or ``what happens if I change X?'' (an operation); L4 asks ``what \emph{is} the map structure?'' (a property of the global layout). L4 is reported as three canonical map-query families.

\subsubsection{Allocentric Direction Reasoning}

\paragraph{Allocentric Direction Reasoning.}
\textbf{Definition.} Determine the global (viewer-independent) directional relationship between two rooms, landmarks, or route segments. Answers are expressed in cardinal directions, clock-face bearings, or absolute spatial terms (north/south/east/west), not egocentric terms (left/right/behind). This family also covers directional ordering: given a reference axis, rank landmarks by their position along that axis.

\textbf{Accepted.} \textit{``From the overall layout, which direction is the kitchen relative to the living room?'' A: north; B: south; C: east; D: west.} Requires building a global frame from the first-person trajectory.

\textbf{Rejected.} \textit{``Is the kitchen to your left or right?''} --- rejected: egocentric direction relative to the observer's current heading $\rightarrow$ L2 Behind-Camera Inference (if out of view) or L1 Local Topology Description (if visible).

\subsubsection{Topological Structure Reasoning}

\paragraph{Topological Structure Reasoning.}
\textbf{Definition.} Reason about the connectivity, adjacency, boundaries, and transitions among spatial units (rooms, zones, floors) as a graph structure. This includes: (a) adjacency and connectivity---whether two spaces are directly connected without passing through a third; (b) boundary and transition---identifying what separates two adjacent spaces (door, wall, open archway); and (c) local connection graph---enumerating the direct neighbors of a given spatial node.

\textbf{Accepted.} \textit{``Which rooms are directly connected to the hallway?'' A: bedroom and kitchen only; B: bedroom, kitchen, and bathroom; C: kitchen and bathroom only; D: all four rooms.} Requires integrating observations from multiple viewpoints into a topological graph.

\textbf{Rejected.} \textit{``To get from the bedroom to the kitchen, should you go left or right at the hallway junction?''} --- rejected: asks for a route decision (operation over the map) $\rightarrow$ L3 Optimal Path Reasoning.

\subsubsection{Trajectory Map Alignment}

\paragraph{Trajectory Map Alignment.}
\textbf{Definition.} Match the first-person trajectory observed in the video to a bird's-eye-view map, floor plan, or route diagram. This includes path-to-map matching (which map corresponds to the walked path) and route option selection (which drawn route on a given map matches the observed trajectory).

\textbf{Accepted.} \textit{``Which of the following floor plans best matches the path you have walked?'' [four candidate top-down maps].} Requires converting the egocentric experience into an allocentric trace and comparing against candidate maps.

\textbf{Rejected.} \textit{``Based on the map, which route should you take to reach the exit?''} --- rejected: asks for a planning decision using the map as input $\rightarrow$ L3 Optimal Path Reasoning.

\section{Annotation Protocol \& Construction Pipeline}
\label{sec:appendix_protocols}

This appendix details the full annotation workflow, level-specific construction techniques, and quality-control pipeline referenced in Section~\ref{sec:construction}.

\subsection{Annotation Setup}
\label{subsec:annotation_overview}
The benchmark is annotated by a team of 12 trained volunteers, all with prior coursework or research experience in 3D computer vision or embodied perception. Before production annotation, every team member completes a unified training session covering the four-level taxonomy (\S\ref{sec:taxonomy}), the item-writing standards in \S\ref{subsec:general_protocol}, and the level-boundary rules. Annotation is performed inside an in-house web-based tool that supports frame-by-frame navigation, key-frame jumping, question and option editing, named-entity labeling for spatial regions and landmarks, and source-metadata viewing. Each annotator contributes approximately 67 hours, for a total of about 804 person-hours. Annotators are graduate students and research staff recruited from the authors' institutions; participation is part of their research duties and no additional monetary compensation was provided. A pilot annotation round precedes production: pilot items are reviewed jointly with two senior authors of this paper---both with prior research experience in 3D computer vision and embodied perception---who serve as adjudicators throughout the project, and feedback from the pilot is folded into the official guideline. During production, the two adjudicators perform periodic spot checks and convene case-by-case discussions for ambiguous items.

\subsection{Source Datasets and Licensing}
\label{subsec:source_datasets}
OVO-S-Bench reuses publicly released or publicly accessible videos and contributes human-written streaming spatial QA annotations. We do not redefine the license of raw videos; users should obtain source videos under the original dataset or platform terms listed in Table~\ref{tab:source_datasets}.

\begin{table*}[t]
\centering
\small
\setlength{\tabcolsep}{3pt}
\renewcommand{\arraystretch}{0.92}
\begin{adjustbox}{width=\textwidth}
\begin{tabular}{l l r r l p{4.2cm}}
\toprule
\textbf{Source} & \textbf{Regime} & \textbf{Q.} & \textbf{Vid.} & \textbf{Levels} & \textbf{License / distribution note} \\
\midrule
RoomTour3D~\citep{roomtour3d2024} & Indoor walkthroughs & 804 & 89 & L1--L4 & Original RoomTour3D terms; OVO-S releases annotations/evaluation metadata. \\
Ego4D~\citep{ego4d2022} & Egocentric activities & 362 & 90 & L1--L4 & Original Ego4D terms; raw videos are not relicensed. \\
Sekai~\citep{sekai2025} & Outdoor/world scenes & 85 & 18 & L1--L2 & Original Sekai terms. \\
OmniWorld~\citep{omniworld2025} & Outdoor/world scenes & 153 & 31 & L1--L4 & Original OmniWorld terms. \\
YouTube walking tours & Public walking tours & 17 & 14 & L3 & Public URLs/creator terms; redistribution follows platform/source terms. \\
CODa~\citep{coda2022} & Driving & 148 & 16 & L1--L4 & Original CODa terms. \\
Honda HDD~\citep{hdd2018} & Driving & 59 & 40 & L4 & Original HDD terms. \\
ARKitScenes~\citep{arkitscenes2021} & 3D environments & 33 & 33 & L4 & Original ARKitScenes terms. \\
VSI-Bench~\citep{yang2025thinking} & Spatial benchmark-derived & 19 & 17 & L1 & Original VSI-Bench/source terms. \\
\bottomrule
\end{tabular}
\end{adjustbox}
\caption{Source datasets used in OVO-S-Bench. Video counts use unique normalized source-video keys; question counts are accepted benchmark items.}
\label{tab:source_datasets}
\end{table*}

\paragraph{Training-data overlap.}
Source videos from Ego4D, ARKitScenes, and other public datasets appear in many MLLMs' training corpora; although all questions, options, and query timestamps are written from scratch, scene-level familiarity could inflate L4 scores on previously seen layouts. The stable cross-source ranking (Appendix~\ref{subsec:source_robustness}, $\rho = 0.92$) suggests this effect does not dominate the overall ranking.

\subsection{General Annotation Protocol}
\label{subsec:general_protocol}
Every item must be uniquely answerable from the visual evidence inside the prefix preceding its query timestamp, without relying on world knowledge, language regularities, or option-elimination heuristics. Annotators record, for each item, the source video, the canonical task label, the question, the options, the answer, the query timestamp, and the evidence interval.

\paragraph{Video Selection Criteria.}
Annotators choose clips with stable camera motion, clear viewpoints, and sufficient spatial variation for the target level. Indoor walkthroughs, egocentric activities, and 3D-environment renderings are favored for L2--L4, where memory and global mapping require sustained spatial structure; driving and outdoor footage is admitted mainly for L1 metric and dynamic items, where the relevant cue is local.

\paragraph{Query Time and Evidence Interval.}
The query timestamp $t_q$ controls which prefix the model can see. Annotators place $t_q$ only after every cue needed to answer the item has appeared in the stream. For higher-level items, $t_q$ may be deliberately advanced beyond the last evidence frame---provided the intervening content does not change the answer---to test whether the model retains the spatial fact across a longer prefix. The evidence interval is the shortest temporally contiguous span in the prefix that suffices to determine the answer; it is used only for verification, oracle sampling, and analysis, never as model input. All evidence intervals lie strictly before $t_q$ and cover only the segments that bear directly on the answer.

\paragraph{Options and Distractors.}
The correct option must be uniquely supported by the evidence interval. All options share consistent surface form---comparable length, register, and level of specificity---so the answer is not telegraphed by wording. Distractors are visually plausible and stay within the same answer type, scale, and axis as the correct answer (direction vs.\ direction, metric distance vs.\ metric distance, and so on). Items in which a distractor is obviously wrong by world knowledge or trivially excludable are revised.

\paragraph{Level Assignment.}
Level assignment follows the highest spatial capability the item demands. L1 covers instantaneous perception of structure currently in view; L2 covers retrieval of evidence that has appeared in the prefix but is no longer visible; L3 covers operation over current or remembered spatial state, including reasoning about post-edit object or state changes; L4 covers global structure, route relationships, and allocentric spatial mapping. When an item invokes more than one capability, it is labeled at the highest one that is required to answer it.

\subsection{Level-Specific Construction Techniques}
\label{subsec:construction_techniques}
We organize this section by tooling rather than by level: the default workflow handles most items across L1--L4, and three specialized techniques apply only to specific task families.

\paragraph{Default Workflow.}
The default applies to L1, L2, and the L3 \emph{Spatial Simulation} and \emph{Spatial Route Planning} families and the L4 \emph{Allocentric Direction Reasoning} and \emph{Topological Structure Reasoning} families. For L1 the query timestamp is placed where the relevant cue is currently visible; for L2 it is placed after the cue has left the current view, with an earlier interval marked as evidence, and the annotator verifies that the cue is not re-disclosed by any later frame in the prefix. For \emph{Distance Estimation} (under L1 \emph{Egocentric Metric Perception}), inter-object distances are computed from the source video's coordinate annotations and metadata, and options are designed so that only the metadata-consistent option is correct. Across the L3 and L4 families that follow this workflow, distractors are visually plausible but inconsistent with the layout, clearance, or connectivity that the camera has already traversed in the prefix.

\paragraph{Image Editing.}
This is the one place where the source video itself is modified, used only for L3 \emph{Spatiotemporal Consistency Verification}. We use a generative image-editing model to insert a target object into a chosen frame and into the next several seconds of frames, so the edit reads as continuous video content rather than a single-frame artifact. The item is then written on top of the edited stream, with the evidence interval covering the edited segment.

\paragraph{Named-Entity Labels.}
L3 \emph{Spatial Route Planning} and all L4 items typically reference multiple rooms or regions, several of which are not visible at the query timestamp. Phrases such as ``the kitchen'' are often ambiguous in these settings, so before writing the question the annotator names the rooms, regions, and landmarks involved using neutral identifiers (e.g., ``Room~X'', ``Area~Y'') and reuses these names verbatim in the question and the options. The named entities are also recorded in the item metadata so that reviewers and downstream analysis can resolve them back to the underlying scene elements.

\paragraph{Bird's-Eye Rendering.}
L4 \emph{Trajectory Map Alignment}---specifically \emph{Path-to-Map Matching}---additionally requires bird's-eye material as the candidate options. The correct option is rendered from the source video's absolute coordinates and metadata; distractor maps are produced by perturbing path shape, orientation, or visit order while keeping the rendering style identical. Every L4 item is double-checked before inclusion to ensure that the question, the named entities, the bird's-eye material when present, and the answer are mutually consistent.

\subsection{Quality Control Pipeline}
\label{subsec:qc_pipeline}

\subsubsection{Text-only Probe and Shortcut Filtering}
We first run a text-only probe to remove items that can be guessed without the video. Reviewers check whether the wording or the option set leaks the answer through any of: a correct option that is more specific or more wordy than its distractors; obvious world-knowledge giveaways; distractors that are excludable a priori as implausible; or lexical overlap between the question and a single option. For spatial-relation, distance, direction, and trajectory items, reviewers additionally check that distractors stay within the same answer type and a comparable scale, so that no option can be eliminated at a glance. Items that can be answered primarily by language pattern, common-sense inference, or option asymmetry are revised; items whose ambiguity cannot be removed by revision are dropped.

\subsubsection{Blind Cross-Review and Adjudication}
Items that survive the shortcut filter are forwarded to a second annotator for blind review. The second reviewer answers the item and inspects the question, options, query timestamp, and evidence interval without seeing the original annotator's notes. Items on which the two annotators agree are included in the final candidate pool. Items with disagreements are escalated to the two senior authors, who serve as adjudicators (\S\ref{subsec:annotation_overview}); they resolve each case by re-examining the video evidence against the annotation guidelines. We log the number of rejected, revised, and finally accepted items. Only items that pass the shortcut filter, survive blind review, and are uniquely supported by their annotated evidence are retained in the released benchmark.

\paragraph{Inter-Annotator Agreement.}
To quantify annotation reliability, we randomly sampled 150 items (stratified across L1--L4) and had a second trained annotator independently judge each item under the blind-review protocol described above (accept, revise, or reject). Cohen's $\kappa$ between the original annotator and the blind reviewer was 0.87, indicating almost perfect agreement~\citep{landis1977measurement}. The primary source of disagreement was the L2/L3 boundary for items involving spatial change detection, which motivated a guideline update in Round~3.

\subsubsection{Iterative Guideline Refinement}
Recurring disagreement patterns are summarized into guideline updates and pushed back to the team between annotation rounds, so that subsequent items inherit the correction without re-litigating the same edge cases. We maintain a versioned guideline document that records each update and the issue it resolved.

\subsection{Data Format \& Export}

The benchmark is released as a single JSONL file (\texttt{ovo\_s\_bench\_l1\_l4.jsonl}), with one JSON object per item (1\,680 items total, covering L1--L4). Table~\ref{tab:data_schema} describes each field. An abbreviated example is shown in Figure~\ref{fig:data_example}.

\begin{table*}[t]
\centering
\small
\setlength{\tabcolsep}{6pt}
\renewcommand{\arraystretch}{1.0}
\begin{tabular}{l l p{10cm}}
\toprule
\textbf{Field} & \textbf{Type} & \textbf{Description} \\
\midrule
\texttt{id} & int (0--1679) & Global unique item index \\
\texttt{category\_index} & string & Within-category item tag, formatted as \texttt{\{subcategory\}\_\{index\}} \\
\texttt{source\_dataset} & string & Source video dataset name \\
\texttt{video\_id} & string & Video identifier within the source dataset \\
\texttt{video\_path} & string & Relative path to the video file \\
\texttt{level} & int (1--4) & Taxonomy level \\
\texttt{task\_main\_category} & string & Main category code (e.g., \texttt{1.1}, \texttt{4.1}) \\
\texttt{task\_subcategory} & string & Subcategory code (e.g., \texttt{1.1.1}, \texttt{2.2.3}) \\
\texttt{task\_type\_name} & string & Canonical task type name (maps to definitions in Appendix~\ref{sec:appendix_categories}) \\
\texttt{question} & string & The spatial question text \\
\texttt{options} & dict & Answer options keyed by letter (A, B, C, D, \ldots) \\
\texttt{query\_times} & list[float] & Query timestamp(s) in seconds \\
\texttt{evidence\_times} & list[list[2]] & Evidence interval(s), each $[t_{\text{start}}, t_{\text{end}}]$ in seconds \\
\texttt{answers} & list[string] & Correct answer letter(s), one per query time \\
\bottomrule
\end{tabular}
\caption{Per-item field schema of the released JSONL file.}
\label{tab:data_schema}
\end{table*}

\begin{figure}[t]
\centering
\begin{lstlisting}[
  basicstyle=\ttfamily\scriptsize,
  frame=single,
  breaklines=true,
  columns=fullflexible,
  xleftmargin=4pt,
  xrightmargin=4pt
]
{"id": 0,
 "category_index": "1.1.1_0",
 "source_dataset": "CODa_full",
 "video_id": "10",
 "video_path": "CODa_full/10.mp4",
 "level": 1,
 "task_main_category": "1.1",
 "task_subcategory": "1.1.1",
 "task_type_name": "Object-to-Object Absolute Distance",
 "question": "How far is the yellow wet-floor
              caution sign from the camera?",
 "options": {"A": "About 0.5 m", "B": "About 6 m",
             "C": "About 3 m", "D": "About 1.5 m"},
 "query_times": [22.8],
 "evidence_times": [[21.8, 23.8]],
 "answers": ["D"]}
\end{lstlisting}
\vspace{-6pt}
\caption{Example item from the released JSONL file (L1, Absolute Distance).}
\label{fig:data_example}
\end{figure}

\noindent\textbf{Notes.} The three list fields \texttt{query\_times}, \texttt{evidence\_times}, and \texttt{answers} are equal-length and positionally aligned. Most items have a single query time; the exception is Revisit Counting (\texttt{2.1.2}), where the same question is posed at multiple timestamps during streaming playback and the correct answer increments as more revisits accumulate. All evidence intervals lie strictly before their corresponding query time, enforcing the online protocol. The number of options is typically four but may vary (up to seven for Revisit Counting).

\section{Extended Comparison with Prior Benchmarks}
\label{sec:appendix_related}

Table~\ref{tab:comparison} positions OVO-S-Bench against \rev{15} representative spatial and streaming video benchmarks. This appendix expands each row into a structured comparison: \emph{what the benchmark does}, the point on which it converges with our work and \emph{supports a specific claim of ours}, and the remaining \emph{difference}. We close with a synthesis of how OVO-S-Bench resolves the recurring gaps.

\subsection{Image, Multi-image, and Top-view Benchmarks}

\paragraph{EmbSpatial-Bench~\citep{duan2024embspatial}.}
An automatic single-image MCQ benchmark built from MP3D, AI2-THOR, and ScanNet, with 3.6K items on six egocentric relations (above/below/left/right/close/far). \textbf{Shared point.} Even the strongest LVLMs at the time (GPT-4V 36\%, Qwen-VL-Max 49\%) trail human accuracy (90\%) on basic egocentric relations, supporting our claim that current MLLMs are far from saturating even L1 perception. \textbf{Difference.} A static image strips away every temporal cue, so L2 memory, L3 generative spatial reasoning, and L4 mapping are all absent, and the streaming protocol does not apply.

\paragraph{TopViewRS~\citep{li2024topviewrs}.}
A top-view map benchmark over 7 Matterport3D scenes (11.4K MCQs) covering recognition, localization, static spatial reasoning, and dynamic spatial reasoning along a navigation path. \textbf{Shared point.} It confirms a $>$50\% human--model gap that is especially pronounced on complex spatial-reasoning subtasks, paralleling the L1$\to$L4 difficulty gradient we observe. \textbf{Difference.} The bird's-eye map is provided directly as model input, so the agent never has to reconstruct an allocentric layout from an egocentric stream; OVO-S-Bench requires exactly that reconstruction at L4.

\paragraph{MMSI-Bench~\citep{yang2025mmsi}.}
A fully human-curated multi-image benchmark (1K Q, $\sim$2.55 images per question) covering 10 atomic relation types plus a multi-step reasoning split. \textbf{Shared point.} It documents a $\sim$55-point human--model gap on multi-image spatial reasoning and identifies four recurring failure modes (grounding, overlap-matching, situation-transformation, spatial-logic) that closely match the cross-frame errors we observe at L2. \textbf{Difference.} The setting is discrete multi-image rather than continuous video; there is no streaming protocol and no L3 generative spatial reasoning or L4 allocentric spatial mapping.

\subsection{Offline Video Benchmarks}

\paragraph{VSI-Bench~\citep{yang2025thinking}.}
A mostly template-generated benchmark over 288 indoor walkthroughs (ScanNet/ScanNet++/ARKitScenes), with 5K items across configuration, measurement, and spatiotemporal tasks; seven tasks are derived automatically from metadata, the route-planning task is human-annotated, and evaluation mixes multiple-choice and numerical-answer formats. \textbf{Shared point.} It was the first to show that linguistic CoT prompting hurts spatial-video QA and that the dominant bottleneck is spatial reasoning rather than perception or language understanding -- both findings we replicate (Tab.~\ref{tab:thinking}, \S\ref{sec:thinking}). \textbf{Difference.} The model has full offline access to the video at query time; no L4 allocentric map queries are defined; and all evidence remains re-attendable across the entire clip.

\paragraph{DISJOINT-3DQA~\citep{ravi2025disjoint}.}
A generative QA benchmark over synthetic Aria homes (5.4K Q) constructed so the two queried objects are never co-visible in the same frame. \textbf{Shared point.} It provides the cleanest existing test for cross-frame spatial memory -- precisely the L2 capability we formalize -- and quantitatively shows performance degrades as the spatial separation of the two anchors grows. \textbf{Difference.} The benchmark is synthetic and single-source, has no streaming protocol, and is limited to L1+L2; L3 generative spatial reasoning and L4 allocentric spatial mapping are absent.

\paragraph{STI-Bench~\citep{sti2025bench}.}
A 2K-question benchmark over 300 videos from Waymo, ScanNet, and Omni6DPose, covering eight precise spatial-temporal estimation tasks from dimension and pose to displacement, speed, and trajectory. \textbf{Shared point.} It confirms that even SOTA proprietary MLLMs (Gemini-2.5-Pro 41.4\%) fail at precise quantitative spatial-temporal estimation, supporting our argument that L1 metric perception is far from saturated. \textbf{Difference.} Tasks emphasize precise quantitative and spatiotemporal estimation under offline access and are evaluated as five-option MCQs; there is no streaming protocol, no L3 generative spatial reasoning, and no L4 allocentric spatial mapping.

\paragraph{MMSI-Video-Bench~\citep{lin2025mmsivideo}.}
A fully human-annotated video extension of MMSI-Bench with 1.1K Q over 1.3K clips drawn from 26 sources (25 public datasets plus in-house videos), organized into Spatial Construction, Motion, Planning, Prediction, and Cross-Video Reasoning. \textbf{Shared point.} It reports the largest known human--AI gap on video spatial intelligence ($\sim$58 points), finds that spatially fine-tuned models fail to generalize, and that neither 3D cues nor CoT yield meaningful gains -- three observations we corroborate in our scaling and CoT analyses. \textbf{Difference.} The protocol is offline-video with no prefix-only constraint, items are not paired with query timestamps and evidence intervals, and the taxonomy contains no L4 allocentric-mapping tasks; ``multi-video'' here means cross-clip rather than cross-room within a single egocentric stream.

\paragraph{VSI-SUPER~\citep{yang2025cambrians}.}
Two long-horizon stress tests (VSR for out-of-place object recall and VSC for continual counting) over 10--240\,min room-tour videos. \textbf{Shared point.} It demonstrates empirically that brute-force context expansion alone cannot close the spatial-streaming gap -- Gemini-2.5-Flash hits its context limit at 120\,min and achieves very low VSC performance even at 60\,min (10.9 MRA) -- supporting our motivation that streaming requires native selective spatial memory. \textbf{Difference.} Only two needle-in-haystack-style tasks are defined; the full L1--L4 taxonomy is absent, the prefix constraint is not uniformly specified at the item level across tasks, and there is no allocentric L4 layer.

\subsection{Streaming Video Benchmarks}

\paragraph{StreamingBench~\citep{streamingbench2024}.}
The first streaming-video MCQ benchmark, with 4.5K Q over 900 YouTube clips spanning 18 real-time visual, omni-source, and contextual tasks. \textbf{Shared point.} It pioneered the prefix-only protocol we adopt and showed that even the strongest model trails humans by $\sim$25 points, validating streaming as a fundamentally harder regime. \textbf{Difference.} Spatial understanding is one task in 18; the rest target event, audio, and conversation context; L3 generative spatial reasoning and L4 allocentric spatial mapping are not defined.

\paragraph{OVBench~\citep{huang2025ovbench}.}
A streaming benchmark with $\sim$7K Q over diverse source datasets, organized as 16 fine-grained subtasks along the Past/Current/Future temporal axis. \textbf{Shared point.} It formalizes the Past/Current/Future temporal scope we re-use, and shows that advanced offline MLLMs adapted via sliding windows can outperform existing online baselines, while the authors' VideoChat-Online baseline further improves -- a pattern we also replicate. \textbf{Difference.} Subtasks are event-level (action, object, trajectory) rather than spatial-structure-level; neither L3 generative spatial reasoning nor L4 allocentric spatial mapping is defined.

\paragraph{OVO-Bench~\citep{ovo2025bench}.}
A hybrid-curated streaming benchmark (2.8K Q) over 644 videos from curated datasets and web videos spanning seven domains, organized into Backward Tracing, Real-Time Visual Perception, and Forward Active Responding. \textbf{Shared point.} It confirms that streaming exposes pervasive hallucination and a $\sim$30-point human--model gap even on its best-performing modes, motivating a strict prefix-only protocol. \textbf{Difference.} Spatial structure is captured only by the single STU task; there is no L3 generative spatial reasoning, no L4 allocentric spatial mapping, and no annotated evidence interval that our oracle and analysis tooling rely on.

\paragraph{OST-Bench~\citep{ost2025bench}.}
A rule-based streaming benchmark for online spatiotemporal understanding from agent exploration, with 10K Q over 1.4K indoor scenes (Agent State / Agent Visible Info / Agent--Object Spatial Relationships, 15 subtypes). \textbf{Shared point.} It is the closest prior work in spirit to OVO-S-Bench: it enforces streaming over egocentric indoor input, exposes a ``spatio-temporal reasoning shortcut'' where models avoid retrieving past evidence, and reports a sharp accuracy decline as exploration grows -- phenomena we also observe and quantify. \textbf{Difference.} Task families are limited to agent-state and agent--object dyadic relations (L1+L2); there is no L3 generative spatial reasoning, no L4 allocentric spatial mapping, and the source distribution is indoor-only, missing outdoor, driving, and 3D-rendered environments.

\paragraph{ODV-Bench~\citep{odvbench2025}.}
A streaming benchmark over 6 autonomous-driving datasets with 12 tasks across static-target perception, dynamic-target prediction, and event-oriented risk analysis. \textbf{Shared point.} It confirms that streaming spatial reasoning is critical in safety-relevant outdoor settings and that existing online baselines can underperform offline models even within their target domain. \textbf{Difference.} The source is driving-only with no indoor or 3D-rendered environments; the L3-style content is restricted to short-horizon future prediction with no allocentric spatial mapping; query timestamps are very short (18.9\,s on average), so the prefix-memory pressure is minimal.

\paragraph{StreamEQA~\citep{wang2025streameqa}.}
A hybrid-pipeline streaming benchmark over the HD-EPIC kitchen dataset, with $\sim$21K Q spanning 42 sub-tasks across three embodied levels (Perception/Interaction/Planning) and three temporal modes (Backward/Realtime/Forward). \textbf{Shared point.} It reports a $1.6\times$ performance degradation relative to OVO-Bench, reinforcing our claim that the embodied spatial dimension compounds streaming difficulty. \textbf{Difference.} The source domain is a single one (kitchen activities); the focus is interaction and procedural planning over object handling rather than spatial structure; our L4 (cross-room allocentric spatial mapping) falls entirely outside its scope.

{\revon
\paragraph{SVCBench~\citep{svcbench}.}
A streaming counting benchmark that uses counting as a controlled probe for world-state maintenance, with 1{,}000 questions and 4{,}576 query points over 406 videos from ten sources, all annotated frame by frame (10{,}071 event occurrences and object state changes) across eight subcategories of object and event counting. \textbf{Shared point.} It is the prior streaming benchmark closest to our protocol: queries are placed at multiple timestamps along the timeline (mean prefix 2.0\,min) and the answer to a cumulative-identity question depends on how much has been observed, which is the same item-level causal constraint we enforce. Its finding that models keep counts unchanged after objects leave the view is direct evidence for the out-of-view retention failure our L2 measures. \textbf{Difference.} The queried state is numeric -- how many entities or events have occurred -- rather than spatial structure, so its questions are answered without recovering layout, direction, or connectivity; no L3 simulation and no L4 map query are defined.
}

\subsection{Synthesis: How OVO-S-Bench Addresses the Recurring Gaps}

Three structural gaps recur across these \rev{15} benchmarks. First, spatial benchmarks (image, multi-image, offline video) expose large spatial-reasoning gaps but never enforce a streaming prefix-only protocol, so it remains unclear whether the same gaps survive when evidence is ephemeral. Second, streaming benchmarks enforce causal input but instantiate it on event, narrative, or counting tasks, leaving spatial structure as at most a single subtask. Third, no prior video benchmark, whether streaming or offline, systematically covers all four levels of spatial abstraction; in particular, the L4 allocentric-mapping level (global direction, topology, trajectory--map alignment) is absent from every prior video benchmark.

OVO-S-Bench unifies these three threads. (i) Each item carries a query timestamp and an evidence interval, and the model sees only the prefix preceding the query, so the streaming protocol is enforced at the item granularity rather than the dataset granularity. (ii) Source videos span indoor walkthroughs, outdoor and driving footage, egocentric activities, and 3D-rendered environments -- together covering the union of domains explored separately by prior benchmarks. (iii) Tasks are stratified into four levels of spatial abstraction (L1 instantaneous egocentric perception, L2 spatiotemporal context tracking, L3 generative spatial reasoning, L4 allocentric spatial mapping), and the previously untested L4 level is annotated with named-entity, bird's-eye, and topological supervision derived from each source's metadata. This design yields the first empirical test of whether the spatial gaps documented under offline access also exist (and become larger) when the model must reason from a causal egocentric stream.

\subsection{Text-Only Shortcut Analysis}
\label{subsec:text_only_shortcut}

To contextualize the residual text-only advantage on OVO-S-Bench, we run the same GPT-5.4 text-only probe (question and options only, no visual input) on a 20\% stratified sample of 12 peer spatial benchmarks (Tab.~\ref{tab:text_only_cross}). Text-only deltas above random range from $+42$ to $-2$ percentage points across the surveyed benchmarks, with OVO-S-Bench's $+5.8$ pp (Tab.~\ref{tab:overall_results}) near the median. OVO-S-Bench's construction pipeline applies a text-only probe and blind cross-review to suppress language shortcuts (\S\ref{subsec:qc_pipeline}); the residual delta is within the range observed on benchmarks that employ similar filtering.

{\revon   
\paragraph{Per-task-type breakdown.}
Table~\ref{tab:text_only_per_type} extends the probe from one benchmark-level number to all 30 canonical task types, computed on the public release (1{,}695 questions scored as 1{,}722 queries; \S\ref{sec:statistics}). The random baseline is $1/|\mathcal{O}_q|$ per query rather than a flat $25\%$, since option counts range from 2 to 7. A type is marked significant when the Wilson $95\%$ lower bound on its accuracy clears its own baseline.

Seven types, covering $20\%$ of queries, clear their baseline by $+14.6$ to $+38.6$ pp: occlusion recognition, local topology description, ego-motion translation, object-motion separation, obstacle-aware rerouting, reachability judgment, and spatial-relationship inference. Their options describe layouts and outcomes in natural language, which carries everyday plausibility that item-level filtering reduces without eliminating. Eight types sit at or below their baseline, and they include all six streaming-memory types (2.1.1, 2.1.2, 2.2.3, 2.3.2, 3.2.1, 4.3): where the answer exists only in memory of what has left the view, text alone performs no better than guessing. Two of these fall well below baseline (3.2.1 at $9.5\%$ and 1.3.4 at $10.0\%$, both against $25\%$), the expected effect of the construction-time text-only filter (\S\ref{subsec:qc_pipeline}), which removes items answerable from text and leaves survivors whose most plausible-sounding option is wrong. The single map-level family that clears its baseline, spatial-relationship inference at $56.7\%$, still trails the human streaming score on the same family by 30 points ($86.7\%$, Tab.~\ref{tab:overall_results}). Text recovers plausibility, not the answer.
}       

\begin{table*}[t]   
\centering
\revon
\small
\setlength{\tabcolsep}{5pt}
\renewcommand{\arraystretch}{0.92}
\begin{tabular}{l l r r r r}
\toprule
\textbf{Type} & \textbf{Task} & \textbf{N} & \textbf{Text-only (\%)} & \textbf{Random (\%)} & $\Delta$ \textbf{(pp)} \\
\midrule
\multicolumn{6}{l}{\textit{L1: Instantaneous Egocentric Perception}} \\
1.1.1 & Object-to-Object Absolute Distance & 37  & 21.6 & 25.0 & $-3.4$ \\
1.1.2 & Relative Physical Scale           & 49  & 44.9 & 35.7 & $+9.2$ \\
1.1.3 & Passability Affordance            & 30  & 56.7 & 50.0 & $+6.7$ \\
1.1.4 & Viewpoint Height Perception       & 51  & 31.4 & 25.0 & $+6.4$ \\
1.2.1 & Containment Relation              & 65  & 36.9 & 26.3 & $+10.6$ \\
1.2.2 & Occlusion Recognition             & 50  & 40.0 & 25.0 & $+15.0^{*}$ \\
1.2.3 & Support Relation                  & 49  & 28.6 & 25.7 & $+2.9$ \\
1.2.4 & Local Topology Description        & 30  & 60.0 & 25.0 & $+35.0^{*}$ \\
1.3.1 & Ego-motion Translation            & 96  & 31.2 & 16.7 & $+14.6^{*}$ \\
1.3.2 & Ego-motion Rotation               & 70  & 35.7 & 33.3 & $+2.4$ \\
1.3.3 & Object Motion Separation          & 37  & 40.5 & 25.0 & $+15.5^{*}$ \\
1.3.4 & Relative Speed Judgment           & 50  & 10.0 & 25.0 & $-15.0$ \\
1.3.5 & Judgment of Motion State          & 15  & 53.3 & 50.0 & $+3.3$ \\
\midrule
\multicolumn{6}{l}{\textit{L2: Spatiotemporal Context Tracking}} \\
2.1.1 & Revisit Detection                 & 120 & 49.2 & 50.0 & $-0.8$ \\
2.1.2 & Revisit Counting                  & 105 & 21.0 & 24.9 & $-3.9$ \\
2.2.1 & Out-of-View Localization          & 48  & 31.2 & 24.3 & $+6.9$ \\
2.2.2 & Behind-Camera Inference           & 100 & 39.0 & 31.0 & $+8.0$ \\
2.2.3 & Transient Count Memory            & 20  & 20.0 & 25.0 & $-5.0$ \\
2.3.1 & Chronological Recall              & 100 & 39.0 & 32.2 & $+6.8$ \\
2.3.2 & Temporal Depth Estimation         & 47  & 42.6 & 50.0 & $-7.4$ \\
\midrule
\multicolumn{6}{l}{\textit{L3: Generative Spatial Reasoning}} \\
3.1.1 & Mental Rotation \& Fitting        & 50  & 52.0 & 50.0 & $+2.0$ \\
3.1.2 & Spatial What-If Consequence       & 50  & 52.0 & 50.0 & $+2.0$ \\
3.1.3 & Physical Stability \& Feasibility & 50  & 58.0 & 50.0 & $+8.0$ \\
3.2.1 & Object-Level Change Detection     & 63  & 9.5  & 25.0 & $-15.5$ \\
3.3.1 & Optimal Path Reasoning            & 22  & 40.9 & 25.0 & $+15.9$ \\
3.3.2 & Obstacle-Aware Rerouting          & 22  & 63.6 & 25.0 & $+38.6^{*}$ \\
3.3.3 & Reachability Judgment             & 22  & 54.5 & 25.0 & $+29.5^{*}$ \\
\midrule
\multicolumn{6}{l}{\textit{L4: Allocentric Spatial Mapping (three map-query families)}} \\
4.1 & Map Sketching                       & 131 & 35.1 & 30.7 & $+4.4$ \\
4.2 & Spatial Relationship Inference      & 90  & 56.7 & 25.0 & $+31.7^{*}$ \\
4.3 & Trajectory Matching                 & 53  & 20.8 & 25.0 & $-4.2$ \\
\midrule
\multicolumn{2}{l}{\textbf{All}}          & \textbf{1722} & \textbf{37.7} & \textbf{31.7} & \textbf{+6.1} \\
\bottomrule
\end{tabular}
\caption{GPT-5.4 text-only accuracy per canonical task type, computed on the public release (1{,}695 questions scored as 1{,}722 queries). The random baseline is the per-query mean of $1/|\mathcal{O}_q|$. $^{*}$marks types whose Wilson $95\%$ lower bound clears their own baseline. The benchmark-level $+6.1$ pp here and the $+5.8$ pp of Tab.~\ref{tab:overall_results} differ because the latter is computed on the 1{,}680-item evaluation set.}
\label{tab:text_only_per_type}
\end{table*}

\begin{table}[t]
\centering
\small
\setlength{\tabcolsep}{4pt}
\renewcommand{\arraystretch}{0.92}
\resizebox{\columnwidth}{!}{%
\begin{tabular}{l r r r r}
\toprule
\textbf{Benchmark} & \textbf{N} & \textbf{Text-only (\%)} & \textbf{Random (\%)} & $\Delta$ \textbf{(pp)} \\
\midrule
EmbSpatial-Bench~\citep{duan2024embspatial}   & 728  & 67.5 & 25.0 & +42.5 \\
STI-Bench~\citep{sti2025bench}          & 413  & 37.1 & 20.0 & +17.1 \\
SpatialTree-Bench~\citep{xiao2025spatialtree}  & 850  & 32.1 & 21.9 & +10.3 \\
ViewSpatial-Bench~\citep{li2025viewspatial}  & 1143 & 34.1 & 25.9 & +8.2  \\
Mind-Cube~\citep{yin2025spatial}          & 4231 & 39.7 & 32.5 & +7.2  \\
\textbf{OVO-S-Bench (ours)} & \textbf{1680} & \textbf{37.1} & \textbf{31.3} & \textbf{+5.8} \\
OST-Bench~\cite{ost2025bench}          & 1112 & 44.2 & 40.5 & +3.7  \\
ERQA~\citep{yin2025spatial}               & 119  & 27.7 & 26.1 & +1.6  \\
SpatialScore-Hard~\citep{wu2025spatialscore}  & 68   & 30.9 & 30.6 & +0.3  \\
MMSI-Video-Bench~\citep{lin2025mmsivideo}   & 222  & 22.1 & 23.2 & -1.1  \\
MMSI-Bench~\citep{yang2025mmsi}         & 190  & 25.8 & 27.0 & -1.2  \\
RoboBench~\citep{luo2025robobench}          & 106  & 23.6 & 25.0 & -1.4  \\
VSI-Bench~\citep{yang2025thinking}          & 498  & 27.1 & 29.3 & -2.2  \\
\bottomrule
\end{tabular}%
}
\caption{GPT-5.4 text-only accuracy on spatial benchmarks (question and options only, no visual input). $\Delta$ is text-only minus random baseline. OVO-S-Bench row from Tab.~\ref{tab:overall_results}; others from a 20\% stratified sample.}
\label{tab:text_only_cross}
\end{table}

\subsection{Source-Robustness Analysis}
\label{subsec:source_robustness}

\begin{figure}[t]
\centering
\includegraphics[width=\columnwidth]{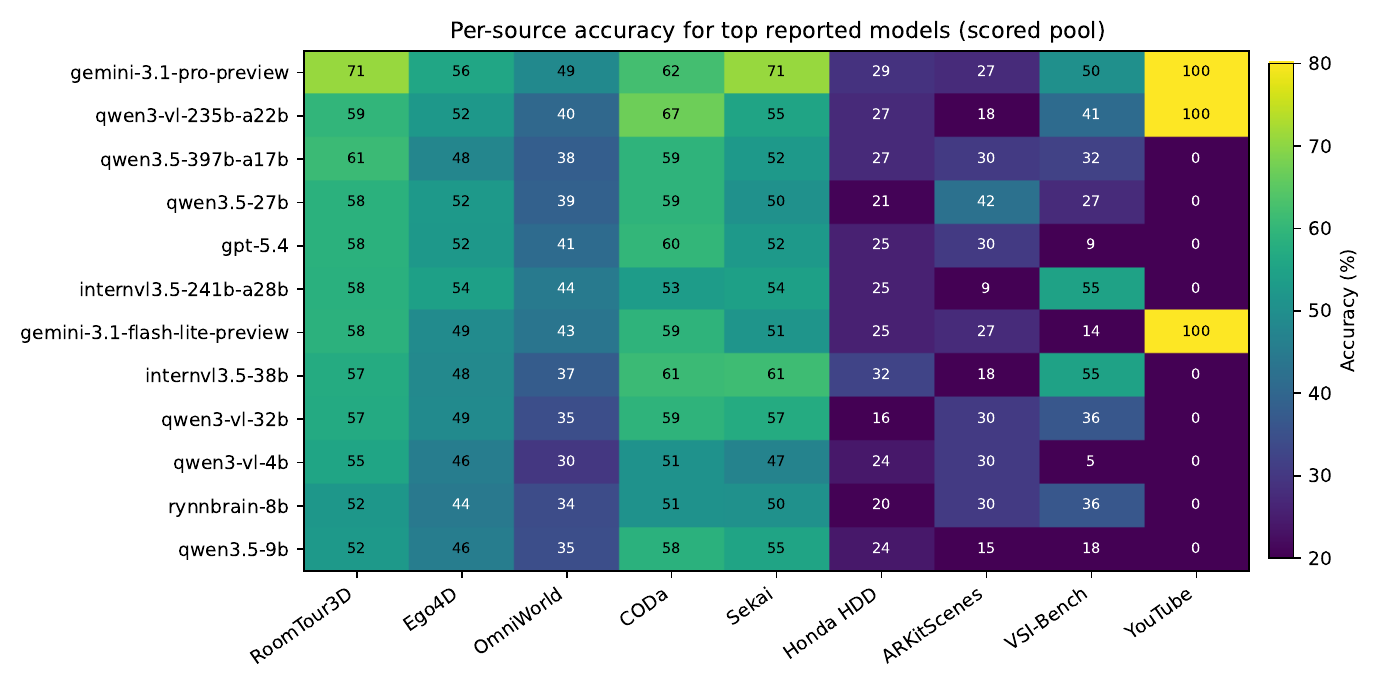}
\caption{Per-source accuracy (\%) for the top-8 reported models. Sources ordered left-to-right by question count (descending).}
\label{fig:source_heatmap}
\end{figure}

RoomTour3D contributes 48\% of benchmark questions (Tab.~\ref{tab:source_datasets}). To test whether this concentration drives the main-table ranking, we recompute overall accuracy after removing all RoomTour3D items. Tab.~\ref{tab:source_robustness} reports the top-10 models under both rankings: the top two (Gemini-3.1-Pro, Qwen3-VL-235B-A22B) hold their positions, and the Spearman rank correlation across all 38 reported systems is $\rho = 0.92$ ($p < 10^{-4}$). Per-source accuracy varies widely (Fig.~\ref{fig:source_heatmap}), reflecting genuine cross-domain difficulty rather than a single-source artifact.

\begin{table}[t]
\centering
\small
\setlength{\tabcolsep}{3pt}
\renewcommand{\arraystretch}{0.92}
\resizebox{\columnwidth}{!}{%
\begin{tabular}{l r r r r r}
\toprule
\textbf{Model} & \textbf{Paper (\%)} & \textbf{R}$_\text{paper}$ & \textbf{w/o RT3D (\%)} & \textbf{R}$_\text{w/o}$ & $\Delta$\textbf{R} \\
\midrule
Gemini-3.1-Pro         & 59.2 & 1  & 54.4 & 1  &  0 \\
Qwen3-VL-235B-A22B    & 53.6 & 2  & 49.7 & 2  &  0 \\
InternVL-3.5-241B     & 50.9 & 6  & 48.8 & 3  & -3 \\
InternVL-3.5-38B      & 49.1 & 8  & 47.9 & 4  & -4 \\
GPT-5.4                & 50.9 & 5  & 47.8 & 5  &  0 \\
Qwen3.5-27B            & 51.7 & 4  & 47.6 & 6  & +2 \\
Gemini-3.1-Flash-Lite  & 50.8 & 7  & 46.8 & 7  &  0 \\
Qwen3.5-397B-A17B     & 52.1 & 3  & 46.0 & 8  & +5 \\
Qwen3-VL-32B           & 48.8 & 9  & 45.7 & 9  &  0 \\
Qwen3.5-9B             & 45.8 & 12 & 43.6 & 10 & +2 \\
\bottomrule
\end{tabular}%
}
\caption{Ranking stability after removing RoomTour3D (48\% of items). R$_\text{paper}$ and R$_\text{w/o}$ are rankings among 38 reported systems. Spearman $\rho = 0.92$ ($p < 10^{-4}$) across all 38 systems.}
\label{tab:source_robustness}
\end{table}

{\revon   
\paragraph{Source $\times$ level matrix.}
Table~\ref{tab:source_level} resolves the heatmap by level, macro-averaging accuracy over the 38 reported systems within each cell and reporting the item count that supports it. Three patterns hold. Difficulty varies widely by source within every level, from 24.0 (Honda HDD) to 52.4 (RoomTour3D) on L4 alone, which locates the difficulty in cross-domain variation rather than in one source. The two L4-only sources occupy the hardest cells in the matrix (Honda HDD 24.0, ARKitScenes 24.2, both near their random baselines), consistent with the overall allocentric-mapping bottleneck. Within a single source the level ordering follows that source's task-type composition, so ``L4 is the lowest level'' is a benchmark-wide statistic (32 of 38 systems) rather than a within-source law, and L4 is the highest level in none of the multi-level sources.
}       

\begin{table}[t]   
\centering
\revon
\small
\setlength{\tabcolsep}{3pt}
\renewcommand{\arraystretch}{0.92}
\resizebox{\columnwidth}{!}{%
\begin{tabular}{l cc cc cc cc}
\toprule
\multirow{2}{*}{\textbf{Source}} & \multicolumn{2}{c}{\textbf{L1}} & \multicolumn{2}{c}{\textbf{L2}} & \multicolumn{2}{c}{\textbf{L3}} & \multicolumn{2}{c}{\textbf{L4}} \\
\cmidrule(lr){2-3}\cmidrule(lr){4-5}\cmidrule(lr){6-7}\cmidrule(lr){8-9}
 & N & \% & N & \% & N & \% & N & \% \\
\midrule
RoomTour3D    & 174 & 43.6 & 308 & 48.7 & 163 & 55.1 & 159 & 52.4 \\
Ego4D         & 166 & 43.1 & 113 & 47.4 & 74  & 31.1 & 9   & 38.0$^{\dagger}$ \\
OmniWorld     & 111 & 32.1 & 22  & 39.1 & 8   & 33.9$^{\dagger}$ & 12 & 33.8 \\
CODa          & 112 & 43.3 & 17  & 59.1 & 17  & 64.6 & 2  & 38.2$^{\dagger}$ \\
Sekai         & 32  & 46.3 & 53  & 42.0 & --  & --         & -- & -- \\
VSI-Bench     & 19  & 44.8 & --  & --         & --  & --         & -- & -- \\
YouTube tours & --  & --         & --  & --         & 17  & 54.9 & -- & -- \\
Honda HDD     & --  & --         & --  & --         & --  & --         & 59 & 24.0 \\
ARKitScenes   & --  & --         & --  & --         & --  & --         & 33 & 24.2 \\
\midrule
\textbf{All}  & \textbf{614} & & \textbf{513} & & \textbf{279} & & \textbf{274} & \\
\bottomrule
\end{tabular}%
}
\caption{Source $\times$ level matrix over the 1{,}680-item evaluation set. \% is accuracy macro-averaged over the 38 reported systems within the cell; N is the item count. $^{\dagger}$marks cells with fewer than 10 items. Row and column totals match Tab.~\ref{tab:source_datasets} and Fig.~\ref{fig:taxonomy}d respectively.}
\label{tab:source_level}
\end{table}

{\revon   
\paragraph{Release-date probe.}
Familiarity with a source video would surface as a consistent advantage on sources released before typical training cutoffs. We compare pre-2023 sources (Ego4D) against 2025 releases (OmniWorld, Sekai) within each task type, which controls for task composition; ARKitScenes and Honda HDD carry L4 items only and hold no paired types, so the comparison reduces to egocentric-indoor against outdoor world scenes. Table~\ref{tab:release_date_probe} reports the 16 task types that keep at least 5 items in both groups. Per-type differences flip sign across the range $-17$ to $+21$ pp, with a residual mean of $+5.0$ pp that tracks the indoor--outdoor domain split rather than release year: the largest positive gap falls on relative speed judgment, where near-field egocentric motion is easier to read than outdoor walking and driving, and the largest negative gap falls on temporal depth estimation, where long outdoor traversals make elapsed time easier to estimate. Varying the minimum cell size from 3 to 10 items retains 19 to 11 types and moves the mean between $+4.6$ and $+7.2$ pp, leaving the sign pattern unchanged. No configuration produces the uniform pre-2023 advantage that memorization would predict.
}       

\begin{table}[t]   
\centering
\revon
\small
\ifdefined\arxivversion
\setlength{\tabcolsep}{3pt}
\renewcommand{\arraystretch}{0.95}
\footnotesize
\resizebox{\columnwidth}{!}{%
\begin{tabular}{l *{16}{c} @{\hspace{8pt}} c}
\toprule
\textbf{Type} & 1.1.2 & 1.1.4 & 1.2.1 & 1.2.2 & 1.2.3 & 1.3.1 & 1.3.2 & 1.3.3 & 1.3.4 & 2.1.1 & 2.1.2 & 2.2.2 & 2.3.1 & 2.3.2 & 3.2.1 & 4.2 & \textbf{Mean} \\
\midrule
pre-2023 \textbf{N}  & 12 & 14 & 18 & 15 & 13 & 22 & 19 & 10 & 16 & 20 & 16 & 14 & 15 & 9 & 18 & 6 & \\
pre-2023 \textbf{\%} & 49.0 & 35.1 & 43.9 & 42.0 & 35.1 & 46.3 & 42.5 & 39.1 & 51.1 & 55.4 & 53.3 & 38.1 & 42.8 & 39.4 & 39.1 & 44.8 & \\
2025 \textbf{N}      & 10 & 11 & 14 & 13 & 9 & 18 & 16 & 8 & 12 & 16 & 12 & 10 & 12 & 7 & 6 & 7 & \\
2025 \textbf{\%}     & 35.0 & 39.1 & 34.9 & 34.0 & 41.1 & 34.3 & 35.5 & 36.1 & 30.1 & 39.4 & 48.3 & 49.1 & 40.8 & 56.4 & 31.1 & 31.8 & \\
\midrule
$\Delta$ \textbf{(pp)} & $+14.0$ & $-4.0$ & $+9.0$ & $+8.0$ & $-6.0$ & $+12.0$ & $+7.0$ & $+3.0$ & $+21.0$ & $+16.0$ & $+5.0$ & $-11.0$ & $+2.0$ & $-17.0$ & $+8.0$ & $+13.0$ & $\mathbf{+5.0}$ \\
\bottomrule
\end{tabular}%
}
\else
\setlength{\tabcolsep}{3.5pt}
\renewcommand{\arraystretch}{0.92}
\begin{tabular}{l cc cc r}
\toprule
\multirow{2}{*}{\textbf{Type}} & \multicolumn{2}{c}{\textbf{pre-2023}} & \multicolumn{2}{c}{\textbf{2025}} & \multirow{2}{*}{$\Delta$ \textbf{(pp)}} \\
\cmidrule(lr){2-3}\cmidrule(lr){4-5}
 & N & \% & N & \% & \\
\midrule
1.1.2 & 12 & 49.0 & 10 & 35.0 & $+14.0$ \\
1.1.4 & 14 & 35.1 & 11 & 39.1 & $-4.0$ \\
1.2.1 & 18 & 43.9 & 14 & 34.9 & $+9.0$ \\
1.2.2 & 15 & 42.0 & 13 & 34.0 & $+8.0$ \\
1.2.3 & 13 & 35.1 & 9 & 41.1 & $-6.0$ \\
1.3.1 & 22 & 46.3 & 18 & 34.3 & $+12.0$ \\
1.3.2 & 19 & 42.5 & 16 & 35.5 & $+7.0$ \\
1.3.3 & 10 & 39.1 & 8 & 36.1 & $+3.0$ \\
1.3.4 & 16 & 51.1 & 12 & 30.1 & $+21.0$ \\
2.1.1 & 20 & 55.4 & 16 & 39.4 & $+16.0$ \\
2.1.2 & 16 & 53.3 & 12 & 48.3 & $+5.0$ \\
2.2.2 & 14 & 38.1 & 10 & 49.1 & $-11.0$ \\
2.3.1 & 15 & 42.8 & 12 & 40.8 & $+2.0$ \\
2.3.2 & 9 & 39.4 & 7 & 56.4 & $-17.0$ \\
3.2.1 & 18 & 39.1 & 6 & 31.1 & $+8.0$ \\
4.2 & 6 & 44.8 & 7 & 31.8 & $+13.0$ \\
\midrule
\multicolumn{5}{l}{\textbf{Mean} $\Delta$ over 16 types} & \textbf{$+5.0$} \\
\bottomrule
\end{tabular}
\fi
\caption{Release-date probe. Accuracy macro-averaged over the 38 reported systems within each (task type, source group) cell, for the 16 types holding at least 5 items in both groups. pre-2023 is Ego4D; 2025 is OmniWorld and Sekai. $\Delta$ is pre-2023 minus 2025.}
\label{tab:release_date_probe}
\end{table}

\section{Evaluation Methods Details}
\label{sec:appendix_eval}
This appendix specifies the inference configuration used to produce every score in Section~\ref{sec:experiments}. We give the model inventory and provenance (\S\ref{subsec:eval_models}), the decoding-parameter recipes (\S\ref{subsec:eval_decoding}), the default frame-sampling policy (\S\ref{subsec:eval_sampling}), the streaming-model ingestion protocol (\S\ref{subsec:eval_streaming}), the prompt templates used at the model interface (\S\ref{subsec:eval_prompts}), the verbatim policy-specific prompts used in the \S\ref{sec:appendix_sampling} matrix (\S\ref{subsec:eval_policy_prompts}), the deterministic answer-extraction rule (\S\ref{subsec:eval_parsing}), and the hardware and software stack (\S\ref{subsec:eval_runtime}). The governing principle throughout is that every per-model choice (resolution, frame count, decoding temperature, reasoning-mode toggle) follows the recipe published by the model's authors; nothing is tuned on OVO-S.

\subsection{Models Evaluated}
\label{subsec:eval_models}
The set of evaluated models is listed in Table~\ref{tab:overall_results} (the main results table, including closed-source proprietary MLLMs, open-source general video MLLMs, streaming video MLLMs, token-compression / memory-based methods, spatially fine-tuned MLLMs, and embodied foundation models). Open-source models are loaded from each author's official HuggingFace release; no weights, tokenizer, or vocabulary are modified, and inference is driven through the authors' released code path (vLLM for models whose card lists vLLM as a supported backend; the authors' provided inference scripts otherwise; see \S\ref{subsec:eval_runtime}). Closed-source models are accessed through their providers' OpenAI-compatible endpoints. To allow re-evaluation against any future API revision, we record the access window: all closed-source runs reported in this paper were collected between 2026-02 and 2026-04 against the providers' \texttt{stable} (non-preview) endpoint tier where one exists, and against the latest preview tier otherwise (\texttt{gemini-3.1-pro-preview}, \texttt{gemini-3.1-flash-lite-preview}). The exact endpoint identifier is the model-name string shown in Table~\ref{tab:overall_results}.

\subsection{Decoding Parameters}
\label{subsec:eval_decoding}
\paragraph{Non-reasoning baseline.}
Every model whose row in Table~\ref{tab:overall_results} does not carry a thinking-mode toggle is run under the same decoding protocol: greedy decoding ($T{=}0$), no nucleus filtering, and a $1{,}024$-token output cap. The cap is comfortably larger than the answer-letter response targeted by our prompt (\S\ref{subsec:eval_prompts}) and accommodates models that emit a short justification before the letter.

\paragraph{Reasoning variants.}
For the reasoning-mode rows of Table~\ref{tab:thinking}, decoding follows the recipe published by each model's authors verbatim. Table~\ref{tab:eval_recipes} reproduces these recipes. Two design choices in the table merit note. First, the non-zero $T{=}0.6$ and $\text{top-}p{=}0.95$ used for InternVL-3.5-thinking are the authors' R1-style anti-repetition recipe documented on the model card; the same model card prescribes the chain-of-thought user template paired with the R1 system prompt, which is why InternVL-3.5-thinking is the only row that overrides the default prompt template (\S\ref{subsec:eval_prompts}). Second, the per-family thinking-token cap (8K--32K) is the published thinking budget for each family; we observed truncation only on Qwen3.5-thinking when the trace exceeds 32K (591 traces across L1--L4, all tagged \texttt{T1a} per \S\ref{subsec:cot_setup}), which is recorded as a model-side failure rather than a configuration artifact.

\begin{table}[t]
\small
\centering
\setlength{\tabcolsep}{4pt}
\resizebox{\columnwidth}{!}{%
\begin{tabular}{l c c c r}
\toprule
\textbf{Family} & $T$ & top-$p$ & \textbf{Thinking toggle} & \textbf{Max tok} \\
\midrule
Qwen3-VL-thinking & 0.0 & -- & \texttt{enable\_thinking=true} & 8{,}192 \\
Qwen3.5-thinking & 0.0 & -- & \texttt{enable\_thinking=true} & 32{,}768 \\
InternVL-3.5-thinking & 0.6 & 0.95 & R1 system prompt + CoT user template & 16{,}384 \\
GLM-4.6V-Flash-thinking & 0.0 & -- & \texttt{chat\_template\_kwargs.enable\_thinking=true} & 16{,}384 \\
Grok-4.1-Fast-Reasoning & 0.0 & -- & built-in (no user toggle) & 4{,}096 \\
\bottomrule
\end{tabular}%
}
\caption{Reasoning-mode decoding recipes for the thinking-variant rows of Tab.~\ref{tab:thinking}. Each row reproduces the recipe documented on the corresponding model card. \emph{Max tok} caps the thinking trace plus answer combined; truncation is tagged \texttt{T1a} (see \S\ref{subsec:cot_setup}).}
\label{tab:eval_recipes}
\end{table}

\subsection{Default Frame Sampling}
\label{subsec:eval_sampling}
For each query at timestamp $t_q$ the default visual input is $N{=}128$ frames sampled uniformly across the prefix $[0, t_q]$. The choice of $N$ is the largest uniform-frame budget that fits inside every evaluated model's published context window without truncation; Section~\ref{sec:appendix_sampling} reports the sensitivity of overall and per-level accuracy to this choice. Frames are resized so the shorter side equals the spatial resolution prescribed by each model's training recipe (this falls in the $336$--$512$\,px range across the evaluation set), and the longer side is preserved up to the model's published per-frame visual-token cap where one exists (\textit{e.g.}\ Qwen3-VL's \texttt{max\_pixels} setting, Flash-VStream's $50{,}176$-pixel cap). Frame extraction is cached: the decoded $N$-frame tensor is keyed by the SHA-1 of the video file and the sampling specification, so a re-run of the same model on the same annotation reads from cache and is bit-for-bit identical to the first run.

\subsection{Streaming-Model Ingestion Protocol}
\label{subsec:eval_streaming}
\paragraph{Native protocol.}
Streaming-architecture models (StreamingVLM, StreamForest, Flash-VStream in the main results; HERMES, InfiniPot-V, FluxMem, StreamingTOM in the token-compression block) consume the video as a sequential stream and apply a model-side memory or KV-compression mechanism that is calibrated for a specific ingest rate. We therefore do \emph{not} impose the uniform $128$-frame default on these models for their main-result row in Table~\ref{tab:overall_results}; instead the video is read from the start until $t_q$ at the model's published streaming rate, and answers are generated using the resulting compressed state. Table~\ref{tab:eval_streaming_protocol} lists, for each streaming model, the ingest rate and the chunk or compression schedule we use. Every entry is the value provided by the model's authors (the model card, the released inference script, or the corresponding paper); no streaming hyperparameter was tuned on OVO-S, and the compression-side hyperparameters (\textit{e.g.}\ StreamingTOM's contrastive-token retention thresholds) take the authors' released defaults.

\begin{table}[t]
\small
\centering
\setlength{\tabcolsep}{4pt}
\resizebox{\columnwidth}{!}{%
\begin{tabular}{l c l}
\toprule
\textbf{Model} & \textbf{Ingest rate} & \textbf{Memory / compression unit} \\
\midrule
StreamingVLM & 2\,fps & continuous (linear-attention context) \\
StreamForest & 1\,fps & $64$-frame budget \\
Flash-VStream & 2\,fps & continuous, $50{,}176$-pixel per-frame cap \\
HERMES & 2\,fps & $16$-frame chunk, KV pruned per chunk \\
InfiniPot-V & 1\,fps & KV-budget per chunk \\
FluxMem & 1\,fps & flux-memory replacement schedule \\
StreamingTOM & 1\,fps & $32$-frame encoder batch, contrastive-token retention \\
\bottomrule
\end{tabular}%
}
\caption{Streaming ingest rate and compression schedule for each streaming-architecture model. All values from the corresponding paper, model card, or released inference script; no streaming hyperparameter is tuned on OVO-S.}
\label{tab:eval_streaming_protocol}
\end{table}

\subsection{Prompt Templates}
\label{subsec:eval_prompts}
\paragraph{Default template.}
With the exception of the InternVL-3.5-thinking row (see below), every model in every evaluation track receives the same prompt template. The question text and option strings are substituted into the placeholders verbatim:
\begin{quote}\small\ttfamily
You are evaluating a video understanding task. Based on the video frames provided, answer the following multiple choice question.\\[2pt]
Question: \{question\}\\[2pt]
Options:\\
\{options\_text\}\\[2pt]
Instructions:\\
- Respond with ONLY the letter of your answer (e.g., ``A'' or ``B'').\\
- Do not include any explanation or additional text.\\[2pt]
Your answer:
\end{quote}

\paragraph{Chain-of-thought template (reasoning models with a CoT recipe).}
InternVL-3.5-thinking is the only main-text row whose model card prescribes an explicit chain-of-thought user template alongside the R1 system prompt; for that row we replace the \emph{Instructions} block above with the following, while leaving the question and option blocks unchanged:
\begin{quote}\small\ttfamily
Instructions:\\
- Think step by step about the spatial relationships and scene content shown in the frames.\\
- After your reasoning, output your final answer on a new line in the format: Answer: X
\end{quote}
\noindent All other thinking-mode rows (Qwen3-VL-thinking, Qwen3.5-thinking, GLM-4.6V-Flash-thinking, Grok-4.1-Fast-Reasoning) use the default template above; the reasoning trace is gated by the model-side thinking-mode flag listed in Table~\ref{tab:eval_recipes}, not by the user prompt.

\subsection{Policy-Specific Inputs}
\label{subsec:eval_policy_prompts}
A practical clarification on the matrix in \S\ref{sec:appendix_sampling}: the eight frame-sampling policies (\emph{single@query}, \emph{nearest-16f@4fps}, \emph{uniform-32}, \emph{uniform-64}, \emph{uniform-128}, \emph{uniform-256}, \emph{log-decay-128}, \emph{oracle-evidence}) differ \emph{only} in the visual input fed to the model, not in the textual prompt. Every policy submits exactly the same multiple-choice prompt produced by the template above, with the same question and option strings. The \emph{text-only} control baseline reported in the \emph{Baseline \& Controls} block of Table~\ref{tab:overall_results} is the same prompt submitted with an empty image list, and is evaluated under GPT-5.4 to bound how much of the benchmark is answerable from textual cues alone.

\subsection{Answer Extraction}
\label{subsec:eval_parsing}
We adopt multiple-choice scoring throughout because it enables deterministic, reproducible evaluation without relying on LLM-as-judge for open-ended answers, which would itself introduce the spatial grounding errors this benchmark is designed to measure.

Model outputs are scored by a deterministic post-processor. For thinking-mode outputs we first remove any \texttt{<think>\ldots</think>} span so that subsequent matching applies only to the post-thinking tail; outputs without a closing \texttt{</think>} after the published thinking budget are treated as truncated and tagged \texttt{T1a} per \S\ref{subsec:cot_setup}. The extractor then tries the following patterns in order on the (de-thinked) response, restricted to the letter range \texttt{A}--\texttt{G}, and returns the first match:
\begin{enumerate}\itemsep0pt
\item Within the last $300$ characters: a tail \texttt{Answer:\,X} (including \emph{final answer}, \emph{final}).
\item Within the last $300$ characters: a Cosmos-style \texttt{<answer>\,X} tag.
\item Within the last $300$ characters: a bare single letter at the very end (\textit{e.g.}\ ``\ldots therefore B'').
\item Within the last $300$ characters: a GLM-style \texttt{<|begin\_of\_box|>\,X} marker.
\item A single letter at the start of the stripped response.
\item Anywhere in the response: \texttt{answer\,/\,choice\,/\,option(s):\,X} with a true separator.
\item Anywhere in the response: a parenthesized or bracketed letter, \textit{e.g.}\ ``(B)'' or ``[C]''.
\end{enumerate}
\noindent Responses for which none of these patterns matches are scored as wrong; they are not skipped, and they contribute to the denominator of every per-task and overall accuracy in Section~\ref{sec:experiments}. The order above is the tail-first order that minimizes accidental matches against letters embedded in reasoning prose; the only step that scans the full response is step (vi), which requires the trigger word \emph{answer}, \emph{choice}, or \emph{option(s)} followed by a real separator.

\subsection{Hardware and Software}
\label{subsec:eval_runtime}
All open-source inference is run on NVIDIA H-series GPUs (H800 80\,GB and H200 141\,GB). Models whose authors release a vLLM-compatible inference path are served with vLLM $\geq{}0.11$ in offline batched mode; models without such a path (predominantly the streaming-architecture and token-compression entries) are served through the authors' released inference scripts at the dependency versions pinned in those repositories. All closed-source inference goes through the providers' OpenAI-compatible HTTPS endpoints. Video frame decoding uses \texttt{decord} as the primary backend with an OpenCV fallback for codecs unsupported by \texttt{decord}. The decoded $N$-frame tensor for each (model, query, sampling specification) tuple is cached on disk under a content-addressed key (the SHA-1 of the video file together with the sampling specification), so repeated evaluation against the same query reads from cache and produces a bit-for-bit identical visual input; the cache footprint for the full benchmark is on the order of $50$\,GB.

{\revon   
\section{Response-Type Decomposition}
\label{sec:appendix_response_type}

OVO-Bench~\citep{ovo2025bench} separates \emph{when} a model should answer into three regimes: \emph{Backward} (the answer lies in past observation), \emph{Real-Time} (the answer lies in the current view), and \emph{Forward} (the model must decide when to speak). OVO-S-Bench instantiates the first two with spatial content and states Forward as a non-goal (Section~\ref{sec:intro}). This appendix quantifies that scope so the claim is checkable rather than asserted.

\paragraph{Labeling rule.}
Let $t_q$ be the query timestamp, $e_q$ the end of the annotated evidence interval, and $g_q = t_q - e_q$ the evidence-to-query gap. Three of the four levels are determined by the level itself and the fourth by $g_q$:
\begin{itemize}
\itemsep0em
\item L1 is \emph{Real-Time}: the evidence interval brackets the query and the answer is readable from the current view.
\item L2 and L4 are \emph{Backward}: the answer requires an observation that has left the view (L2) or an integration over the entire explored region (L4).
\item L3 is \emph{Real-Time} when $g_q \leq \tau$ and \emph{Backward} otherwise, with $\tau = 2.0\,$s.
\end{itemize}
L3 is the only level carrying both labels: its median evidence span equals L1's ($2.0$\,s, \S\ref{sec:statistics}) while a minority of its items reference earlier observation. The threshold matches that median, so evidence ending within $2$\,s of the query counts as still in view.

\paragraph{Counts and threshold sensitivity.}
Table~\ref{tab:response_type} gives the decomposition and its sensitivity to $\tau$. Sweeping $\tau$ from $0.5$ to $10$\,s moves the overall split by 2.0 points (50.3\% to 52.3\% Real-Time) and reassigns 33 of 1{,}722 queries in total, because 85\% of L3 gaps fall below $0.5$\,s. The decomposition is therefore a property of the taxonomy rather than of the threshold.

\begin{table}[t]   
\centering
\revon
\small
\setlength{\tabcolsep}{5pt}
\renewcommand{\arraystretch}{0.92}
\begin{tabular}{l r r}
\toprule
\textbf{Level} & \textbf{Real-Time} & \textbf{Backward} \\
\midrule
L1 & 629 & 0 \\
L2 & 0   & 540 \\
L3 & 253 & 26 \\
L4 & 0   & 274 \\
\midrule
\textbf{Total} & \textbf{882} (51\%) & \textbf{840} (49\%) \\
\bottomrule
\end{tabular}

\vspace{4pt}
\begin{tabular}{l r r r}
\toprule
$\tau$ \textbf{(s)} & \textbf{L3 RT} & \textbf{L3 BW} & \textbf{Overall RT (\%)} \\
\midrule
0.5  & 238 & 41 & 50.3 \\
1.0  & 248 & 31 & 50.9 \\
\textbf{2.0} & \textbf{253} & \textbf{26} & \textbf{51.2} \\
3.0  & 257 & 22 & 51.5 \\
5.0  & 263 & 16 & 51.8 \\
10.0 & 271 & 8  & 52.3 \\
\bottomrule
\end{tabular}
\caption{Response-type decomposition over the 1{,}722 scored queries of the public release (top) and its sensitivity to the L3 gap threshold $\tau$ (bottom). The row in bold is the setting reported in Section~\ref{sec:setup}.}
\label{tab:response_type}
\end{table}

\section{Open-Ended Answering Pilot}
\label{sec:appendix_open_ended}

Multiple-choice scoring keeps every reported number deterministic and reproducible, and it compresses partial spatial knowledge into a single letter. We measure both effects on a stratified sample of 200 items (50 per level), with GPT-5.4 answering in free form under the same frame input as the main protocol.

\paragraph{Scoring.}
An independent judge (Gemini-3.1-Pro) scores each free-form answer against the gold spatial fact on a three-point scale: 2 when the key relation, region, count, or layout matches; 1 when the relation or region is right but the specifics are imprecise; 0 when the answer contradicts the gold fact or is too vague to recover it. A second judge agrees with the first on the binary fully-correct decision for 96\% of items. Prompts and both judges' scores are released with the code.

\paragraph{The two distortions run in opposite directions.}
Strict open-ended accuracy is 45.0 against 55.5 under multiple choice on the same items: options scaffold answers the model cannot produce unaided. In the other direction, 20\% of free-form answers are partially correct, and among items failed under multiple choice, 37\% still carry partially or fully correct spatial content in free form. One trace identifies that the pillows lie ``side by side in rows'' while missing their global triangular arrangement, which the letter-level score cannot distinguish from a model that has no layout at all. Awarding half credit for partial answers yields 55.0 against 55.5, so the two distortions cancel in aggregate while the item-level information they carry is recoverable only under graded scoring.

\paragraph{Level breakdown.}
Table~\ref{tab:open_ended} reports per-level counts. The ordering of multiple-choice accuracy across levels follows GPT-5.4's full-benchmark profile (Tab.~\ref{tab:overall_results}), with L4 lowest. These 200 items score 55.5 under multiple choice against GPT-5.4's $50.9$ over the full benchmark, so the sample is easier than the benchmark average and the pilot supports within-sample comparisons.
}       

\begin{table}[t]   
\centering
\revon
\small
\setlength{\tabcolsep}{4pt}
\renewcommand{\arraystretch}{0.92}
\begin{tabular}{l r r r r r r}
\toprule
\multirow{2}{*}{\textbf{Level}} & \multicolumn{3}{c}{\textbf{Open-ended}} & \multirow{2}{*}{\textbf{Strict}} & \multirow{2}{*}{\textbf{Graded}} & \multirow{2}{*}{\textbf{MCQ}} \\
\cmidrule(lr){2-4}
 & full & part & none & & & \\
\midrule
L1 & 26 & 9  & 15 & 52.0 & 61.0 & 60.0 \\
L2 & 28 & 9  & 13 & 56.0 & 65.0 & 64.0 \\
L3 & 21 & 12 & 17 & 42.0 & 54.0 & 56.0 \\
L4 & 15 & 10 & 25 & 30.0 & 40.0 & 42.0 \\
\midrule
\textbf{All} & \textbf{90} & \textbf{40} & \textbf{70} & \textbf{45.0} & \textbf{55.0} & \textbf{55.5} \\
\bottomrule
\end{tabular}
\caption{Open-ended pilot on 200 stratified items (50 per level). \emph{full}/\emph{part}/\emph{none} are judge scores 2/1/0. \emph{Strict} counts only full credit; \emph{Graded} awards $0.5$ for partial; \emph{MCQ} is the multiple-choice score on the same items.}
\label{tab:open_ended}
\end{table}

\section{Future Work}
\label{sec:future_work}

{\revon   
\paragraph{Scaling annotation with humans in the loop.}
Every item in this release is written by hand, which is what buys verified query timestamps, verified evidence intervals, and shortcut screening, and it is also what caps the scale at 1{,}680 items and 804 person-hours. The route to a larger release keeps the human as verifier rather than author: a model proposes candidate question--option--answer tuples, and annotators run the same blind double review that governs the current pipeline (\S\ref{subsec:qc_pipeline}). Adapting an automated spatial-supervision pipeline such as Holi-Spatial~\citep{gong2026holispatial} to emit streaming annotations, including a query timestamp and an evidence interval per item, would supply those candidates directly.

\paragraph{From diagnosis to remedy.}
The compression diagnostics locate the L4 failure before reasoning begins: evidence recall on L4 falls to $0.14$--$0.42$ against $0.60$--$0.76$ on L1 (Section~\ref{sec:method_vs_base}), so an intervention has to change which spatial evidence survives, not how the surviving tokens are read. Two families of remedy follow directly. Explicit spatial memory keeps an allocentric structure updated as the stream arrives, through 3D-aware memory, test-time adaptation, world-model imagination, or structured maps. External geometric grounding supplies at inference what the representation lacks, through depth, open-vocabulary detection, or reconstruction tools invoked by an agent, as in Think3D~\citep{zhang2026think3d}; OVO-S-Bench measures such interventions per level, which is where a grounding remedy should show its effect. The open question our diagnostics sharpen is when to invoke such tools, since always-on tool use adds scaffolding cost on the levels a model already handles.

\paragraph{Evaluation protocol.}
Graded open-ended scoring (Appendix~\ref{sec:appendix_open_ended}), richer human baselines, and interactive embodied tasks each recover information the current protocol compresses away. We also plan a contamination-aware release protocol, with a lightweight public subset and a protected evaluation split if the benchmark is used for leaderboard evaluation.
}       

\section{Additional Examples}
\label{sec:appendix_examples}

The following pages present representative examples across all levels and task families of OVO-S-Bench. Each page pairs streaming spatial questions with visual evidence, query timestamps, and answers.

\begin{figure*}[p]
\centering
\includegraphics[page=1,width=\textwidth,height=0.92\textheight,keepaspectratio]{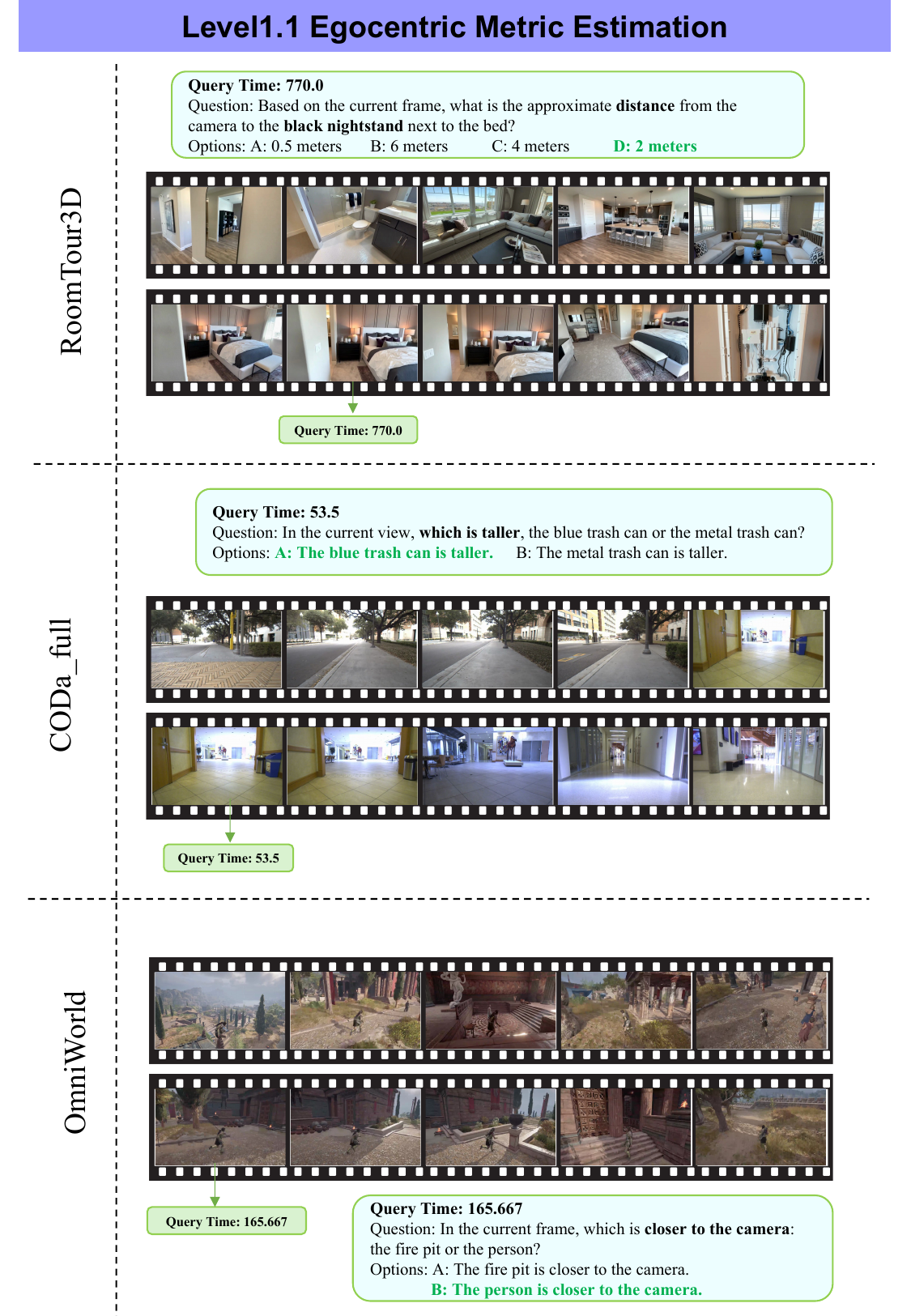}
\end{figure*}

\begin{figure*}[p]
\centering
\includegraphics[page=2,width=\textwidth,height=0.92\textheight,keepaspectratio]{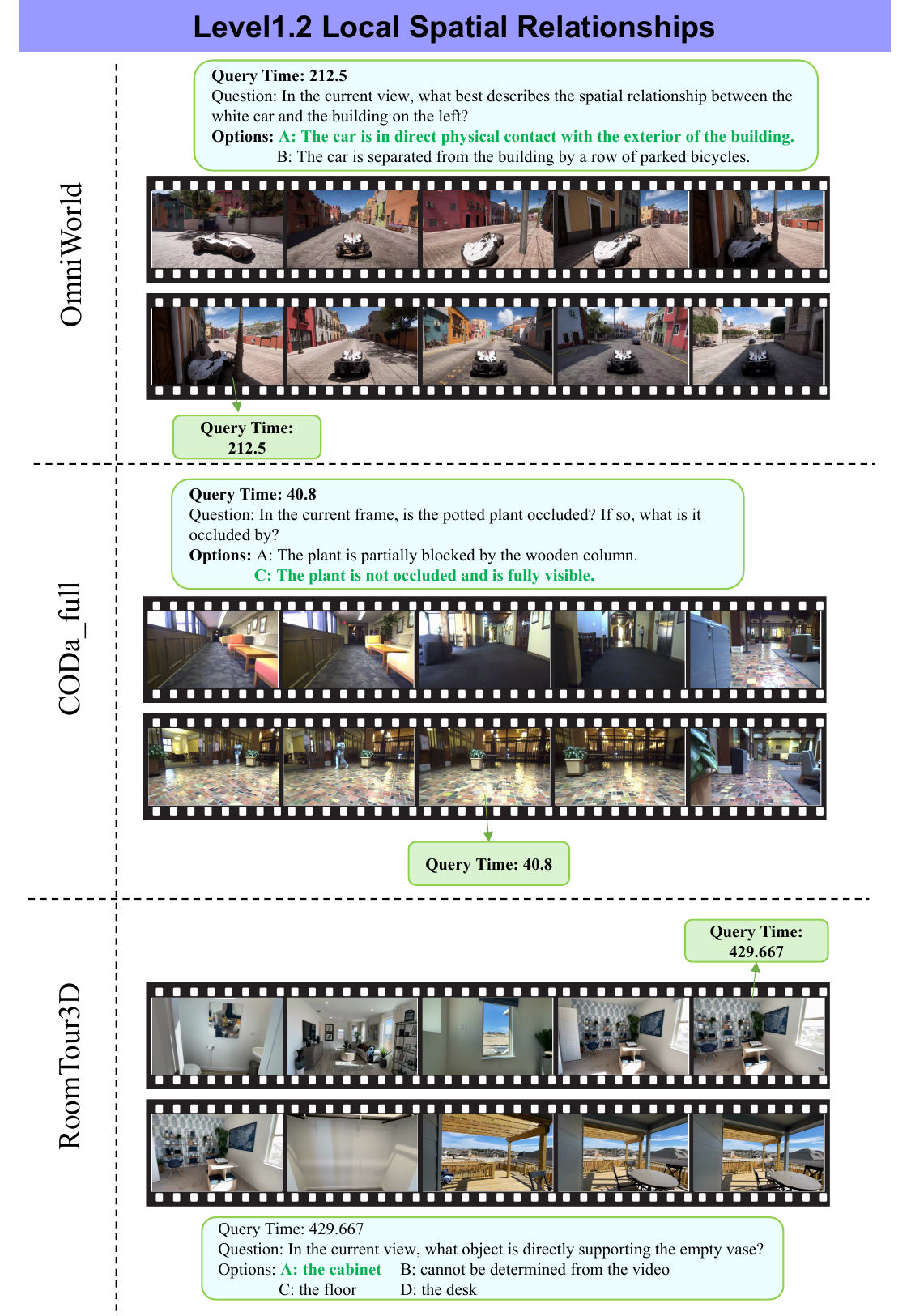}
\end{figure*}

\begin{figure*}[p]
\centering
\includegraphics[page=3,width=\textwidth,height=0.92\textheight,keepaspectratio]{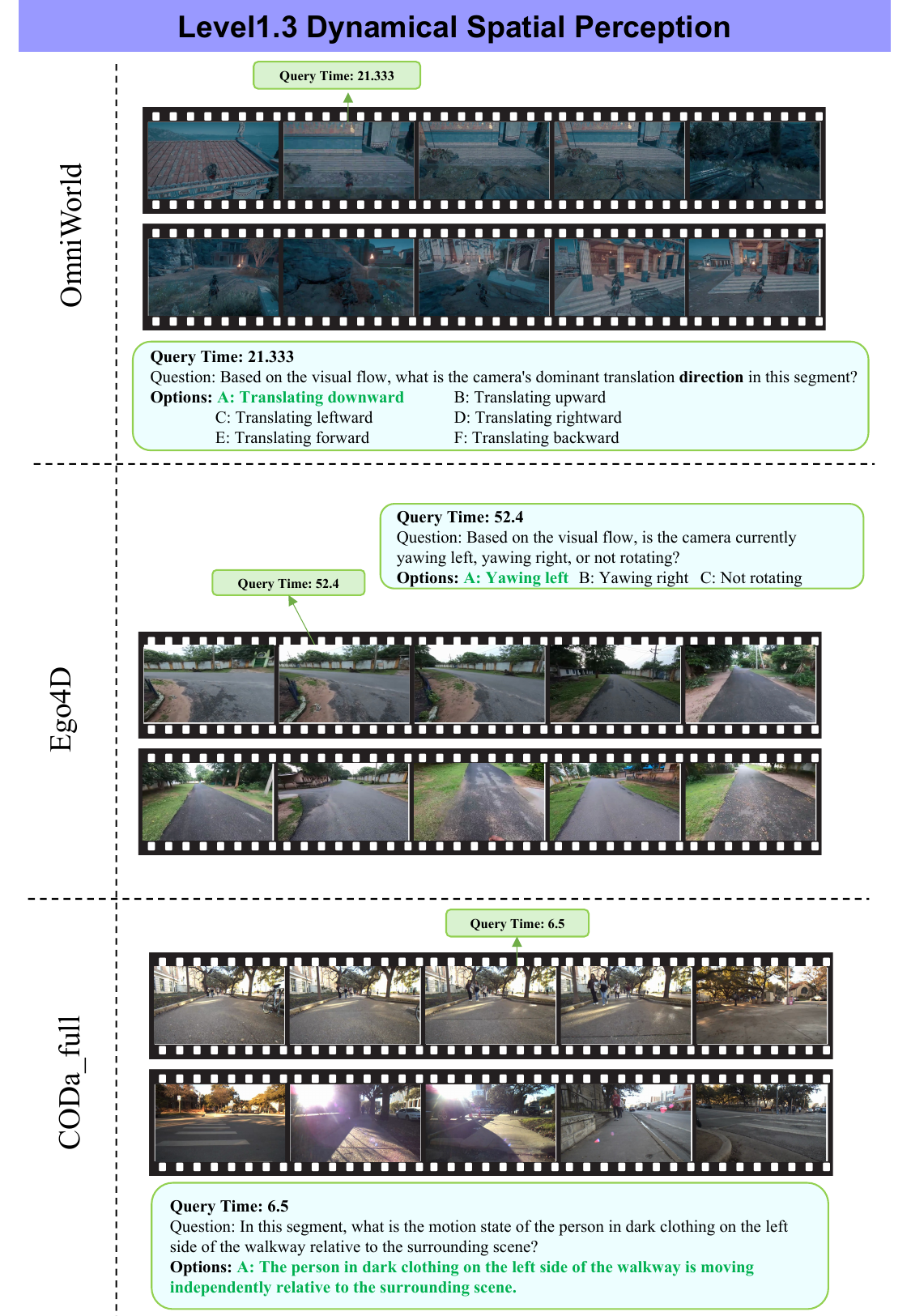}
\end{figure*}

\begin{figure*}[p]
\centering
\includegraphics[page=4,width=\textwidth,height=0.92\textheight,keepaspectratio]{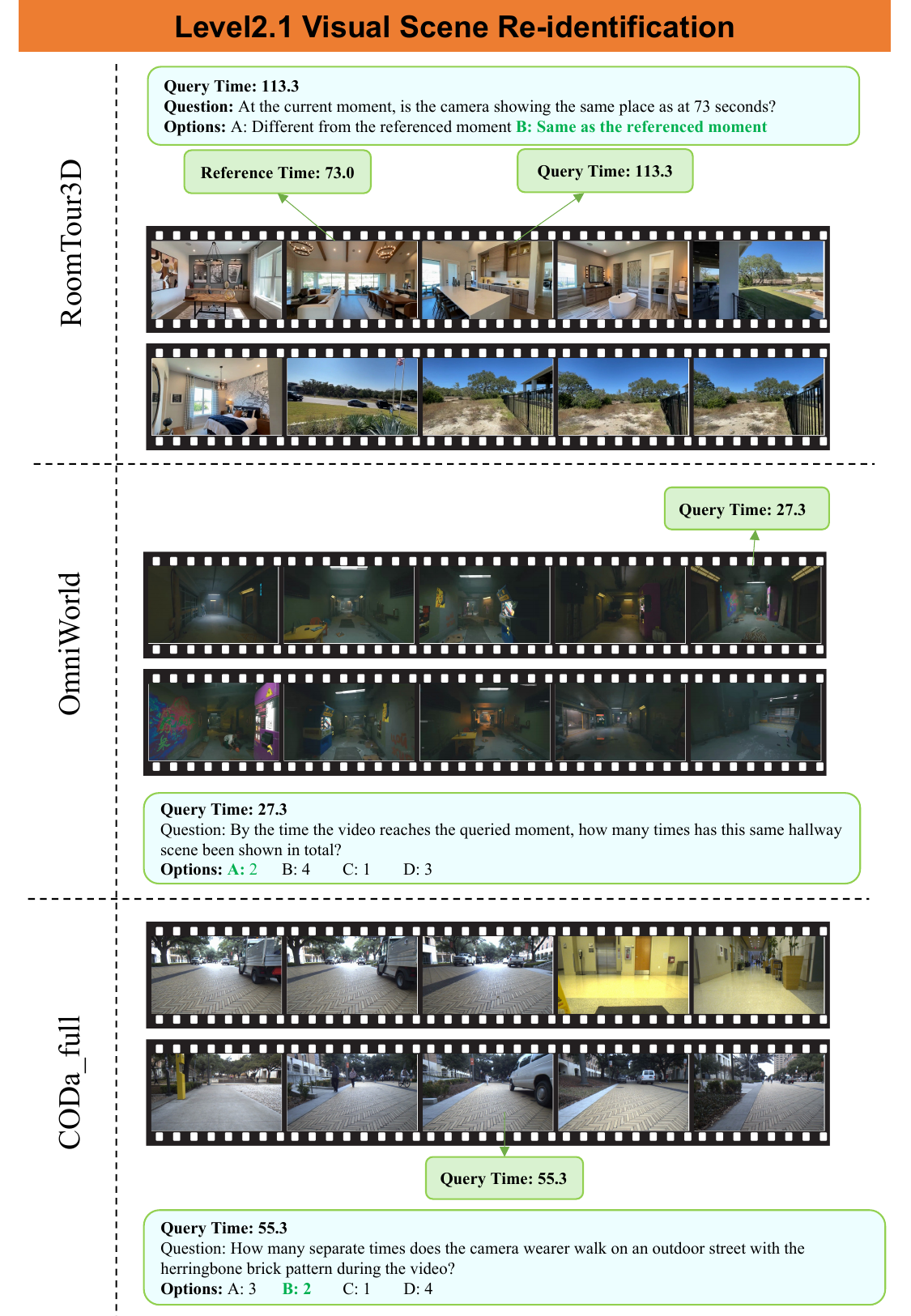}
\end{figure*}

\begin{figure*}[p]
\centering
\includegraphics[page=5,width=\textwidth,height=0.92\textheight,keepaspectratio]{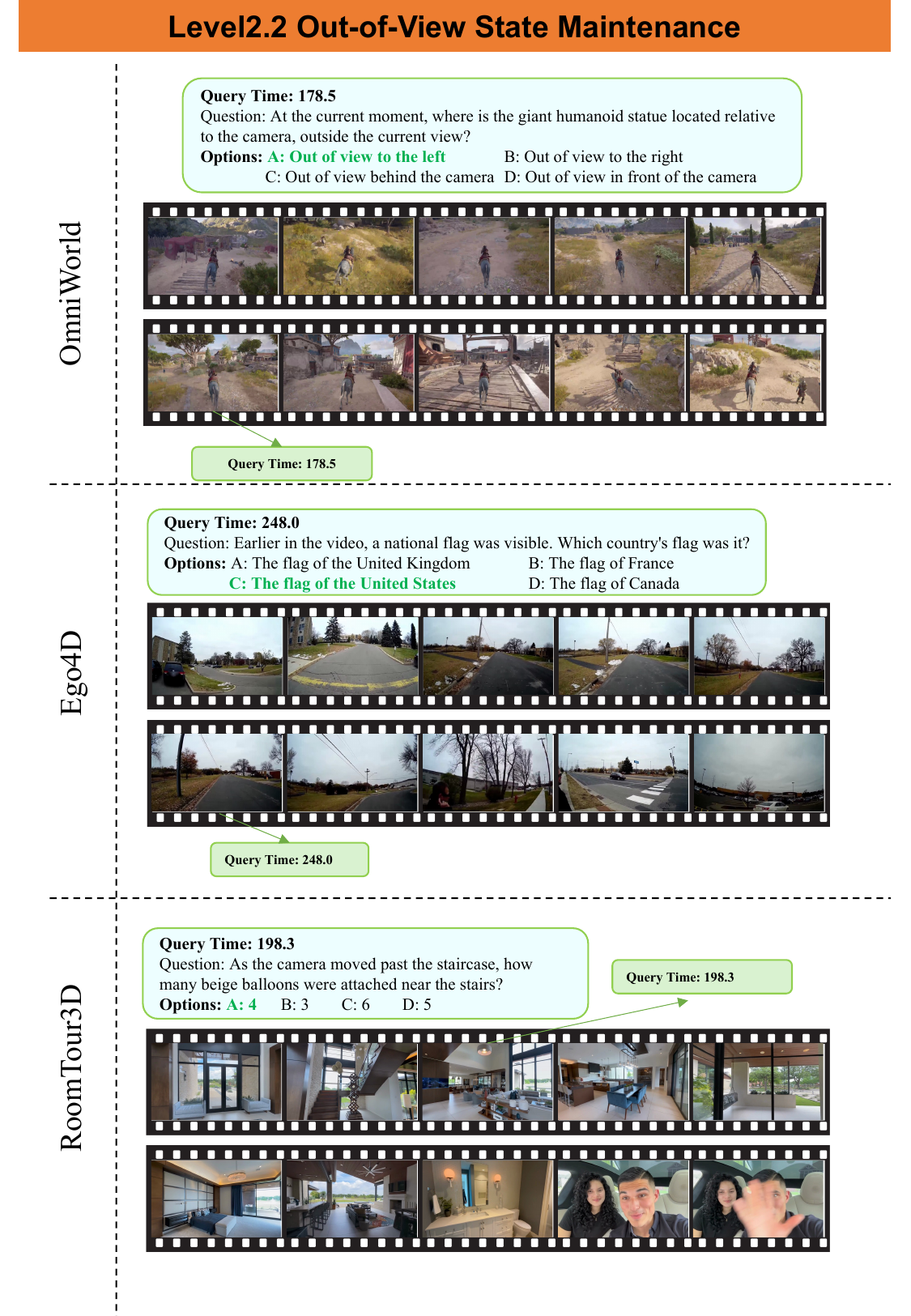}
\end{figure*}

\begin{figure*}[p]
\centering
\includegraphics[page=6,width=\textwidth,height=0.92\textheight,keepaspectratio]{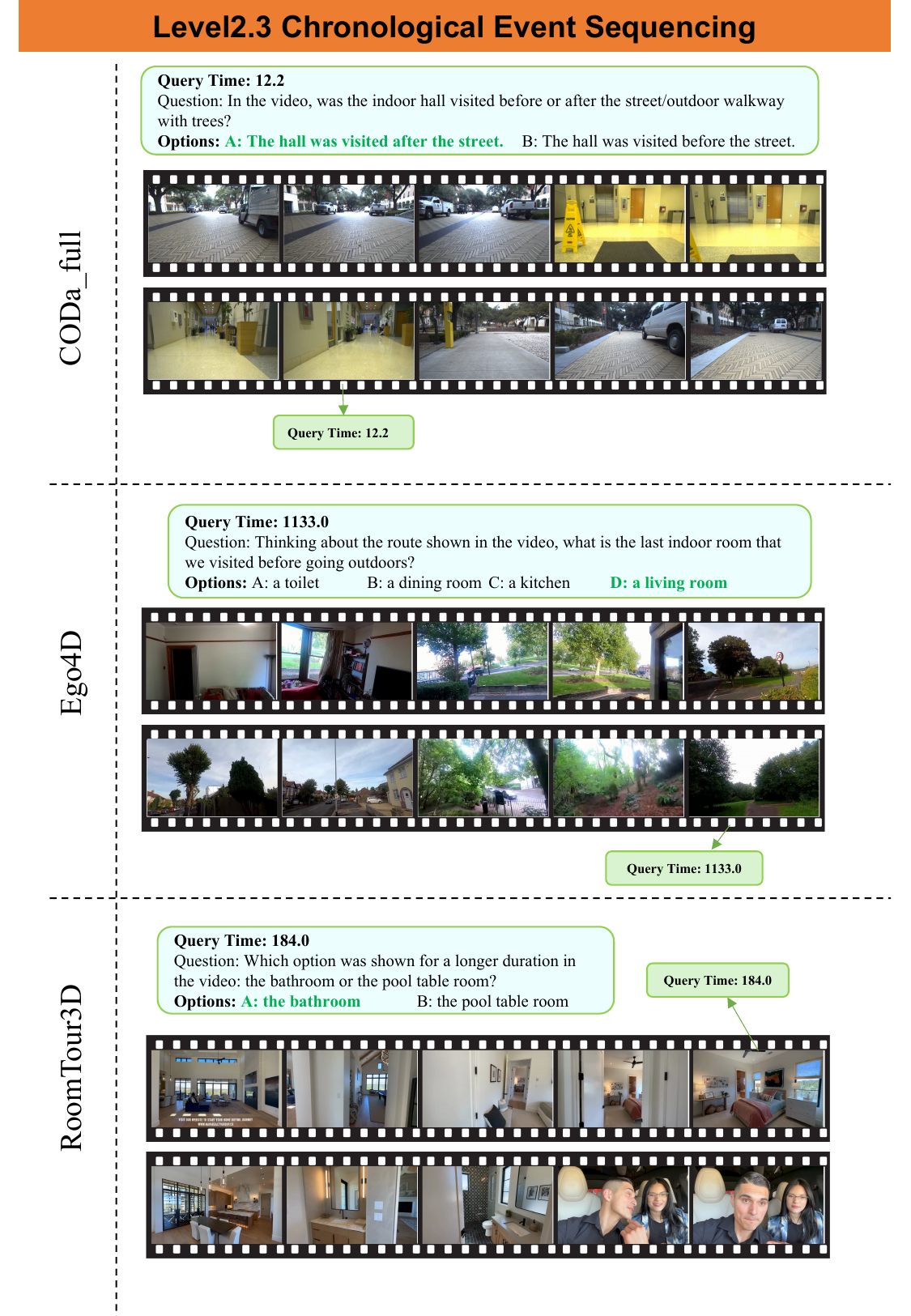}
\end{figure*}

\begin{figure*}[p]
\centering
\includegraphics[page=7,width=\textwidth,height=0.92\textheight,keepaspectratio]{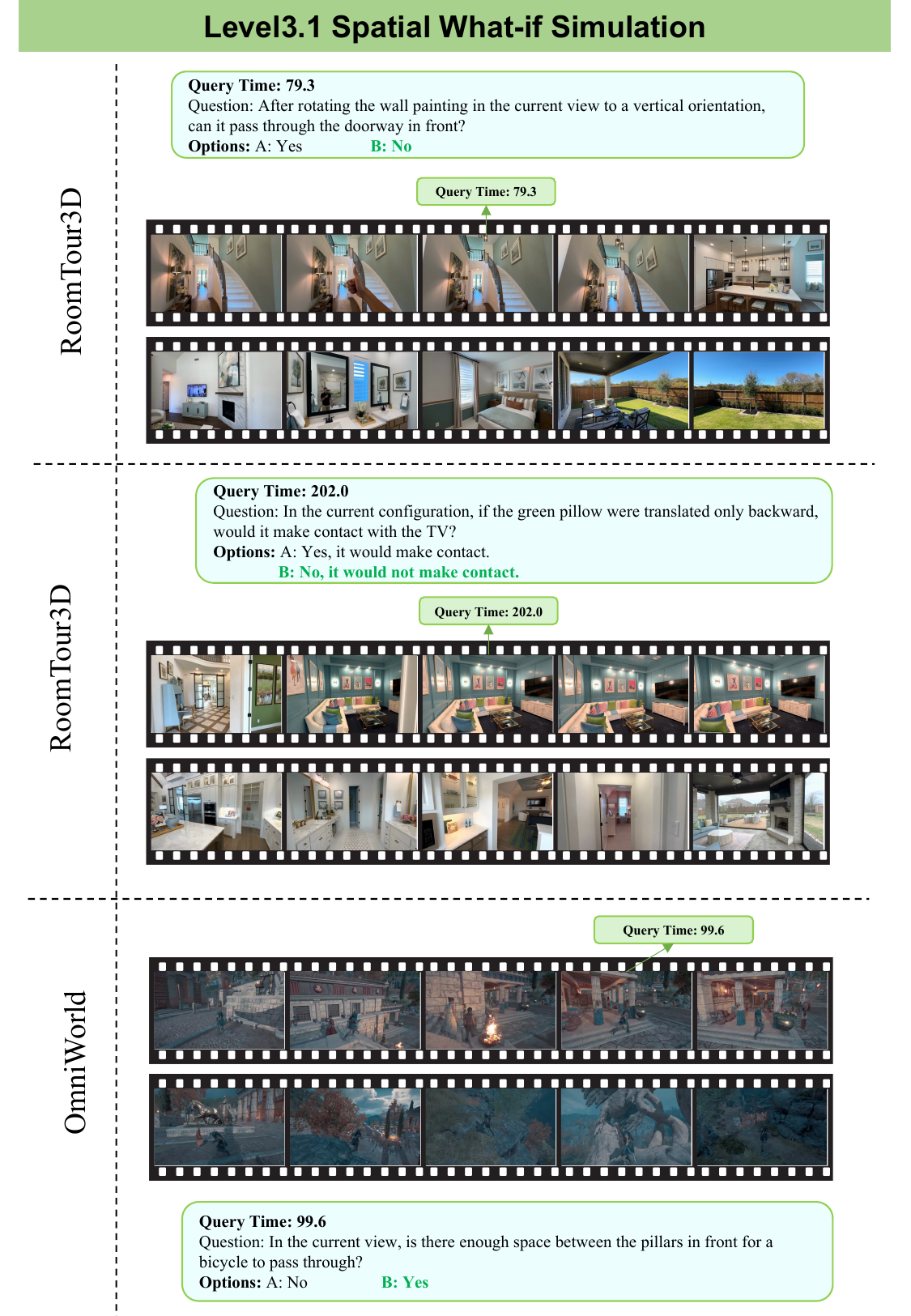}
\end{figure*}

\begin{figure*}[p]
\centering
\includegraphics[page=8,width=\textwidth,height=0.92\textheight,keepaspectratio]{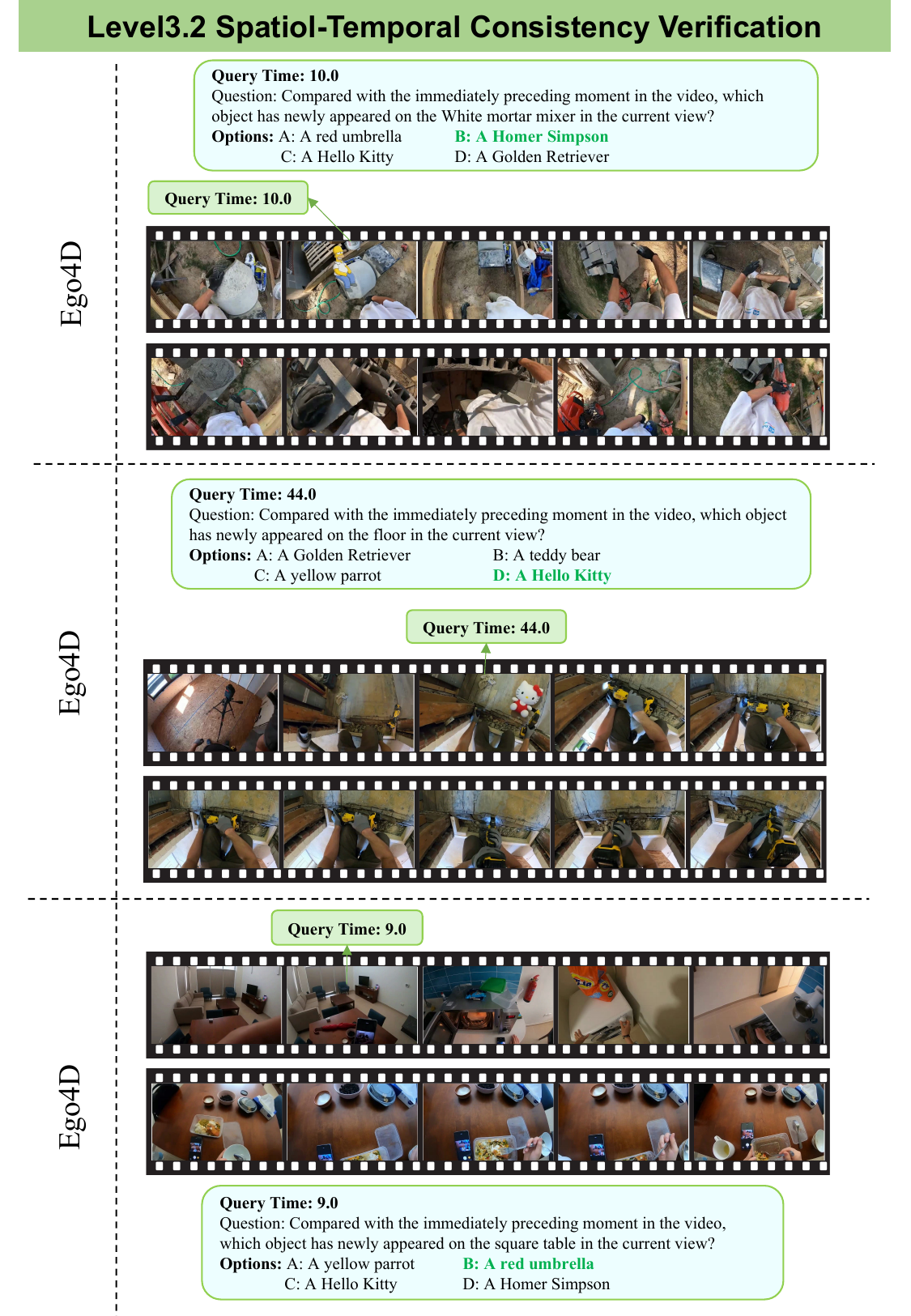}
\end{figure*}

\begin{figure*}[p]
\centering
\includegraphics[page=9,width=\textwidth,height=0.92\textheight,keepaspectratio]{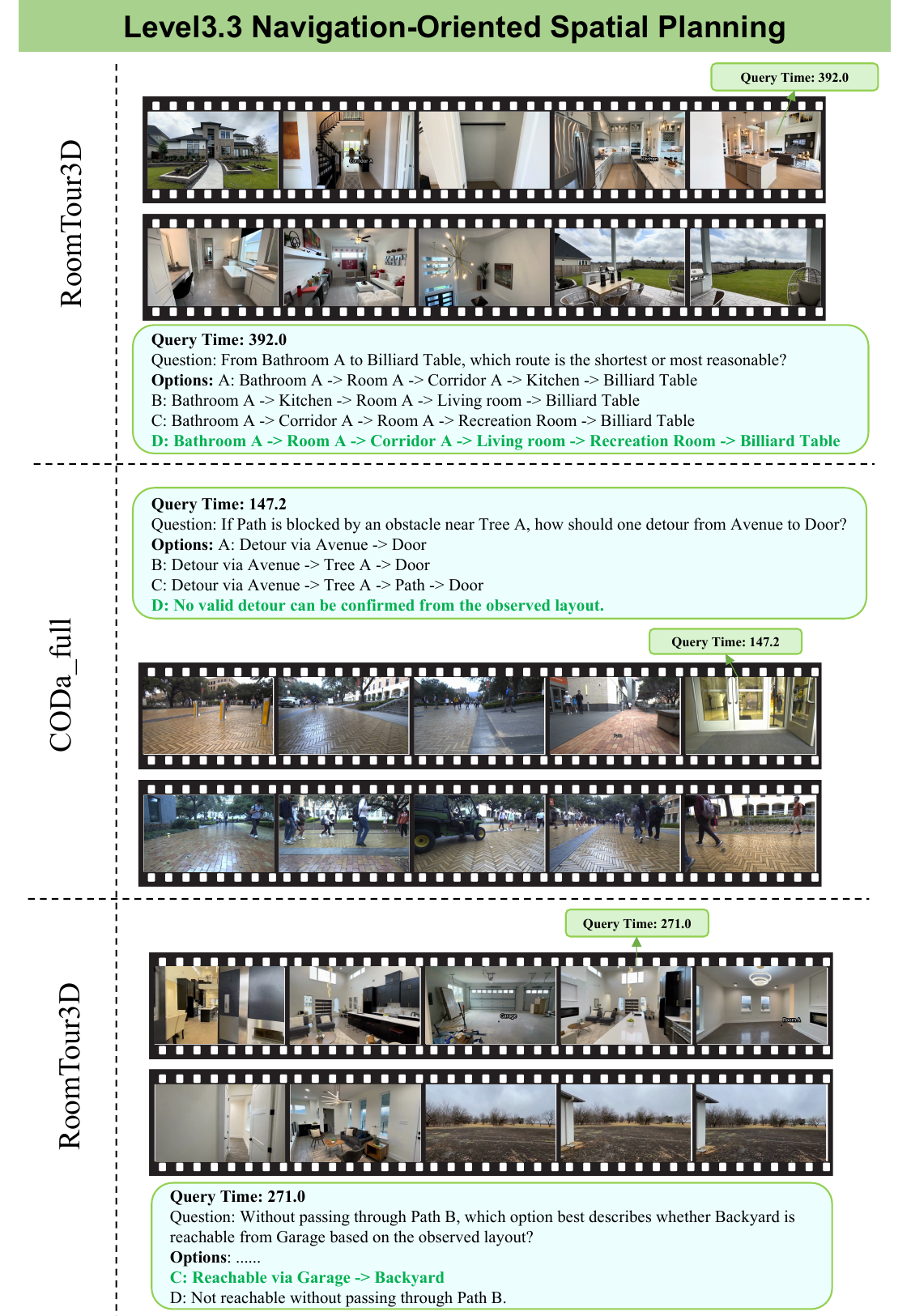}
\end{figure*}

\begin{figure*}[p]
\centering
\includegraphics[page=10,width=\textwidth,height=0.92\textheight,keepaspectratio]{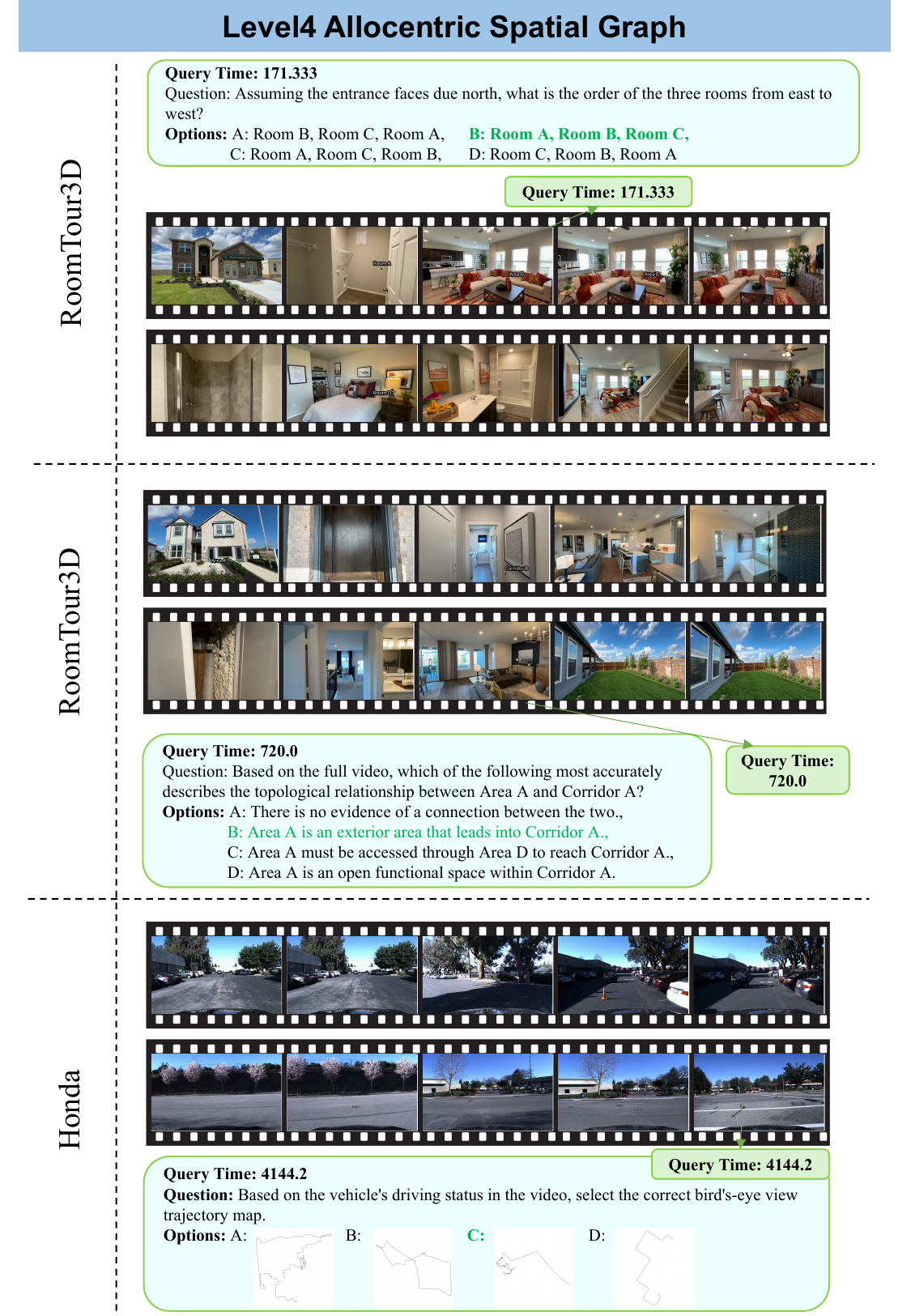}
\end{figure*}


\end{document}